\documentclass{article}

    \PassOptionsToPackage{numbers, compress}{natbib}

\usepackage[final]{neurips_2025}
\usepackage{amsmath}
\usepackage{longtable,array}
\usepackage{amssymb}
\usepackage[table]{xcolor}
\usepackage{breqn}
\usepackage{tabularx}
\usepackage[most]{tcolorbox}              
\usetikzlibrary{shadows}                  
\usepackage{enumitem,colortbl}
\usepackage{ltablex}     
\keepXColumns            

\usepackage[utf8]{inputenc} 
\usepackage[T1]{fontenc}    
\usepackage{hyperref}       
\usepackage{url}            
\usepackage{booktabs}       
\usepackage{amsfonts}       
\usepackage{nicefrac}       
\usepackage{microtype}      
\usepackage{xcolor}         
\usepackage{graphicx} 
\usepackage{booktabs}
\usepackage{multirow}
\usepackage{adjustbox}
\usepackage{booktabs}
\usepackage{tabularx}
\usepackage{subcaption}
\usepackage{xspace}
\usepackage{algorithm}
\usepackage{algpseudocode}
\usepackage{wrapfig}
\usepackage{caption}
\usepackage{amsmath}
\usepackage{placeins}
\def\model{Athena\xspace}

\newtcolorbox[auto counter, number within=section]{hintbox}[2][]{
  enhanced,
  colback=gray!10,           
  colframe=gray!60,          
  coltitle=black,          
  fonttitle=\bfseries,       
  arc=3mm,                   
  boxrule=0.8pt,             
  drop shadow={              
    shadow xshift=0.5mm,
    shadow yshift=-0.5mm,
    opacity=0.2
  },
  title={Prompt~\thetcbcounter: #2},  
  label=#1                   
}

\definecolor{blue}{gray}{0.40}
\definecolor{mygray}{gray}{0.75}
\definecolor{lightBlue}{gray}{0.90}

\newcolumntype{L}{>{\raggedright\arraybackslash}X} 
\newcolumntype{R}{>{\raggedright\arraybackslash}X} 

\title{Personalized Decision Modeling: Utility Optimization or Textualized-Symbolic Reasoning}

\author{
  Yibo Zhao\\
  Department of Civil and Systems Engineering \\
  Johns Hopkins University \\
  \And
  Yang Zhao\\
  Department of Civil and Systems Engineering \\
  Johns Hopkins University \\
  \AND
  Hongru Du \thanks{This work was completed while Hongru Du was at Johns Hopkins University.} ~\thanks{Correspondence to: Hongru Du and Hao Frank Yang.}\\
  Department of Systems and Information Engineering \\
  University of Virginia \\
  \texttt{hongrudu@virginia.edu}
  \And
  Hao Frank Yang \footnotemark[\value{footnote}]\\
  Department of Civil and Systems Engineering \\
  Johns Hopkins Data Science and AI Institute\\
  Johns Hopkins University \\
  \texttt{haofrankyang@jhu.edu}
}

\begin{document}

\maketitle

\begin{abstract}
Decision-making models for individuals, particularly in high-stakes scenarios like vaccine uptake, often diverge from population optimal predictions. This gap arises from the uniqueness of the individual decision-making process, shaped by numerical attributes (e.g., cost, time) and linguistic influences (e.g., personal preferences and constraints).
Developing upon Utility Theory and leveraging the textual-reasoning capabilities of Large Language Models (LLMs), this paper proposes an Adaptive Textual-symbolic Human-centric Reasoning framework (\textbf{\textsc{\model}}) to address the optimal information integration.
\textsc{\model} uniquely integrates two stages: First, it discovers robust, group-level symbolic utility functions via LLM-augmented symbolic discovery; Second, it implements individual-level semantic adaptation, creating personalized semantic templates guided by the optimal utility to model personalized choices. Validated on real-world travel mode and vaccine choice tasks, \textsc{\model} consistently outperforms utility-based, machine learning, and other LLM-based models, lifting F1 score by at least 6.5\% over the strongest cutting-edge models.  Further, ablation studies confirm that both stages of \textsc{\model} are critical and complementary, as removing either clearly degrades overall predictive performance. By organically integrating symbolic utility modeling and semantic adaptation, \textsc{\model} provides a new scheme for modeling human-centric decisions. The project page can be found at \url{https://yibozh.github.io/Athena}.


\end{abstract}

\section{Introduction}
\label{intro}


Consider the widely debated \textit{vaccine dilemma} \cite{fu2011imitation}, from a population-level perspective aimed at optimizing collective well-being (e.g., achieving herd immunity at minimal societal cost), models would invariably predict near-universal vaccine adoption. However, this population optimum consistently fails to predict actual individual behavior. The reality is a broad spectrum of personal choices, because each individual undertakes their own unique \textit{cognitive calculus}: balancing a vaccine's perceived efficacy and protection against its perceived risks \cite{bond2011making}, all filtered through their personal beliefs, constraints, and even considerations of potential "free-riding" on the immunity of others \cite{betsch2017benefits}. This divergence between the theoretical population optimum and observed individual actions highlights a critical limitation: \textbf{models designed to maximize collective outcomes do not adequately explain or predict individuals' choices.} Instead, individual choices are profoundly shaped by who we are, when the decision is made, and the unique constraints we face. This internal \textit{'cognitive calculus'}, unique to each individual and situation, presents a profound challenge for human decision modeling.

For decades, researchers have attempted to model this \textit{'cognitive calculus'} with Utility Theory \cite{von2007theory,rabin2013risk,quiggin2012generalized}, which assumes individuals select options that maximize expected gain. Operationally, this involves defining a parametric utility function, denoted as $f: \mathcal{X} \to \mathbb{R}$, that maps a vector of structured attributes $\mathcal{X}$ (e.g., monetary cost and time) for each option to a scalar utility score. Such pre-defined and explicit specifications of $f$ are the basis for classic discrete choice models \cite{ben1985discrete},
which have been widely adopted across economics \cite{small1981applied,haghani2021landscape,berbeglia2022comparative}, transportation \cite{ben1985discrete,zheng2021equality,salas2022systematic,ameli2022departure}, and public policy \cite{bier2020game,li2022government,loria2024public}. In these models, the utility scores derived from $f$ for each available option are used to probabilistically determine the likelihood of an individual selecting a particular option. However, even these utility-based models encounter fundamental barriers when attempting to capture the full depth of human decision-making. Real-world human decisions, as seen in the \textit{vaccine dilemma}, frequently deviate from these mathematical formulations. Individuals exhibit behaviors that appear inconsistent or irrational \cite{thomas2024modelling, glimcher2022efficiently}, yet these are often driven by subjective feelings and personal experiences. Such deviations reflect that traditional models, with their reliance on pre-specified functions, struggle to capture personalized decisions \cite{schoemaker1982expected}. The clue for deciphering this deviation is covered within individual attributes, some of which are structured and quantifiable, while others are unstructured and semantic (e.g., individual preference and constraints).

Addressing these unstructured and semantic dimensions, which are pivotal for capturing personalized decision mechanisms, calls for new modeling paradigms. LLMs with their strong textual-reasoning capability offer a clear advance \cite{achiam2023gpt}, providing new mechanisms for identifying the utility function $f$ and for integrating semantic individual context directly into the decision modeling process. Specifically, LLMs enhance our ability to model human choice by: a) Guiding the discovery of more accurate and robust parametric forms for $f$. Through LLM-augmented symbolic regression \cite{romera2024mathematical, llmsr, grayeli2024symbolic}, it becomes feasible to identify data-driven mathematical structures that capture underlying group-level choice patterns more effectively than pre-specified forms. b) Enabling the infusion of individual-level textual information into the human decision modeling \cite{aher2023using, schramowski2022large, hagendorff2023human}. By encoding personal preferences, constraints, and narratives, LLMs allow each decision to reflect the nuanced motivations and situational factors that traditional numeric features alone cannot convey.

This paper introduces an \textbf{Adaptive Textual-symbolic Human-centric Reasoning framework (\textsc{\model})}. \textsc{\model} achieves personalized decision modeling by uniquely integrating two sequentially structured steps: First, at the group level, it focuses on discovering robust, symbolic utility functions. Second, it implements individual-level, LLM-powered semantic adaptation guided by optimal utility functions discovered in previous steps. The outcome is a customized semantic template for each person, specifically designed to empower an LLM to model their choices by incorporating their unique preferences and constraints. We empirically validate \textsc{\model} on two real-world human decision-making tasks: travel mode choice and vaccine uptake decisions. The model consistently outperforms traditional utility-based, machine learning, and LLM-based models, with at least $6.5\%$ improvement in F1 score. Further ablation experiments reveal that removing either the group-level symbolic utility search or the individual semantic adapter lowers performance by at least $18\%$, underscoring the merit of the full \textsc{\model} framework. 




\vspace{-1em} 
\section{Related Work}
\label{gen_inst}

\textbf{Utility-based Decision-Making Models.} Initial explorations into this complex domain were predominantly by utility-based models \cite{Ramos02012014,su13084332,DeVos2016,Ettema2016}. These methods aim to capture human behavior within explicit mathematical functions, formulated from empirical data. Established methodologies such as Discrete Choice Models (DCMs) have been widely used, attributable to their interpretability and robust statistical underpinnings \cite{vij2024hybrid,berbeglia2022comparative,lancsar2017discrete,wang2025rational,yu2012four,habib2023rational,deneke2024transportation,10077454}. While offering tractability, they may also limit the ability to fully capture complex non-linear patterns and diverse preferences in modern high-dimensional data.

\textbf{Machine Learning–Driven Decision-Making Models.} ML-driven decision-making models aim to directly learn from rich, diverse features. Tree-based ensemble methods, including Random Forests \cite{zhang2023understanding}, Gradient-Boosting Trees \cite{pineda2022assessing}, XGBoost \cite{kim2021analysis,martin2023prediction}, and LightGBM \cite{su151411414}, alongside neural network architectures \cite{tang2024predicting,koushik2025explaining,9954273}, exhibited a notable proficiency in fitting complex, non-additive interaction effects. These models effectively integrated large-scale data, but their decision-making processes often lacked transparency despite strong predictive performance. Efforts to enhance transparency via post-hoc explanation frameworks, for instance, SHAP \cite{SHAPdeep2024,SADEGHI2024109370,e23010018} and Integrated Gradients \cite{unknown,IntegratedDecision}, have provided some insights for human behavior. A persistent challenge is these models’ vulnerability to distribution shifts, lack of transparency, and limited ability to provide symbolic, interpretable insights needed for personalized utility reasoning.


\textbf{Symbolic Regression with LLMs.} 
Classical symbolic regression (SR) typically uses genetic programming to evolve populations of candidate equations via stochastic mutation and crossover \cite{augusto2000symbolic,zhong2018multifactorial,skanderova2023self}. The goal is to find formulas that balance simplicity, generalizability, and human interpretability \cite{pysr,10.1609/aaai.v33i01.33014780,10.1007/s10462-023-10622-0}. The recent rise of LLMs has revitalized symbolic regression, enabling new possibilities in scientific discovery when combined with advanced evolutionary algorithms \cite{ye2024reevo}. For example, LLM-SR integrates LLM with evolutionary symbolic regression by treating equations as executable programs. It leverages LLMs’ scientific prior knowledge and code generation abilities to iteratively generate, refine, and optimize equation skeletons \cite{llmsr}. LASR integrated LLM-driven abstract textual concepts within evolutionary frameworks, achieving notable performance enhancements on benchmarks, such as the Feynman equation set \cite{NEURIPS2024_4ec3ddc4}. The DiSciPLE framework extended these contributions by emphasizing the interpretability and reliability of generated scientific hypotheses, incorporating critical evaluation and simplification to ensure hypotheses are both scientifically rigorous and computationally efficient \cite{disciple-25}.


\textbf{LLM-based Decision-Making Models.} The advent of LLMs has offered a new opportunity, establishing these models as human-like reasoning engines \cite{guo2025deepseek,grattafiori2024llama3herdmodels,jiang2023mistral7b,jiang2024mixtralexperts,geminiteam2024gemini15unlockingmultimodal,openai2024gpt4technicalreport}. Techniques such as instruction tuning \cite{zhang2023instruction,longpre2023flan,peng2023instruction,liu2023visual}, chain-of-thought reasoning \cite{wei2022chain,lyu2023faithful,xia2024beyond,NEURIPS2024_00d80722,chen2024unlocking} are elevating LLMs move beyond basic text generation to handle more complex tasks involving step-by-step reasoning and symbolic or numerical problem-solving  \cite{ouyang2022traininglanguagemodelsfollow,yao2023treethoughtsdeliberateproblem,wang2023selfconsistencyimproveschainthought,yao2023reactsynergizingreasoningacting,gao2023retrieval,shazeer2017outrageously,bi2024program}. 
Within the specific domain of personalized decision making, preliminary findings suggest that zero-shot and few-shot prompting strategies can enhance the behavioral alignment of LLMs \cite{liu2024travel,mo2023llmtravel}. Because a model’s knowledge is inherited from generic pre-training priors, its reasoning defaults to universally salient factors -- e.g., cost and time in travel mode choice -- while overlooking personal preferences such as rail-pass loyalty or transfer aversion, thereby introducing systematic bias \cite{gallegos2024self}. Techniques like persona loading partially mitigate this gap by conditioning responses on inferred preference structures \cite{liu2024travel,coletta2024personallm}. 
Beyond basic prompting, decision-centric systems add explicit structure to improve reliability and transparency: \emph{DeLLMa} enumerates plausible states, elicits utilities via pairwise comparisons, and maximizes expected utility; \emph{STRUX} distills inputs into fact tables with self-reflective evidence; \emph{OptiGuide} compiles natural-language "what-if" queries into optimization code and invokes solvers; \emph{Agent-Driver} coordinates tool calls, commonsense/experience memory, and chain-of-thought planning; \emph{Personalized Oncology} evaluations show chat-LLMs still trail experts, motivating structured, evidence-grounded pipelines \cite{liu2025dellma,lu-etal-2025-strux,li2023large,mao2024a,benary2023leveraging}. Nevertheless, many deployments still treat LLMs as opaque scoring mechanisms, falling short of fully recovering explicit, personalized utility logic.

\section{Methods}
\label{method}





We consider a classic discrete choice problem, where an individual $i$ faces a finite set of choices $\mathcal{J} = \{1, 2, \dots, J\}$, where $J = |\mathcal{J}|$. The decision-making process assumes individuals select the option $j$ that maximizes their utility. Each individual's observed choice behavior is represented by a dataset $\mathcal{D} = \{(X_i, y_i)\}_{i=1}^N$. For each observation $i$, $X_i = \{x_{ij}\}_{j=1}^J$ represents the set of feature vectors, where $x_{ij} \in \mathbb{R}^K$ captures the features for choice $j$ and $y_i \in \mathcal{J}$ denotes the observed choice.
\begin{figure}
    \centering
    \includegraphics[width=\linewidth]{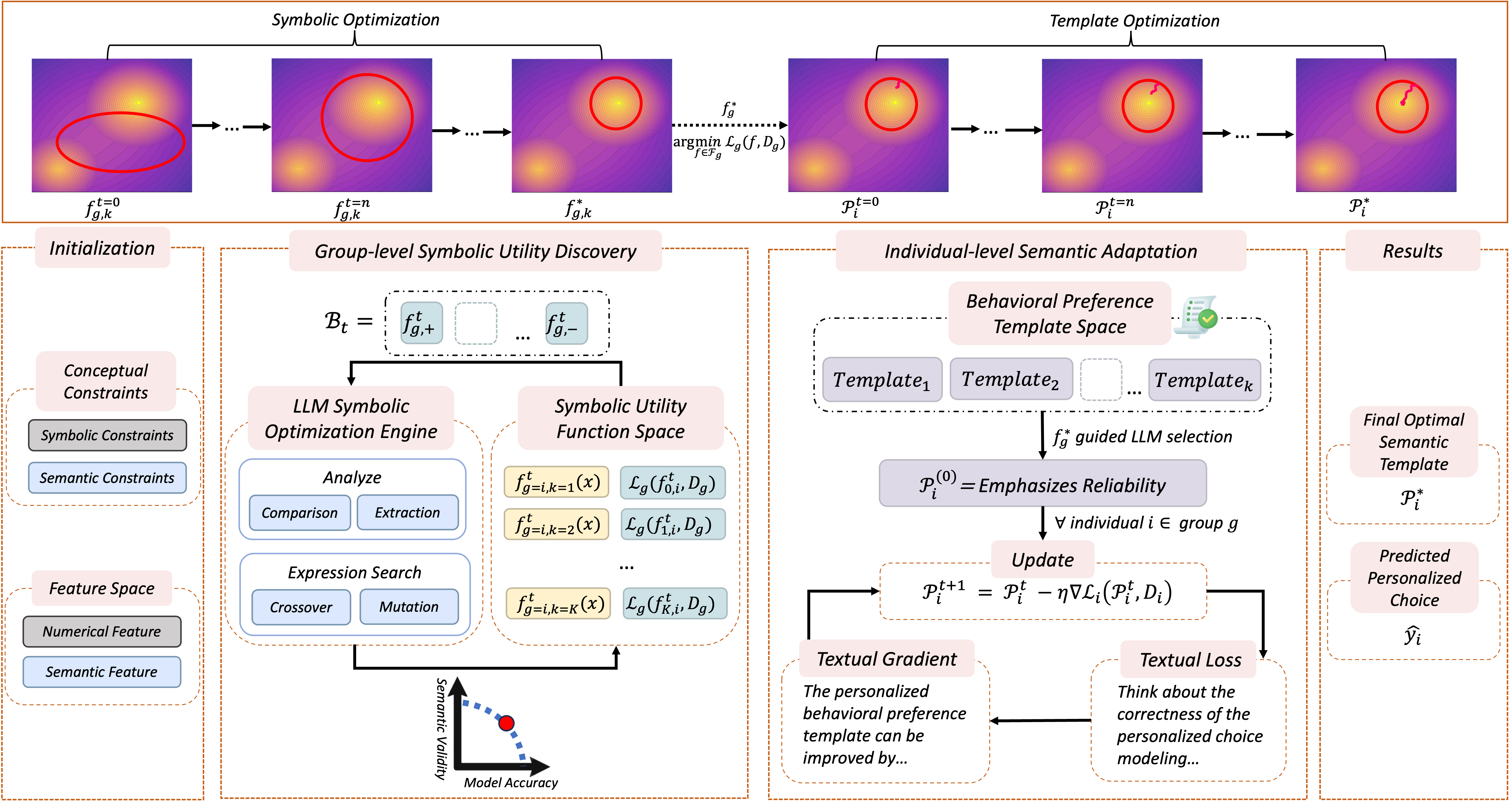}
    \caption{\textbf{Overview of the proposed \textsc{\model} framework.}
\textit{Group-level symbolic utility discovery}:  Symbolic \& semantic constraints library feed an LLM-driven symbolic-optimization engine that iteratively proposes candidate utility functions, scores them with loss $\mathcal{L}_{g}$, and prunes the search via analysis, crossover, and mutation.  Red rings in the contour maps illustrate how the feasible solution space shrinks across iterations until the optimal formula $f^{*}_{g}$ is selected.
\textit{Individual-level semantic adaptation}: The optimal group utility $f^{*}_{g}$ seeds a personalized template space.  For each individual $i$, TextGrad computes textual gradients of an individual loss and updates the template $\mathcal{P}^{t}_{i}$ into a more personalized decision rule $\mathcal{P}^{t+1}_{i}$. Finally, the optimal $\mathcal{P}^{*}_{i}$ is used to predict personal decisions.}
\label{fig:1}
\vspace{-3mm}
\end{figure}


The standard approach to modeling discrete choices is the Random Utility Maximization (RUM) framework \cite{manski1977structure}. It assumes the latent utility for each alternative $j$ is described by $U_{ij} = f(X_i, j) + \epsilon_{ij}$, where $f(X_i, j)$ is systematic component of utility and $\epsilon_{ij}$ is the random error. Assuming $\epsilon_{ij}$ are  independently and identically drawn (i.i.d.) and follow a Type I Extreme Value distribution \cite{mcfadden1972conditional}, the probability of individual $i$ choosing alternative $j$ is given by:

\begingroup
\setlength{\abovedisplayskip}{-3pt}      
\setlength{\belowdisplayskip}{6pt}      
\setlength{\abovedisplayshortskip}{0pt} 
\setlength{\belowdisplayshortskip}{3pt}
\begin{equation}
P(y_i = j \mid X_i) = \frac{e^{f(X_i, j;\theta)}}{\sum_{k \in \mathcal{J}} e^{f(X_i, k;\theta)}}.
\label{eq:choice}
\end{equation}
\endgroup


A central challenge lies in specifying the systematic utility component $f$. Traditional applications of RUM often rely on pre-specified functional forms for $f$ using domain expertise and observed data \cite{evans1991estimation}. This approach may result in a suboptimal representation of the true decision mechanism, while also neglecting individual heterogeneity in choices. Furthermore, traditional decision-making models are not designed to incorporate non-structured semantic information. To address these limitations, we introduce \textsc{\model} for personalized decision modeling designed to identify suitable utility function representations while simultaneously capturing individual-specific preferences.
As shown in Fig. \ref{fig:1}, \textsc{\model} structures the decision modeling process into two sequential stages:

\begin{enumerate}[leftmargin=*]
    \item \textbf{Group-Level Symbolic Utility Discovery:} This initial stage focuses on identifying optimal symbolic utility components that capture common decision patterns within distinct demographic groups. The discovery is achieved through a feedback-informed symbolic discovery process powered by LLMs.
    \item \textbf{Individual-Level Semantic Adaptation:} Then, the optimal group-level utility functions serve as guidance for the LLM-driven optimization of personalized semantic templates. This adaptation process is designed to incorporate individual-specific preferences and constraints, leveraging the rich semantic reasoning capabilities of LLMs.
\end{enumerate}

\subsection{Group-Level Symbolic Utility Discovery}
The first stage aims to discover an optimal parametric utility function, denoted as $f_g^*$, for each demographic group $g \in \mathcal{G}$. This function $f_g^*$ should be constructible from symbolic building blocks and optimally explain the group's choice behavior. Following the choice probability defined earlier Eq. \eqref{eq:choice}, the objective is to find the optimal $f_g^*$ and its associated parameters $\theta_g^*$ such that: 
\begin{equation} \label{eq:group_loss}
(\theta_g^*, f_g^*) = \underset{f, \theta}{\arg\min} \; \mathcal{L}_{g}(f(X_i, y_i; \theta); \mathcal{D}_g) = - \sum_{(X_i, y_i) \in \mathcal{D}_g} \log \left( \frac{e^{f(X_i, y_i; \theta)}}{\sum_{k \in \mathcal{J}} e^{f(X_i, k; \theta)}} \right).
\end{equation}


To automate the symbolic utility discovery of $f_g^*$, we design an iterative, feedback-informed generation process powered by LLMs. To effectively guide the automated discovery of utility functions, we constructed two foundational libraries: a domain knowledge concept library ($\mathcal{C}$) and a symbolic library ($\mathcal{S}$). The library $\mathcal{C}$, developed based on input from domain experts, covers high-level conceptual knowledge about domain-specific human behavior. The library $\mathcal{S}$ provides the fundamental syntactic building blocks needed for constructing all candidate utility expressions. 

Inspired by evolutionary algorithms \cite{back1993overview}, the core discovery process proceeds iteratively. In each iteration $t$ for each demographic group $g$, the LLM samples a set of $K$ candidate symbolic utility functions, $\{f_{g,k}^{t}\}_{k=1}^K$. This sampling is performed from the LLM's learned distribution $\phi$ \cite{achiam2023gpt, team2023gemini}, conditioned on the group profile $g$, the domain concept $\mathcal{C}$, the available symbolic building block $\mathcal{S}$, and a feedback $\mathcal{B}^{t-1}$ from preceding iteration:
\vspace{-1em} 

\begin{equation}
\{f_{g,k}^{t}\}_{k=1}^K \sim \phi(\cdot | g, \mathcal{C}, \mathcal{S}, \mathcal{B}^{t-1})
\end{equation}

The feedback $\mathcal{B}$ is essential in refining the LLM's sampling strategy. Specifically, $\mathcal{B}^{t}$ is constructed at the end of each iteration t and comprises the best-performing and worst-performing candidate functions from that iteration:
\vspace{-1.2em} 


\begin{equation}
\mathcal{B}^t = \{f_{g,+}^t, f_{g,-}^t\},
\label{eq:bt}
\end{equation}

where $f_{g,+}^t = \underset{k \in K}{\arg\min} \: \mathcal{L}_g(f_{g,k}^{t}, \mathcal{D}_g)$ and $ f_{g,-}^t = \underset{k \in K}{\arg\max} \: \mathcal{L}_g(f_{g,k}^{t}, \mathcal{D}_g)$. Here $\mathcal{L}_g$ is the group-level loss function, with a similar format as Eq. \eqref{eq:group_loss}.
This feedback $\mathcal{B}^t$ is used to refine the LLM's sampling distribution $\phi$ through stochastic \textit{mutation} or \textit{crossover} \cite{augusto2000symbolic,zhong2018multifactorial,skanderova2023self}, pushing the generation towards more promising types of functions. The iterative discovery process for group $g$ is considered to have converged at iteration $T$ if the absolute difference in the loss of the best-performing function from the current iteration $t$ and that of the previous iteration $t-1$ falls below a predefined threshold $\delta$:
\vspace{-0.05em} 
\begin{equation}
|\mathcal{L}_g(f_{g,+}^{t}, \mathcal{D}_g) - \mathcal{L}_g(f_{g,+}^{t-1}, \mathcal{D}_g)| < \delta.
\label{eq:iteration_stop}
\end{equation}

Upon convergence at iteration at $T$, the optimal group-level symbolic utility function ($f_g^*$) is determined as the function that achieved the minimum loss across all generated candidate functions throughout the entire iterative process:

\begin{equation}
f_g^* = \arg \min_{f \in \mathcal{F}_g} \mathcal{L}_g(f, D_g),
\end{equation}

where $\mathcal{F}_g = \bigcup_{t=1}^{T}\{f_{g,k}^{t}\}_{k=1}^K$ and $T$ is the iteration at which convergence occurred. This discovered function $f_g^*$, along with its fitted parameters $\theta_g^*$, serves as the learned representation of the systematic utility for group $g$.

\subsection{Individual-Level Semantic Adaptation}

Following the determination of the group-level optimal symbolic utility functions $f_g^*$, the framework transitions to the second stage,  leveraging an LLM conditioned on $f_g^*$ to model individual choice behavior more accurately. While $f_g^*$ captures the central tendencies of utility for group $g$, significant intra-group heterogeneity often persists. To account for this, we introduce an individual-level adaptation stage to personalize the utility representation by generating and refining an individual-specific semantic template. 

For each individual $i \in g$, the initial semantic template, denoted as $\mathcal{P}_i^{0}$ is generated by the LLM ($\phi$). The generation of the initial semantic template is represented as a sampling process from the LLM's distribution: $\mathcal{P}_i^{0} \sim \phi( \cdot | f_g^*, i, \mathcal{C})$.  
In this formulation, $\phi$ conditions on the optimal group-level symbolic function $f_g^*$, the specific individual context $i$, and the high-level domain concepts from $\mathcal{C}$ to generate $\mathcal{P}_i^{0}$. This initial template $\mathcal{P}i^{0}$ is a semantic representation that is designed to be adaptable in subsequent optimization steps. Then, the semantic template $\mathcal{P}_i^{0}$ undergoes an iterative refinement process for each individual $i$. This optimization is driven by TextGrad \cite{yuksekgonul2025optimizing}, which optimizes the template based on the individual's specific data $\mathcal{D}_i = (X_i, y_i)$. The update rule is given by:

\begin{equation}
\mathcal{P}_i^{t+1} \leftarrow 
\mathcal{P}_i^{t} - \eta \nabla \mathcal{L}_{i}(\mathcal{P}_i^{t}, D_i).
\label{eq:textgrad}
\end{equation}

The term $\nabla \mathcal{L}_{i}(\mathcal{P}_i^{t}, D_i)$ represents the "textual gradient" of the loss function with respect to the semantic template $\mathcal{P}_i^{t}$. Since $\mathcal{P}_i^{t}$ is the textual template, this gradient is not a vector of partial derivatives in the mathematical sense. Instead, it indicates the direction and nature of textual modifications to $\mathcal{P}_i^{t}$ that would lead to the most improvement in loss. This iterative refinement process continues until a maximum number of iterations $T'$ is reached. Then the final optimal semantic template for individual $i$, denoted as $\mathcal{P}_i^{*}$, is determined. The predicted personalized choice $\hat{y}_i$ is then represented as sampling from the LLM's output distribution: 

\begingroup
\setlength{\abovedisplayskip}{3pt}      
\setlength{\belowdisplayskip}{6pt}      
\setlength{\abovedisplayshortskip}{0pt}
\setlength{\belowdisplayshortskip}{3pt}
\begin{equation}
\hat{y}_i \sim \phi(\underbrace{\mathcal{P}_i^*, X_i}_{\text{Semantic Adaptation}}
\,|\,\overbrace{f_g^*(X_i; \theta_g^*)}^{\text{Symbolic Utility Discovery}})
\label{eq:pred}
\end{equation}
\endgroup

The overall procedure of \textsc{\model} is summarized in Algorithm~\ref{alg:athena}.

\section{Experiments}
\label{exp}





\begin{wrapfigure}{R}{0.6\textwidth}  
\vspace{-50pt}
\begin{minipage}{0.57\textwidth}
\begin{algorithm}[H]
\footnotesize
\caption{\textsc{\model}\ Optimization Flow}
\label{alg:opt}
\begin{algorithmic}[1]
\Require Demographic group $g$, dataset \(\mathcal{D}_g\), domain concept $\mathcal{C}$, symbolic building block $\mathcal{S}$
\State Initialize \(\mathcal{B}_0\leftarrow\texttt{None}\)

\Statex \textbf{ \textit{// Stage~1: Group-Level Symbolic Utility Discovery}}
\For{$t = 1$ to $T$}
    \State Sample symbolic utility functions \(\{f_{g,k}^{t}\}_{k=1}^K \sim \phi(\cdot \mid g, \mathcal{C}, \mathcal{S}, \mathcal{B}^{t-1})\)
    \State Update \(\mathcal{B}^t \leftarrow \{f_{g,+}^t, f_{g,-}^t\}\) using Eq.~\eqref{eq:bt}
    \State Select best function \(f_g^* \leftarrow \arg\min_{f \in \mathcal{F}_g} \mathcal{L}_g(f, \mathcal{D}_g)\)
    \If{stopping condition in Eq.~\eqref{eq:iteration_stop} is met}
        \State \textbf{break}
    \EndIf
\EndFor
\Statex \textbf{ \textit{// Stage~2: Individual-Level Semantic Adaptation}}
\For{each individual \(i \in g\)}
    \State Initialize semantic template \(\mathcal{P}_i^{0} \sim \phi( \cdot \mid f_g^*, i, \mathcal{C})\)
    \For{$t = 1$ to $T'$}
        \State Update \(\mathcal{P}_i^{t+1} \leftarrow \mathcal{P}_i^{t} - \eta \nabla \mathcal{L}_i(\mathcal{P}_i^{t}, \mathcal{D}_i)\) using Eq.~\eqref{eq:textgrad}
    \EndFor
\EndFor
\State \Return \(\{\mathcal{P}_i^{*}\}_{i \in g}\), predict decisions using Eq. (\ref{eq:pred}).
\end{algorithmic}
\label{alg:athena}
\end{algorithm}
\end{minipage}
\vspace{-30pt}
\end{wrapfigure}

\begin{figure}
    \centering
    \includegraphics[width=1\linewidth]{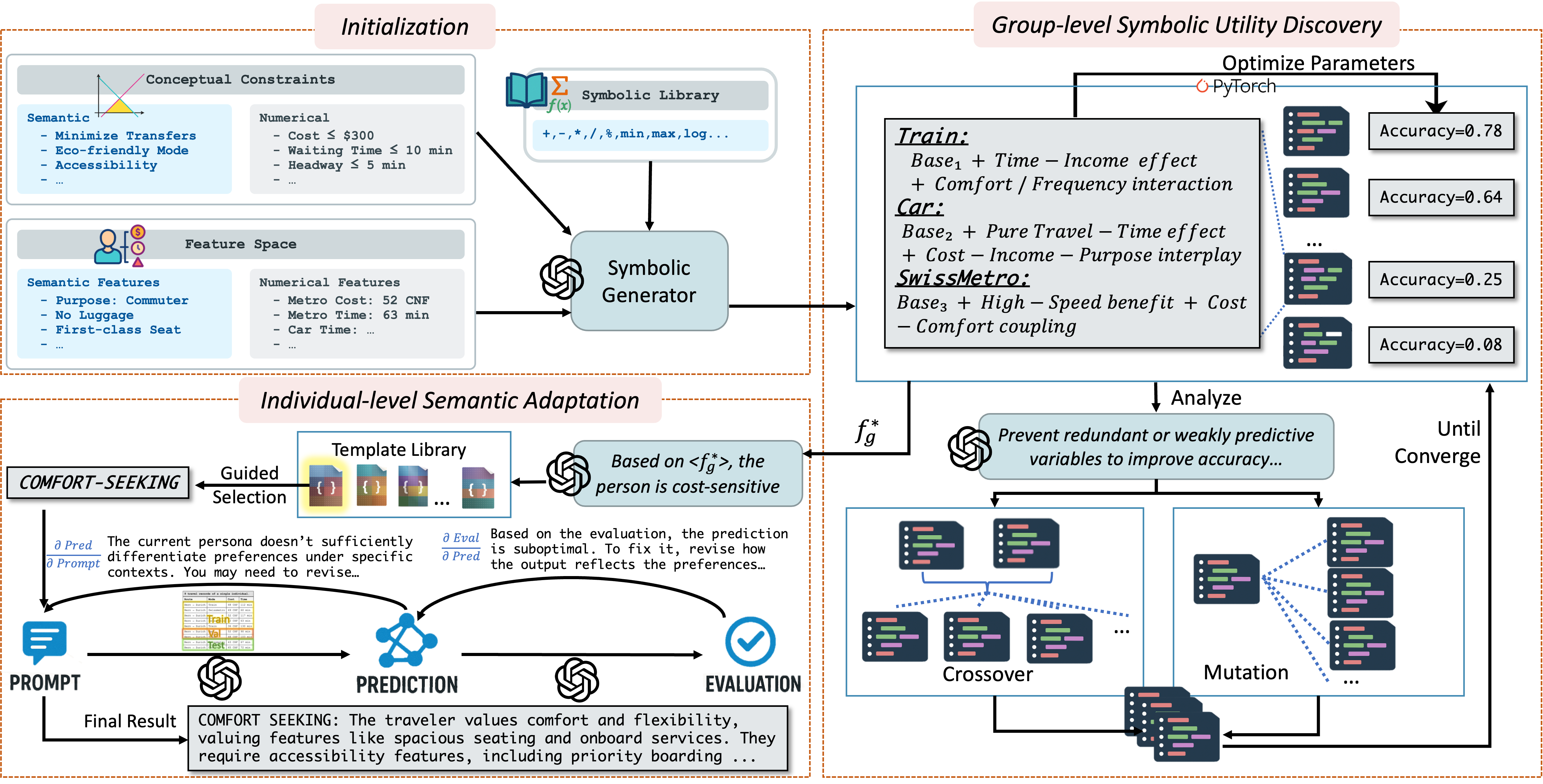}
    \caption{\textbf{\textsc{\model} pipeline applied to a travel–mode choice example.} Here we use \textit{\textbf{Swissmetro}} as an example to illustrate \textsc{\model} framework. The \textit{Initialization} panel encodes conceptual constraints, a mixed semantic–numerical feature space, and a symbolic library of operations. In \textit{Group‑level symbolic optimization}, an LLM samples, scores, and prunes candidate utility expressions for each alternative to produce compact formulas $\{f_g^*\}$ that best explain group behavior. In \textit{Individual semantic adaptation}, each $f_g^*$ seeds a group‑specific prompt template $\mathcal{P}_i^0$, which is refined to a personalized template via TextGrad to capture individual heterogeneity ($\mathcal{P}_i^0 \rightarrow \mathcal{P}_i^*$).}
    \label{fig:pipeline}
\end{figure}

This section empirically validates the value of \textsc{\model}, demonstrating its overall effectiveness in personalized decision-making and its robust capability to apply across diverse application domains. We break down our experimental findings to specifically showcase the distinct value added by each core component of the \textsc{\model} framework: 1) group-level symbolic utility discovery and 2) personalized semantic template adaptation. 
Fig.~\ref{fig:pipeline} illustrates the full pipeline using the travel-mode choice as an example.



\subsection{Experimental Setup}
\textbf{Datasets.}
To test \textsc{\model}’s ability to generalize across different domains and to adapt to individual preferences, we selected two real-world tasks that reflect fundamentally different personalized decision scenarios: daily transportation choices and public health decisions. \textbf{(1) Swissmetro Transportation Choice (\textit{Swissmetro})}: is a widely used benchmark in travel mode choice modeling \cite{swissmetro,pham2022causality,wen2025new,ghorbani2025enhanced,haj2025incorporating}. Each record details a trip between major Swiss cities and includes both traveler characteristics (e.g., income, age) and alternative-specific attributes (e.g., travel time, cost). The dataset has a potential choice set of three transportation modes: \textit{Train}, \textit{Car}, and \textit{Metro}. \textbf{(2) COVID-19 Vaccination Choice (\textit{Vaccine})}: This dataset is derived from a large-scale international survey,  conducted across multiple countries \cite{lazarus2023survey}. The survey was designed to understand factors influencing COVID-19 vaccine uptake and attitudes. For each participant, it captures demographics, prior beliefs about the vaccine, and their self-reported vaccination status. The modeled choices based on this information include: \textit{Unvaccinated}, \textit{Vaccinated initial doses, no booster}, and \textit{Vaccinated initial doses plus booster}.



\textbf{Experiment Configurations.}
To maintain a reasonable budget for the template-adaptation stage, we restricted the experimental sample to a representative subset of each dataset. Specifically, we used: \textit{(1) \textbf{Swissmetro}}: 500 travelers, two trip records per person; \textit{(2) \textbf{Vaccine}}: 300 respondents, one survey record per person. Within each dataset, we first identified key demographic dimensions (gender, age, and income), then sampled approximately balanced subsets across these strata from the full dataset. This ensures \textit{(i)} comparable class priors between training and test splits, and \textit{(ii) }that no demographic group dominates the symbolic-utility discovery process. The predefined demographic grouping follows established practice in choice modeling, supports interpretability, and improves robustness by avoiding the complexity and data requirements of latent clustering methods \cite{10.1257/jel.47.2.448,doi:10.1073/pnas.1207144109,deBruijn2021PovertyAE}.

\textbf{Evaluation metrics.} We report Accuracy, F1, AUC, and Cross-Entropy (CE). CE is included because \textsc{\model} produces probabilistic predictions over choices. A lower CE means the model assigns higher probabilities to actual choices, while F1 and AUC capture classification performance; together, they provide complementary views on accuracy and calibration.

\textbf{Models and baselines.}
Both stages of \textsc{\model}, symbolic-utility discovery and individual semantic adaptation, run on the \texttt{gpt-4o-mini-2024-07-18} and \texttt{gemini-2.0-flash}.  To evaluate its performance, we contrasted \textsc{\model} with three baseline groups.  \textit{(i)} LLM-based methods: a plain zero-shot method \cite{kojima2022large,mo2023large}, a zero-shot chain-of-thought method \cite{kojima2022large}, a five-example few-shot method \cite{brown2020language,liu2024can}, and TextGrad tuning \cite{yuksekgonul2025optimizing}.  \textit{(ii)} Classical discrete-choice models: Multinomial Logit (MNL) \cite{huang2024comparative}, Conditional Logit (CLogit) \cite{truong2021travel}, and Latent-Class MNL \cite{wu2025evaluating}.  \textit{(iii)} Standard machine-learning classifiers: logistic regression, random forest, XGBoost \cite{kashifi2022predicting}, a shallow two-layer MLP \cite{goodfellow2016deep}, TabNet for tabular data \cite{arik2021tabnet}, and a fine-tuned BERT classifier \cite{devlin2019bert}.  This spectrum ranges from end-to-end language-model reasoning through discrete choice models to conventional predictive learners, providing a balanced reference for unique modeling capabilities.

\begin{table}[ht]
  \centering
  \scriptsize
  \caption{Performance comparison of LLM-based, classical choice, and machine learning methods on the three-class Swissmetro and three-class COVID-19 Vaccine choice tasks.}
  \begin{adjustbox}{max width=\textwidth}
    \begin{tabular}{llp{1.8cm}cccccccc}
      \toprule
      & Method      & LLM Model 
        & \multicolumn{4}{c}{Swissmetro} 
        & \multicolumn{4}{c}{Vaccine} \\
      \cmidrule(lr){4-7} \cmidrule(lr){8-11}
      &             & 
        & Acc.$\uparrow$ & F1.$\uparrow$ & CE.$\downarrow$ & AUC.$\uparrow$
        & Acc.$\uparrow$ & F1.$\uparrow$ & CE.$\downarrow$ & AUC.$\uparrow$ \\
      \midrule
      \multirow{10}{*}{\shortstack{\textbf{LLM-}\\\textbf{Based}}}
      & \multirow{2}{*}{Zeroshot}
        & \shortstack{gemini-2.0-flash}
            & 0.5920 & 0.2940 & 0.9257 & 0.6561
            & 0.5800 & 0.5092 & 0.8328 & 0.7607 \\
      &                          & \shortstack{GPT-4o-mini}
            & 0.6300 & 0.2757 & 2.7258 & 0.3657
            & 0.5433 & 0.5387 & 0.8562 & 0.7395 \\
      & \multirow{2}{*}{Zeroshot-CoT}
        & \shortstack{gemini-2.0-flash}
            & 0.5880 & 0.3478 & 0.9415 & 0.6331
            & 0.5800 & 0.5073 & 0.8436 & 0.7526 \\
      &                              & \shortstack{GPT-4o-mini}
            & 0.6420 & 0.2960 & 0.8957 & 0.6237
            & 0.5500 & 0.5353 & 0.8540 & 0.7465 \\
      & \multirow{2}{*}{Fewshot}
        & \shortstack{gemini-2.0-flash}
            & 0.7580 & 0.7027 & 8.7244 & 0.7956
            & 0.5667 & 0.5740 & 12.0324 & 0.7053 \\
      &                         & \shortstack{GPT-4o-mini}
            & 0.6815 & 0.4945 & 7.0029 & 0.7395
            & 0.5067 & 0.5097 & 6.6110 & 0.6891 \\
      & \multirow{2}{*}{TextGrad}
        & \shortstack{gemini-2.0-flash}
            & 0.5568 & 0.2980 & 1.2011 & 0.5400
            & 0.4241 & 0.4014 & 5.7813 & 0.6363 \\
      &                         & \shortstack{GPT-4o-mini}
            & 0.6500 & 0.3620 & 0.9079 & 0.5364
            & 0.5084 & 0.4962 & 4.5412 & 0.6709 \\
      & \multirow{2}{*}{\textbf{\textsc{\model}}}
        & \shortstack{gemini-2.0-flash}
            & 0.7679 & 0.7222 & 0.9041 & 0.8387
            & 0.6797 & 0.5968 & 0.7610 & 0.8370 \\  
      &                              & \shortstack{GPT-4o-mini}
            & \textbf{0.8134} & \textbf{0.7655} & 1.0863 & \textbf{0.8825}
            & \textbf{0.7345} & \textbf{0.7161} & 0.7551 & \textbf{0.8704} \\
      \midrule
      \multirow{3}{*}{\shortstack{\textbf{Utility}\\\textbf{Theory}}}
      & MNL              & \centering / 
            & 0.6101 & 0.3887 & 0.8353 & 0.7074
            & 0.4150 & 0.1955 & 1.0510 & 0.4301 \\
      & CLogit           & \centering /
            & 0.5714 & 0.2424 & 0.8916 & 0.5976
            & 0.4150 & 0.1955 & 1.0510 & 0.5000 \\
      & Latent Class MNL & \centering /
            & 0.6101 & 0.3967 & 0.8175 & 0.7182
            & 0.1950 & 0.1088 & 1.0986 & 0.5000 \\
      \midrule
      \multirow{6}{*}{\shortstack{\textbf{Machine}\\\textbf{Learning}}}
      & Logistic Regression & \centering /
            & 0.5620 & 0.5570 & 0.9310 & 0.7460
            & 0.6500 & 0.6690 & 0.7630 & 0.8330 \\
      & Random Forest       & \centering /
            & 0.7100 & 0.7050 & 0.7380 & 0.8810
            & 0.6300 & 0.6470 & \textbf{0.7290} & 0.8420 \\
      & XGBoost             & \centering /
            & 0.7080 & 0.7050 & 0.7040 & 0.8810
            & 0.6300 & 0.6480 & 1.1420 & 0.8150 \\
      & BERT & \centering /
            & 0.7246 & 0.4994 & \textbf{0.7037} & 0.8811
            & 0.6350 & 0.6541 & 0.7409 & 0.8168 \\
      & TabNet       & \centering /
            & 0.6375 & 0.4060 & 0.7887 & 0.8810
            & 0.6650 & 0.6684 & 0.8968 & 0.8147 \\
      & MLP             & \centering /
            & 0.6475 & 0.6386 & 0.7626 & 0.8350
            & 0.6068 & 0.6062 & 0.9320 & 0.8205 \\
      \bottomrule
    \end{tabular}
  \end{adjustbox}
  \label{tab:baseline_comparison}
  \vspace{-3mm}
\end{table}


\subsection{Overall Performance Analysis}

\textbf{Performance and insights.}
As shown in Table~\ref{tab:baseline_comparison}, on the \textit{Swissmetro} mode choice task, \textsc{\model} with GPT-4o-mini notably outperforms evaluated baselines across Accuracy (Acc), F1-score (F1), and AUC. Over the strongest baseline, it achieves gains of at least $6\%$ in Acc and $6.5\%$ in F1, respectively. Similar improvements are noted on the \textit{Vaccine} dataset. Notably, our proposed method exhibits higher Cross-Entropy (CE) compared to baselines such as XGBoost. We attribute this to the inherent design of \textsc{\model}, which produces more conservative probability distributions rather than extreme certainties. Specifically, unlike models that might predict a choice with $>90\%$ confidence, \textsc{\model}'s framework is less prone to such high probabilities. This characteristic may better reflect the uncertain nature of human decision-making, which our model is designed to accommodate. Overall, the performance enhancements highlight \textsc{\model}'s strength in combining symbolic structures with semantic adaptation for effective personalized decision modeling.



\textbf{Disentangling Semantically Similar Choices.} Prompt-only LLMs and classical choice models frequently fail to distinguish between superficially similar options. For example, the few-shot LLM misclassified $75\%$ of true \textit{Car} trips as the premium \textit{Metro} service. By introducing symbolic-level structure and performing individual-level semantic adaptation, \textsc{\model} more than doubled the number of correctly classified \textit{Car} trips, while maintaining high recall for both \textit{Train} and \textit{Metro}. On the Vaccine task, its learned templates encode key interactions such as age-risk trade-offs and prior-infection hesitancy, allowing it to achieve the highest F1 score despite strong semantic similarity between fully vaccinated and booster options. In practice, these interpretable templates enable a better understanding of individual behavior, for instance, identifying who tends to decline vaccination and why, which is crucial for informing high-stakes decision-making. See Appendix~\ref{cm} for details.

\textbf{Computational Complexity and Scalability.}
With $T$ and $T'$ fixed, \textsc{\model}’s runtime is linear in the number of groups $|\mathcal{G}|$ and individuals $N$, scaled by the average LLM latency $\tau$:
\[
\mathcal{O}\bigl((|\mathcal{G}|KT + N T')\,\tau_{\text{tok}}\bigr).
\]
Both stages parallelize naturally, as group-level searches run independently and individual-level refinements can be batched or distributed. 
Detailed runtime measurements are provided in Appendix~\ref{app:scalability}.

\textbf{Extended backbone LLM comparisons.} On a 100-individual subset, we also tested larger reasoning LLMs (\texttt{Qwen3-32B}, \texttt{DeepSeek-R1-Distill-Qwen-32B}, \texttt{GPT-4o}). With prompt-only baselines, larger reasoning models occasionally yield higher F1/Acc but exhibit volatile calibration (high CE), reflecting the lack of structural constraints. Under \textsc{\model}, backbone differences shrink: the symbolic discovery plus semantic adaptation turns the task into constrained sampling and small, directed improvements, allowing lightweight models to reach near-maximal performance, while stronger reasoning models provide modest, consistent gains on harder interactions (e.g., vaccine risk–trust trade-offs). Full experimental details appear in Appendix~\ref{moreexp} for completeness.



\subsection{Ablation Study}

We evaluate \textsc{\model}’s two components by \textit{(i)} keeping only the group-level symbolic utility discovery and \textit{(ii)} keeping only the individual-level semantic adaptation, under identical data and metrics. We do not include a symbolic-only group-level discovery baseline (Stage 1 without LLM), because the Concept Library is accessible only via the LLM. Excluding it would reduce the hypothesis space to symbolic operators alone, changing the problem definition rather than providing a clean ablation.



\begin{figure}[h]
  \centering
  \includegraphics[width=\linewidth]{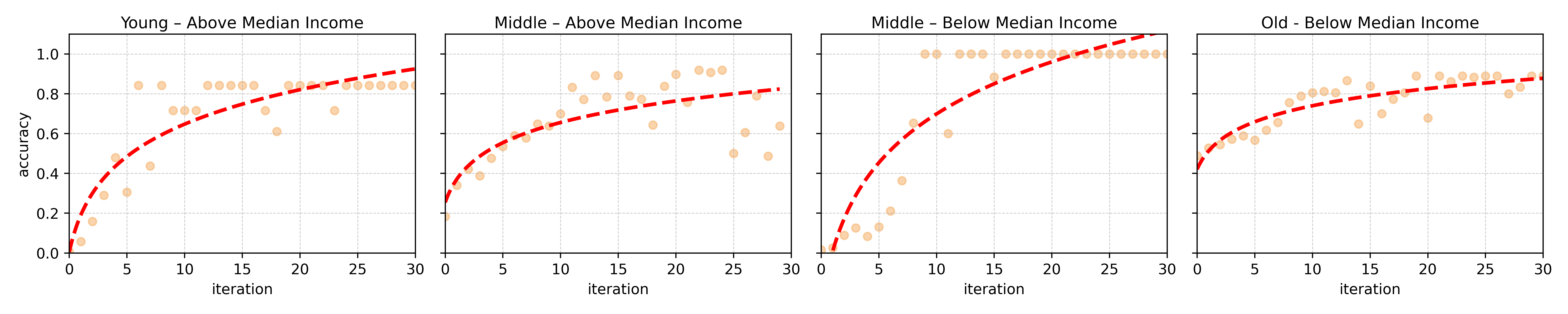}
  \caption{\textbf{Accuracy trajectories of symbolic regression.} As shown here, the accuracy keeps growing in 30 iterations for all four groups in the \textit{vaccine} dataset. Each orange dot is the average accuracy at a given iteration; the red dashed curve is a fit showing the overall upward trend and convergence.}
  \label{fig:symreg_dynamics}
\end{figure}

\textbf{Group-Level Symbolic Utility Discovery: necessary but not sufficient.}
When \textsc{\model} retains only the group-level symbolic component, accuracy exceeds the classical MNL by $4.7\%$ on \textit{Swissmetro} and $19\%$ on \textit{Vaccine} (Table \ref{tab:ablation-study}), indicating that only LLM-generated utility expressions can already encode broad demographic regularities. The accuracy trajectories of this symbolic discovery process over 30 iterations (Fig.~\ref{fig:symreg_dynamics}) further demonstrate its effectiveness, illustrating the gradual learning of these group-level trends. Nevertheless, lower F1 score and AUC and elevated cross-entropy, reflecting limited discriminative capacity for similar alternatives. These results highlight the symbolic stage’s strength in pruning the hypothesis space to interpretable structures, but also expose its limitations in capturing much heterogeneity.

\vspace{-3mm}
\begin{table}[ht]
  \centering
  \scriptsize
    \caption{Component-wise ablation results on the Swissmetro and Vaccine choice tasks, comparing Symbolic Utility Discovery only, Semantic Adaptation only, MNL, and the full \textsc{\model} pipeline.}
  \begin{adjustbox}{max width=\textwidth}
    \begin{tabular}{llcccc|cccc}
      \toprule
      & & \multicolumn{4}{c}{Swissmetro} & \multicolumn{4}{c}{Vaccine} \\
      \cmidrule(r){3-6} \cmidrule(l){7-10}
      & \textbf{Variant}
        & \textbf{Acc.$\uparrow$} & \textbf{F1.$\uparrow$} & \textbf{CE.$\downarrow$} & \textbf{AUC.$\uparrow$}
        & \textbf{Acc.$\uparrow$} & \textbf{F1.$\uparrow$} & \textbf{CE.$\downarrow$} & \textbf{AUC.$\uparrow$} \\
      \midrule
      \multirow{4}{*}{\shortstack{\textbf{Ablation}\\\textbf{Variants}}}
        & Symbolic Utility Discovery Only        & 0.6566 & 0.3785 & 2.6044 & 0.5687
                        & 0.6067 & 0.3596 & 1.0410 & 0.7294 \\
        & Semantic Adaptation Only        & 0.6044 & 0.4950 & 2.2897 & 0.6872
                        & 0.5433 & 0.5348 & 0.8695 & 0.7535 \\
        & MNL  & 0.6101 & 0.3967 & 0.8175 & 0.7182
                        & 0.4150 & 0.1955 & 1.0510 & 0.5000 \\
        & \textbf{Full Pipeline} & \textbf{0.8134} & \textbf{0.7655} & \textbf{1.0863} & \textbf{0.8825}
                        & \textbf{0.7345} & \textbf{0.7161} & \textbf{0.7551} & \textbf{0.8704} \\
      \bottomrule
    \end{tabular}
  \end{adjustbox}
  \label{tab:ablation-study}
\end{table}

\textbf{Individual-Level Semantic Adaptation: powerful only with a solid starting point.}
Conversely, bypassing symbolic discovery and initiating TextGrad from random templates leads to noteworthy degraded performance: As shown in Table \ref{tab:ablation-study}, accuracy drops to $60.4\%$ on the Swissmetro dataset and $54.3\%$ on the Vaccine dataset; Swissmetro’s CE more than doubles ($2.29$), and AUC falls below $0.70$. Without a sound starting point, gradients are likely to converge to local optima and yield erratic probability outputs, reaffirming the unreliability of unguided adaptation in multi-choice settings.

\textbf{Take-away.} The two stages of \textsc{\model} are complementary: symbolic discovery supplies an interpretable, well-regularized search space, while semantic adaptation injects the individual-level nuance that symbolic rules alone miss.

\subsection{Symbolic Utility Discovery Fragment Analysis}

\begin{figure}[htbp]
  \centering
    \begin{subfigure}[b]{0.8\textwidth}
    \centering
    \includegraphics[height=4.5cm]{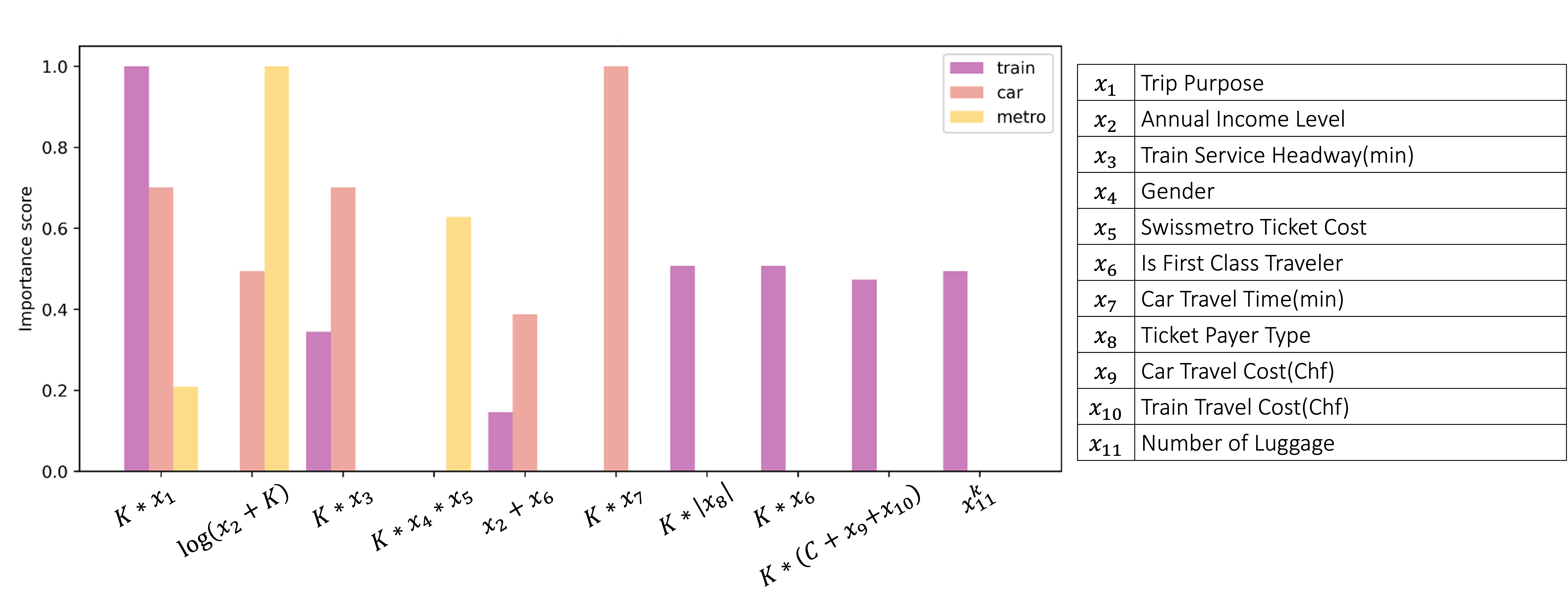}
    \caption{Swissmetro travel-mode choice dataset}
    \label{fig:sub2}
  \end{subfigure}
    \hfill
  
  \begin{subfigure}[b]{0.8\textwidth}
    \centering
    \includegraphics[height=4.5cm]{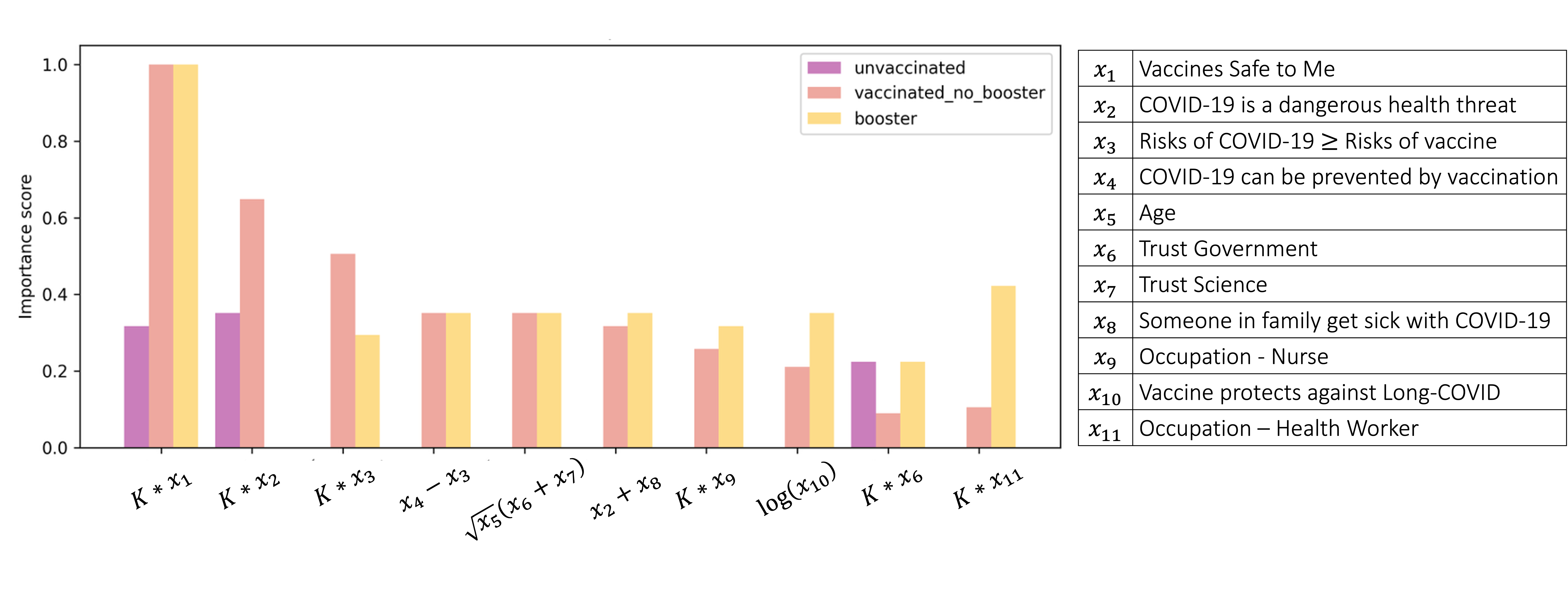}
    \vspace{-5mm} \caption{COVID-19 vaccine-choice dataset}
    \label{fig:sub1}
  \end{subfigure}
  \caption{\textbf{Aggregated fragment importance extracted from the learned symbolic utilities.} For each task, we plot the top-ranked atomic fragments $\varphi_m$ that appear in the three best group-level utility formulas and weight them by fragment importance score \ref{eq:frag_score}. Values shown here are the normalized scores in $[0,1]$.}
  \vspace{-5mm}
  \label{fig:side_by_side}
\end{figure}

Equation~\eqref{eq:pred} shows that an individual prediction is influenced by group-level symbolic utility
\(f_g^*(X_i;\theta_g^*)\).
In this section, we demonstrate the building blocks of those utilities are both behaviorally meaningful and
reusable across groups.
As shown in Fig.~\ref{fig:side_by_side}, each symbolic utility is decomposed into atomic fragments
\(\{\varphi_1,\varphi_2,\dots\}\) and their global importance is quantified.

\textbf{Fragment score.}
For every group \(g\) we retain the
top-K $(K=3)$ utilities ranked by held-out accuracy
\(\mathrm{Acc}(f)\).
The importance score of a fragment \(\varphi_m\) is then

\vspace{-2mm}



\begin{equation}
\label{eq:frag_score}
\operatorname{Score}\left(\varphi_m\right)=\sum_{g \in \mathcal{G}} \sum_{k=1}^K \nVdash\left[\varphi_m \subset\left\{f_{g, k}^*\right\}\right] \cdot \operatorname{Acc}\left(f_{g, k}^*\right)
\end{equation}

So a fragment earns points whenever it (i) appears in the
top-ranked utilities of \emph{many} groups and (ii) is embedded in
highly predictive expressions.

Fig. ~\ref{fig:side_by_side} visualizes the fragment scores for both
datasets. Only a small fraction of fragments dominate, confirming that
\textsc{\model} converges to a compact and interpretable symbolic basis. For example, in \textit{Vaccine}, one of the leading fragment $\varphi_{7}=\sqrt{\text{Age}}*\,(\text{Trust Government}+\text{Trust\_Science})$ softens age’s impact at higher values while amplifying it for individuals who trust government or science, precisely isolating the cohort most likely to take boosters.

Beyond fragment-level analysis, \textsc{\model} also produces fully interpretable symbolic utilities. Representative full formulas and domain-relevant insights for both \textit{Swissmetro} and \textit{Vaccine} are provided in Appendix \ref{symbolic_interpretability}.



\vspace{-2mm}
\section{Conclusion}
\label{conclusion}
\vspace{-2mm}

This research highlights the critical role of textual-semantic information in overcoming the limitations of traditional utility-based models for human decision-making. By introducing \textsc{\model}, an adaptive textual-symbolic and human-centric reasoning framework is proposed that integrates group-level symbolic regression of utility functions with individual-level, LLM-powered semantic modeling, we offer a more comprehensive and personalized view of choice behavior. Our experiments on transportation mode choice and vaccine uptake demonstrate that this co-design approach clearly outperforms three existing model zoos, including classical utility, machine learning, and purely LLM-based approach, underscoring the benefits of capturing both structured attributes and rich semantic context. These findings suggest that textualized-symbolic reasoning can bridge the gap between theoretical utility optimization and real-world individual choices, paving the way for more adaptive and human-centric decision models.


\textbf{Limitations.} The current implementation of  \textsc{\model} has two limitations. 1) Computational Complexity: The proposed framework requires extra computational resources for textual gradient, particularly when scaling to larger populations. 2) Representation on Groups: \textsc{\model} assumes that a shared symbolic utility function can effectively model each demographic group. However, groups with greater internal diversity may produce weaker or less reliable representations. 3) Result Stability: All reported results are based on single representative runs under fixed random seeds, given the computational cost of multi-stage adaptation. Future work will include multi-seed repetitions to further examine the stability of \textsc{\model}'s performance. 

\section*{Acknowledgments}

Yang Zhao acknowledges a fellowship from JHU + Amazon Initiative for Interactive AI.

\bibliographystyle{unsrtnat} 
\bibliography{references}

\begin{thebibliography}{130}
\providecommand{\natexlab}[1]{#1}
\providecommand{\url}[1]{\texttt{#1}}
\expandafter\ifx\csname urlstyle\endcsname\relax
  \providecommand{\doi}[1]{doi: #1}\else
  \providecommand{\doi}{doi: \begingroup \urlstyle{rm}\Url}\fi

\bibitem[Fu et~al.(2011)Fu, Rosenbloom, Wang, and Nowak]{fu2011imitation}
Feng Fu, Daniel~I Rosenbloom, Long Wang, and Martin~A Nowak.
\newblock Imitation dynamics of vaccination behaviour on social networks.
\newblock \emph{Proceedings of the Royal Society B: Biological Sciences}, 278\penalty0 (1702):\penalty0 42--49, 2011.

\bibitem[Bond and Nolan(2011)]{bond2011making}
Lyndal Bond and Terry Nolan.
\newblock Making sense of perceptions of risk of diseases and vaccinations: a qualitative study combining models of health beliefs, decision-making and risk perception.
\newblock \emph{BMC public health}, 11:\penalty0 1--14, 2011.

\bibitem[Betsch et~al.(2017)Betsch, B{\"o}hm, Korn, and Holtmann]{betsch2017benefits}
Cornelia Betsch, Robert B{\"o}hm, Lars Korn, and Cindy Holtmann.
\newblock On the benefits of explaining herd immunity in vaccine advocacy.
\newblock \emph{Nature human behaviour}, 1\penalty0 (3):\penalty0 0056, 2017.

\bibitem[Von~Neumann and Morgenstern(2007)]{von2007theory}
John Von~Neumann and Oskar Morgenstern.
\newblock Theory of games and economic behavior: 60th anniversary commemorative edition.
\newblock In \emph{Theory of games and economic behavior}. Princeton university press, 2007.

\bibitem[Rabin(2013)]{rabin2013risk}
Matihew Rabin.
\newblock Risk aversion and expected-utility theory: A calibration theorem.
\newblock In \emph{Handbook of the fundamentals of financial decision making: Part I}, pages 241--252. World Scientific, 2013.

\bibitem[Quiggin(2012)]{quiggin2012generalized}
John Quiggin.
\newblock \emph{Generalized expected utility theory: The rank-dependent model}.
\newblock Springer Science \& Business Media, 2012.

\bibitem[Ben-Akiva and Lerman(1985)]{ben1985discrete}
Moshe~E Ben-Akiva and Steven~R Lerman.
\newblock \emph{Discrete choice analysis: theory and application to travel demand}, volume~9.
\newblock MIT press, 1985.

\bibitem[Small and Rosen(1981)]{small1981applied}
Kenneth~A Small and Harvey~S Rosen.
\newblock Applied welfare economics with discrete choice models.
\newblock \emph{Econometrica: Journal of the Econometric Society}, pages 105--130, 1981.

\bibitem[Haghani et~al.(2021)Haghani, Bliemer, and Hensher]{haghani2021landscape}
Milad Haghani, Michiel~CJ Bliemer, and David~A Hensher.
\newblock The landscape of econometric discrete choice modelling research.
\newblock \emph{Journal of choice modelling}, 40:\penalty0 100303, 2021.

\bibitem[Berbeglia et~al.(2022)Berbeglia, Garassino, and Vulcano]{berbeglia2022comparative}
Gerardo Berbeglia, Agust{\'\i}n Garassino, and Gustavo Vulcano.
\newblock A comparative empirical study of discrete choice models in retail operations.
\newblock \emph{Management Science}, 68\penalty0 (6):\penalty0 4005--4023, 2022.

\bibitem[Zheng et~al.(2021)Zheng, Wang, and Zhao]{zheng2021equality}
Yunhan Zheng, Shenhao Wang, and Jinhua Zhao.
\newblock Equality of opportunity in travel behavior prediction with deep neural networks and discrete choice models.
\newblock \emph{Transportation Research Part C: Emerging Technologies}, 132:\penalty0 103410, 2021.

\bibitem[Salas et~al.(2022)Salas, De~la Fuente, Astroza, and Carrasco]{salas2022systematic}
Patricio Salas, Rodrigo De~la Fuente, Sebastian Astroza, and Juan~Antonio Carrasco.
\newblock A systematic comparative evaluation of machine learning classifiers and discrete choice models for travel mode choice in the presence of response heterogeneity.
\newblock \emph{Expert Systems with Applications}, 193:\penalty0 116253, 2022.

\bibitem[Ameli et~al.(2022)Ameli, Faradonbeh, Lebacque, Abouee-Mehrizi, and Leclercq]{ameli2022departure}
Mostafa Ameli, Mohamad Sadegh~Shirani Faradonbeh, Jean-Patrick Lebacque, Hossein Abouee-Mehrizi, and Ludovic Leclercq.
\newblock Departure time choice models in urban transportation systems based on mean field games.
\newblock \emph{Transportation Science}, 56\penalty0 (6):\penalty0 1483--1504, 2022.

\bibitem[Bier et~al.(2020)Bier, Zhou, and Du]{bier2020game}
Vicki~M Bier, Yuqun Zhou, and Hongru Du.
\newblock Game-theoretic modeling of pre-disaster relocation.
\newblock \emph{The Engineering Economist}, 65\penalty0 (2):\penalty0 89--113, 2020.

\bibitem[Li et~al.(2022)Li, Wang, and Xie]{li2022government}
Lixu Li, Zhiqiang Wang, and Xiaoqing Xie.
\newblock From government to market? a discrete choice analysis of policy instruments for electric vehicle adoption.
\newblock \emph{Transportation Research Part A: Policy and Practice}, 160:\penalty0 143--159, 2022.

\bibitem[Lor{\'\i}a-Rebolledo et~al.(2024)Lor{\'\i}a-Rebolledo, Abbott, Antunes, Norwood, Ryan, Watson, and Wu]{loria2024public}
Luis~Enrique Lor{\'\i}a-Rebolledo, Michael Abbott, M{\'e}lanie Antunes, Patricia Norwood, Mandy Ryan, Verity Watson, and Hangjian Wu.
\newblock Public preferences and willingness to pay for a net zero nhs: a protocol for a discrete choice experiment in england and scotland.
\newblock \emph{BMJ open}, 14\penalty0 (6):\penalty0 e082863, 2024.

\bibitem[Thomas et~al.(2024)Thomas, Straub, Tatai, Shene, Tosik, Kersting, and Rothkopf]{thomas2024modelling}
Tobias Thomas, Dominik Straub, Fabian Tatai, Megan Shene, T{\"u}mer Tosik, Kristian Kersting, and Constantin~A Rothkopf.
\newblock Modelling dataset bias in machine-learned theories of economic decision-making.
\newblock \emph{Nature Human Behaviour}, 8\penalty0 (4):\penalty0 679--691, 2024.

\bibitem[Glimcher(2022)]{glimcher2022efficiently}
Paul~W Glimcher.
\newblock Efficiently irrational: deciphering the riddle of human choice.
\newblock \emph{Trends in cognitive sciences}, 26\penalty0 (8):\penalty0 669--687, 2022.

\bibitem[Schoemaker(1982)]{schoemaker1982expected}
Paul~JH Schoemaker.
\newblock The expected utility model: Its variants, purposes, evidence and limitations.
\newblock \emph{Journal of economic literature}, pages 529--563, 1982.

\bibitem[Achiam et~al.(2023)Achiam, Adler, Agarwal, Ahmad, Akkaya, Aleman, Almeida, Altenschmidt, Altman, Anadkat, et~al.]{achiam2023gpt}
Josh Achiam, Steven Adler, Sandhini Agarwal, Lama Ahmad, Ilge Akkaya, Florencia~Leoni Aleman, Diogo Almeida, Janko Altenschmidt, Sam Altman, Shyamal Anadkat, et~al.
\newblock Gpt-4 technical report.
\newblock \emph{arXiv preprint arXiv:2303.08774}, 2023.

\bibitem[Romera-Paredes et~al.(2024)Romera-Paredes, Barekatain, Novikov, Balog, Kumar, Dupont, Ruiz, Ellenberg, Wang, Fawzi, et~al.]{romera2024mathematical}
Bernardino Romera-Paredes, Mohammadamin Barekatain, Alexander Novikov, Matej Balog, M~Pawan Kumar, Emilien Dupont, Francisco~JR Ruiz, Jordan~S Ellenberg, Pengming Wang, Omar Fawzi, et~al.
\newblock Mathematical discoveries from program search with large language models.
\newblock \emph{Nature}, 625\penalty0 (7995):\penalty0 468--475, 2024.

\bibitem[Shojaee et~al.(2024)Shojaee, Meidani, Gupta, Barati~Farimani, and Reddy]{llmsr}
Parshin Shojaee, Kazem Meidani, Shashank Gupta, Amir Barati~Farimani, and Chandan Reddy.
\newblock Llm-sr: Scientific equation discovery via programming with large language models, 04 2024.

\bibitem[Grayeli et~al.(2024{\natexlab{a}})Grayeli, Sehgal, Costilla~Reyes, Cranmer, and Chaudhuri]{grayeli2024symbolic}
Arya Grayeli, Atharva Sehgal, Omar Costilla~Reyes, Miles Cranmer, and Swarat Chaudhuri.
\newblock Symbolic regression with a learned concept library.
\newblock \emph{Advances in Neural Information Processing Systems}, 37:\penalty0 44678--44709, 2024{\natexlab{a}}.

\bibitem[Aher et~al.(2023)Aher, Arriaga, and Kalai]{aher2023using}
Gati~V Aher, Rosa~I Arriaga, and Adam~Tauman Kalai.
\newblock Using large language models to simulate multiple humans and replicate human subject studies.
\newblock In \emph{International Conference on Machine Learning}, pages 337--371. PMLR, 2023.

\bibitem[Schramowski et~al.(2022)Schramowski, Turan, Andersen, Rothkopf, and Kersting]{schramowski2022large}
Patrick Schramowski, Cigdem Turan, Nico Andersen, Constantin~A Rothkopf, and Kristian Kersting.
\newblock Large pre-trained language models contain human-like biases of what is right and wrong to do.
\newblock \emph{Nature Machine Intelligence}, 4\penalty0 (3):\penalty0 258--268, 2022.

\bibitem[Hagendorff et~al.(2023)Hagendorff, Fabi, and Kosinski]{hagendorff2023human}
Thilo Hagendorff, Sarah Fabi, and Michal Kosinski.
\newblock Human-like intuitive behavior and reasoning biases emerged in large language models but disappeared in chatgpt.
\newblock \emph{Nature Computational Science}, 3\penalty0 (10):\penalty0 833--838, 2023.

\bibitem[Ramos et~al.(2014)Ramos, Daamen, and and]{Ramos02012014}
Giselle~Moraes Ramos, Winnie Daamen, and Serge~Hoogendoorn and.
\newblock A state-of-the-art review: Developments in utility theory, prospect theory and regret theory to investigate travellers' behaviour in situations involving travel time uncertainty.
\newblock \emph{Transport Reviews}, 34\penalty0 (1):\penalty0 46--67, 2014.
\newblock \doi{10.1080/01441647.2013.856356}.
\newblock URL \url{https://doi.org/10.1080/01441647.2013.856356}.

\bibitem[Al-Salih and Esztergár-Kiss(2021)]{su13084332}
Wissam~Qassim Al-Salih and Domokos Esztergár-Kiss.
\newblock Linking mode choice with travel behavior by using logit model based on utility function.
\newblock \emph{Sustainability}, 13\penalty0 (8), 2021.
\newblock ISSN 2071-1050.
\newblock \doi{10.3390/su13084332}.
\newblock URL \url{https://www.mdpi.com/2071-1050/13/8/4332}.

\bibitem[Vos et~al.(2016)Vos, Mokhtarian, Schwanen, Acker, and Witlox]{DeVos2016}
Jonas~De Vos, Patricia~L. Mokhtarian, Tim Schwanen, Veronique~Van Acker, and Frank Witlox.
\newblock Travel mode choice and travel satisfaction: bridging the gap between decision utility and experienced utility.
\newblock \emph{Transportation}, 43\penalty0 (5):\penalty0 771--796, 2016.
\newblock ISSN 1572-9435.
\newblock \doi{10.1007/s11116-015-9619-9}.
\newblock URL \url{https://doi.org/10.1007/s11116-015-9619-9}.

\bibitem[Ettema et~al.(2016)Ettema, Friman, G{\"a}rling, and Olsson]{Ettema2016}
Dick Ettema, Margareta Friman, Tommy G{\"a}rling, and Lars~E. Olsson.
\newblock \emph{Travel Mode Use, Travel Mode Shift and Subjective Well-Being: Overview of Theories, Empirical Findings and Policy Implications}, pages 129--150.
\newblock Springer Berlin Heidelberg, Berlin, Heidelberg, 2016.
\newblock ISBN 978-3-662-48184-4.
\newblock \doi{10.1007/978-3-662-48184-4_7}.
\newblock URL \url{https://doi.org/10.1007/978-3-662-48184-4_7}.

\bibitem[Vij and Walker(2024)]{vij2024hybrid}
Akshay Vij and Joan~L Walker.
\newblock Hybrid choice models: The identification problem.
\newblock In \emph{Handbook of choice modelling}, pages 522--567. Edward Elgar Publishing, 2024.

\bibitem[Lancsar et~al.(2017)Lancsar, Fiebig, and Hole]{lancsar2017discrete}
Emily Lancsar, Denzil~G Fiebig, and Arne~Risa Hole.
\newblock Discrete choice experiments: a guide to model specification, estimation and software.
\newblock \emph{Pharmacoeconomics}, 35:\penalty0 697--716, 2017.

\bibitem[Wang(2025)]{wang2025rational}
Zexuan Wang.
\newblock Rational inattention: The interplay of stakes and prior beliefs in a laboratory study.
\newblock \emph{Available at SSRN 5140893}, 2025.

\bibitem[Yu and Sun(2012)]{yu2012four}
Lijun Yu and Bin Sun.
\newblock Four types of typical discrete choice models: Which are you using?
\newblock In \emph{Proceedings of 2012 IEEE International Conference on Service Operations and Logistics, and Informatics}, pages 298--301. IEEE, 2012.

\bibitem[Habib(2023)]{habib2023rational}
Khandker~Nurul Habib.
\newblock Rational inattention in discrete choice models: Estimable specifications of ri-multinomial logit (ri-mnl) and ri-nested logit (ri-nl) models.
\newblock \emph{Transportation Research Part B: Methodological}, 172:\penalty0 53--70, 2023.

\bibitem[Deneke et~al.(2024)Deneke, Desta, Afework, T{\'o}th, et~al.]{deneke2024transportation}
Yeshitila Deneke, Robel Desta, Anteneh Afework, J{\'a}nos T{\'o}th, et~al.
\newblock Transportation mode choice behavior with multinomial logit model: work and school trips.
\newblock \emph{Transactions on transport sciences}, 15\penalty0 (1):\penalty0 17--27, 2024.

\bibitem[Rahmani et~al.(2023)Rahmani, Baghbani, Bouguila, and Patterson]{10077454}
Saeed Rahmani, Asiye Baghbani, Nizar Bouguila, and Zachary Patterson.
\newblock Graph neural networks for intelligent transportation systems: A survey.
\newblock \emph{IEEE Transactions on Intelligent Transportation Systems}, 24\penalty0 (8):\penalty0 8846--8885, 2023.
\newblock \doi{10.1109/TITS.2023.3257759}.

\bibitem[Zhang et~al.(2023{\natexlab{a}})Zhang, Zhang, Liu, and Zhang]{zhang2023understanding}
Hui Zhang, Li~Zhang, Yanjun Liu, and Lele Zhang.
\newblock Understanding travel mode choice behavior: Influencing factors analysis and prediction with machine learning method.
\newblock \emph{Sustainability}, 15\penalty0 (14):\penalty0 11414, 2023{\natexlab{a}}.

\bibitem[Pineda-Jaramillo and Arbel{\'a}ez-Arenas(2022)]{pineda2022assessing}
Juan Pineda-Jaramillo and {\'O}scar Arbel{\'a}ez-Arenas.
\newblock Assessing the performance of gradient-boosting models for predicting the travel mode choice using household survey data.
\newblock \emph{Journal of Urban Planning and Development}, 148\penalty0 (2):\penalty0 04022007, 2022.

\bibitem[Kim(2021)]{kim2021analysis}
Eui-Jin Kim.
\newblock Analysis of travel mode choice in seoul using an interpretable machine learning approach.
\newblock \emph{Journal of Advanced Transportation}, 2021\penalty0 (1):\penalty0 6685004, 2021.

\bibitem[Mart{\'\i}n-Baos et~al.(2023)Mart{\'\i}n-Baos, L{\'o}pez-G{\'o}mez, Rodriguez-Benitez, Hillel, and Garc{\'\i}a-R{\'o}denas]{martin2023prediction}
Jos{\'e}~{\'A}ngel Mart{\'\i}n-Baos, Julio~Alberto L{\'o}pez-G{\'o}mez, Luis Rodriguez-Benitez, Tim Hillel, and Ricardo Garc{\'\i}a-R{\'o}denas.
\newblock A prediction and behavioural analysis of machine learning methods for modelling travel mode choice.
\newblock \emph{Transportation research part C: emerging technologies}, 156:\penalty0 104318, 2023.

\bibitem[Zhang et~al.(2023{\natexlab{b}})Zhang, Zhang, Liu, and Zhang]{su151411414}
Hui Zhang, Li~Zhang, Yanjun Liu, and Lele Zhang.
\newblock Understanding travel mode choice behavior: Influencing factors analysis and prediction with machine learning method.
\newblock \emph{Sustainability}, 15\penalty0 (14), 2023{\natexlab{b}}.
\newblock ISSN 2071-1050.
\newblock \doi{10.3390/su151411414}.
\newblock URL \url{https://www.mdpi.com/2071-1050/15/14/11414}.

\bibitem[Tang et~al.(2024)Tang, Tang, Fu, and Ma]{tang2024predicting}
Li~Tang, Chuanli Tang, Qi~Fu, and Changxi Ma.
\newblock Predicting travel mode choice with a robust neural network and shapley additive explanations analysis.
\newblock \emph{IET Intelligent Transport Systems}, 18\penalty0 (7):\penalty0 1339--1354, 2024.

\bibitem[Koushik et~al.(2025)Koushik, Manoj, and Nezamuddin]{koushik2025explaining}
Anil Koushik, M~Manoj, and N~Nezamuddin.
\newblock Explaining deep learning-based activity schedule models using shapley additive explanations.
\newblock \emph{Transportation Letters}, 17\penalty0 (3):\penalty0 442--457, 2025.

\bibitem[Xu et~al.(2023)Xu, Zhong, Kang, Duan, Wang, Lu, and Shi]{9954273}
Aikun Xu, Ping Zhong, Yilin Kang, Jiongqiang Duan, Anning Wang, Mingming Lu, and Chuan Shi.
\newblock Than: Multimodal transportation recommendation with heterogeneous graph attention networks.
\newblock \emph{IEEE Transactions on Intelligent Transportation Systems}, 24\penalty0 (2):\penalty0 1533--1543, 2023.
\newblock \doi{10.1109/TITS.2022.3221370}.

\bibitem[Çevik et~al.(2024)Çevik, Přibyl, and Samandar]{SHAPdeep2024}
Halil Çevik, Ondřej Přibyl, and Shoaib Samandar.
\newblock Understanding travel behavior: A deep neural network and shap approach to mode choice determinants.
\newblock \emph{Neural Network World}, 34:\penalty0 219--241, 01 2024.
\newblock \doi{10.14311/NNW.2024.34.012}.

\bibitem[Sadeghi et~al.(2024)Sadeghi, Alizadehsani, CIFCI, Kausar, Rehman, Mahanta, Bora, Almasri, Alkhawaldeh, Hussain, Alatas, Shoeibi, Moosaei, Hladík, Nahavandi, and Pardalos]{SADEGHI2024109370}
Zahra Sadeghi, Roohallah Alizadehsani, Mehmet~Akif CIFCI, Samina Kausar, Rizwan Rehman, Priyakshi Mahanta, Pranjal~Kumar Bora, Ammar Almasri, Rami~S. Alkhawaldeh, Sadiq Hussain, Bilal Alatas, Afshin Shoeibi, Hossein Moosaei, Milan Hladík, Saeid Nahavandi, and Panos~M. Pardalos.
\newblock A review of explainable artificial intelligence in healthcare.
\newblock \emph{Computers and Electrical Engineering}, 118:\penalty0 109370, 2024.
\newblock ISSN 0045-7906.
\newblock \doi{https://doi.org/10.1016/j.compeleceng.2024.109370}.
\newblock URL \url{https://www.sciencedirect.com/science/article/pii/S0045790624002982}.

\bibitem[Linardatos et~al.(2021)Linardatos, Papastefanopoulos, and Kotsiantis]{e23010018}
Pantelis Linardatos, Vasilis Papastefanopoulos, and Sotiris Kotsiantis.
\newblock Explainable ai: A review of machine learning interpretability methods.
\newblock \emph{Entropy}, 23\penalty0 (1), 2021.
\newblock ISSN 1099-4300.
\newblock \doi{10.3390/e23010018}.
\newblock URL \url{https://www.mdpi.com/1099-4300/23/1/18}.

\bibitem[Qi et~al.(2019)Qi, Khorram, and Li]{unknown}
Zhongang Qi, Saeed Khorram, and Fuxin Li.
\newblock Visualizing deep networks by optimizing with integrated gradients, 05 2019.

\bibitem[Walker et~al.(2024)Walker, Chen, and Ewetz]{IntegratedDecision}
Chase Walker, Kenny Chen, and Rickard Ewetz.
\newblock Integrated decision gradients: Compute your attributions where the model makes its decision.
\newblock \emph{Proceedings of the AAAI Conference on Artificial Intelligence}, 38:\penalty0 5289--5297, 03 2024.
\newblock \doi{10.1609/aaai.v38i6.28336}.

\bibitem[Augusto and Barbosa(2000)]{augusto2000symbolic}
Douglas~Adriano Augusto and Helio~JC Barbosa.
\newblock Symbolic regression via genetic programming.
\newblock In \emph{Proceedings. Vol. 1. Sixth Brazilian symposium on neural networks}, pages 173--178. IEEE, 2000.

\bibitem[Zhong et~al.(2018)Zhong, Feng, Cai, and Ong]{zhong2018multifactorial}
Jinghui Zhong, Liang Feng, Wentong Cai, and Yew-Soon Ong.
\newblock Multifactorial genetic programming for symbolic regression problems.
\newblock \emph{IEEE transactions on systems, man, and cybernetics: systems}, 50\penalty0 (11):\penalty0 4492--4505, 2018.

\bibitem[Skanderova(2023)]{skanderova2023self}
Lenka Skanderova.
\newblock Self-organizing migrating algorithm: review, improvements and comparison.
\newblock \emph{Artificial Intelligence Review}, 56\penalty0 (1):\penalty0 101--172, 2023.

\bibitem[Cranmer(2023)]{pysr}
Miles Cranmer.
\newblock Interpretable machine learning for science with pysr and symbolicregression.jl, 05 2023.

\bibitem[Real et~al.(2019)Real, Aggarwal, Huang, and Le]{10.1609/aaai.v33i01.33014780}
Esteban Real, Alok Aggarwal, Yanping Huang, and Quoc~V. Le.
\newblock Regularized evolution for image classifier architecture search.
\newblock In \emph{Proceedings of the Thirty-Third AAAI Conference on Artificial Intelligence and Thirty-First Innovative Applications of Artificial Intelligence Conference and Ninth AAAI Symposium on Educational Advances in Artificial Intelligence}, AAAI'19/IAAI'19/EAAI'19. AAAI Press, 2019.
\newblock ISBN 978-1-57735-809-1.
\newblock \doi{10.1609/aaai.v33i01.33014780}.
\newblock URL \url{https://doi.org/10.1609/aaai.v33i01.33014780}.

\bibitem[Makke and Chawla(2024)]{10.1007/s10462-023-10622-0}
Nour Makke and Sanjay Chawla.
\newblock Interpretable scientific discovery with symbolic regression: a review.
\newblock \emph{Artif. Intell. Rev.}, 57\penalty0 (1), January 2024.
\newblock ISSN 0269-2821.
\newblock \doi{10.1007/s10462-023-10622-0}.
\newblock URL \url{https://doi.org/10.1007/s10462-023-10622-0}.

\bibitem[Ye et~al.(2024)Ye, Wang, Cao, Berto, Hua, Kim, Park, and Song]{ye2024reevo}
Haoran Ye, Jiarui Wang, Zhiguang Cao, Federico Berto, Chuanbo Hua, Haeyeon Kim, Jinkyoo Park, and Guojie Song.
\newblock Reevo: Large language models as hyper-heuristics with reflective evolution.
\newblock In \emph{The Thirty-eighth Annual Conference on Neural Information Processing Systems}, 2024.
\newblock URL \url{https://openreview.net/forum?id=483IPG0HWL}.

\bibitem[Grayeli et~al.(2024{\natexlab{b}})Grayeli, Sehgal, Costilla-Reyes, Cranmer, and Chaudhuri]{NEURIPS2024_4ec3ddc4}
Arya Grayeli, Atharva Sehgal, Omar Costilla-Reyes, Miles Cranmer, and Swarat Chaudhuri.
\newblock Symbolic regression with a learned concept library.
\newblock In A.~Globerson, L.~Mackey, D.~Belgrave, A.~Fan, U.~Paquet, J.~Tomczak, and C.~Zhang, editors, \emph{Advances in Neural Information Processing Systems}, volume~37, pages 44678--44709. Curran Associates, Inc., 2024{\natexlab{b}}.
\newblock URL \url{https://proceedings.neurips.cc/paper_files/paper/2024/file/4ec3ddc465c6d650c9c419fb91f1c00a-Paper-Conference.pdf}.

\bibitem[Mall et~al.(2025)Mall, Phoo, Chiquier, Hariharan, Bala, and Vondrick]{disciple-25}
Utkarsh Mall, Cheng~Perng Phoo, Mia Chiquier, Bharath Hariharan, Kavita Bala, and Carl Vondrick.
\newblock Disciple: Learning interpretable programs for scientific visual discovery.
\newblock 2025.

\bibitem[Guo et~al.(2025)Guo, Yang, Zhang, Song, Zhang, Xu, Zhu, Ma, Wang, Bi, et~al.]{guo2025deepseek}
Daya Guo, Dejian Yang, Haowei Zhang, Junxiao Song, Ruoyu Zhang, Runxin Xu, Qihao Zhu, Shirong Ma, Peiyi Wang, Xiao Bi, et~al.
\newblock Deepseek-r1: Incentivizing reasoning capability in llms via reinforcement learning.
\newblock \emph{arXiv preprint arXiv:2501.12948}, 2025.

\bibitem[Grattafiori et~al.(2024)Grattafiori, Dubey, Jauhri, Pandey, Kadian, Al-Dahle, Letman, Mathur, Schelten, Vaughan, Yang, Fan, Goyal, Hartshorn, Yang, Mitra, Sravankumar, Korenev, Hinsvark, Rao, Zhang, Rodriguez, Gregerson, Spataru, Roziere, Biron, Tang, Chern, Caucheteux, Nayak, Bi, Marra, McConnell, Keller, Touret, Wu, Wong, Ferrer, Nikolaidis, Allonsius, Song, Pintz, Livshits, Wyatt, Esiobu, Choudhary, Mahajan, Garcia-Olano, Perino, Hupkes, Lakomkin, AlBadawy, Lobanova, Dinan, Smith, Radenovic, Guzmán, Zhang, Synnaeve, Lee, Anderson, Thattai, Nail, Mialon, Pang, Cucurell, Nguyen, Korevaar, Xu, Touvron, Zarov, Ibarra, Kloumann, Misra, Evtimov, Zhang, Copet, Lee, Geffert, Vranes, Park, Mahadeokar, Shah, van~der Linde, Billock, Hong, Lee, Fu, Chi, Huang, Liu, Wang, Yu, Bitton, Spisak, Park, Rocca, Johnstun, Saxe, Jia, Alwala, Prasad, Upasani, Plawiak, Li, Heafield, Stone, El-Arini, Iyer, Malik, Chiu, Bhalla, Lakhotia, Rantala-Yeary, van~der Maaten, Chen, Tan, Jenkins, Martin, Madaan, Malo, Blecher,
  Landzaat, de~Oliveira, Muzzi, Pasupuleti, Singh, Paluri, Kardas, Tsimpoukelli, Oldham, Rita, Pavlova, Kambadur, Lewis, Si, Singh, Hassan, Goyal, Torabi, Bashlykov, Bogoychev, Chatterji, Zhang, Duchenne, Çelebi, Alrassy, Zhang, Li, Vasic, Weng, Bhargava, Dubal, Krishnan, Koura, Xu, He, Dong, Srinivasan, Ganapathy, Calderer, Cabral, Stojnic, Raileanu, Maheswari, Girdhar, Patel, Sauvestre, Polidoro, Sumbaly, Taylor, Silva, Hou, Wang, Hosseini, Chennabasappa, Singh, Bell, Kim, Edunov, Nie, Narang, Raparthy, Shen, Wan, Bhosale, Zhang, Vandenhende, Batra, Whitman, Sootla, Collot, Gururangan, Borodinsky, Herman, Fowler, Sheasha, Georgiou, Scialom, Speckbacher, Mihaylov, Xiao, Karn, Goswami, Gupta, Ramanathan, Kerkez, Gonguet, Do, Vogeti, Albiero, Petrovic, Chu, Xiong, Fu, Meers, Martinet, Wang, Wang, Tan, Xia, Xie, Jia, Wang, Goldschlag, Gaur, Babaei, Wen, Song, Zhang, Li, Mao, Coudert, Yan, Chen, Papakipos, Singh, Srivastava, Jain, Kelsey, Shajnfeld, Gangidi, Victoria, Goldstand, Menon, Sharma, Boesenberg,
  Baevski, Feinstein, Kallet, Sangani, Teo, Yunus, Lupu, Alvarado, Caples, Gu, Ho, Poulton, Ryan, Ramchandani, Dong, Franco, Goyal, Saraf, Chowdhury, Gabriel, Bharambe, Eisenman, Yazdan, James, Maurer, Leonhardi, Huang, Loyd, Paola, Paranjape, Liu, Wu, Ni, Hancock, Wasti, Spence, Stojkovic, Gamido, Montalvo, Parker, Burton, Mejia, Liu, Wang, Kim, Zhou, Hu, Chu, Cai, Tindal, Feichtenhofer, Gao, Civin, Beaty, Kreymer, Li, Adkins, Xu, Testuggine, David, Parikh, Liskovich, Foss, Wang, Le, Holland, Dowling, Jamil, Montgomery, Presani, Hahn, Wood, Le, Brinkman, Arcaute, Dunbar, Smothers, Sun, Kreuk, Tian, Kokkinos, Ozgenel, Caggioni, Kanayet, Seide, Florez, Schwarz, Badeer, Swee, Halpern, Herman, Sizov, Guangyi, Zhang, Lakshminarayanan, Inan, Shojanazeri, Zou, Wang, Zha, Habeeb, Rudolph, Suk, Aspegren, Goldman, Zhan, Damlaj, Molybog, Tufanov, Leontiadis, Veliche, Gat, Weissman, Geboski, Kohli, Lam, Asher, Gaya, Marcus, Tang, Chan, Zhen, Reizenstein, Teboul, Zhong, Jin, Yang, Cummings, Carvill, Shepard, McPhie,
  Torres, Ginsburg, Wang, Wu, U, Saxena, Khandelwal, Zand, Matosich, Veeraraghavan, Michelena, Li, Jagadeesh, Huang, Chawla, Huang, Chen, Garg, A, Silva, Bell, Zhang, Guo, Yu, Moshkovich, Wehrstedt, Khabsa, Avalani, Bhatt, Mankus, Hasson, Lennie, Reso, Groshev, Naumov, Lathi, Keneally, Liu, Seltzer, Valko, Restrepo, Patel, Vyatskov, Samvelyan, Clark, Macey, Wang, Hermoso, Metanat, Rastegari, Bansal, Santhanam, Parks, White, Bawa, Singhal, Egebo, Usunier, Mehta, Laptev, Dong, Cheng, Chernoguz, Hart, Salpekar, Kalinli, Kent, Parekh, Saab, Balaji, Rittner, Bontrager, Roux, Dollar, Zvyagina, Ratanchandani, Yuvraj, Liang, Alao, Rodriguez, Ayub, Murthy, Nayani, Mitra, Parthasarathy, Li, Hogan, Battey, Wang, Howes, Rinott, Mehta, Siby, Bondu, Datta, Chugh, Hunt, Dhillon, Sidorov, Pan, Mahajan, Verma, Yamamoto, Ramaswamy, Lindsay, Lindsay, Feng, Lin, Zha, Patil, Shankar, Zhang, Zhang, Wang, Agarwal, Sajuyigbe, Chintala, Max, Chen, Kehoe, Satterfield, Govindaprasad, Gupta, Deng, Cho, Virk, Subramanian, Choudhury,
  Goldman, Remez, Glaser, Best, Koehler, Robinson, Li, Zhang, Matthews, Chou, Shaked, Vontimitta, Ajayi, Montanez, Mohan, Kumar, Mangla, Ionescu, Poenaru, Mihailescu, Ivanov, Li, Wang, Jiang, Bouaziz, Constable, Tang, Wu, Wang, Wu, Gao, Kleinman, Chen, Hu, Jia, Qi, Li, Zhang, Zhang, Adi, Nam, Yu, Wang, Zhao, Hao, Qian, Li, He, Rait, DeVito, Rosnbrick, Wen, Yang, Zhao, and Ma]{grattafiori2024llama3herdmodels}
Aaron Grattafiori, Abhimanyu Dubey, Abhinav Jauhri, Abhinav Pandey, Abhishek Kadian, Ahmad Al-Dahle, Aiesha Letman, Akhil Mathur, Alan Schelten, Alex Vaughan, Amy Yang, Angela Fan, Anirudh Goyal, Anthony Hartshorn, Aobo Yang, Archi Mitra, Archie Sravankumar, Artem Korenev, Arthur Hinsvark, Arun Rao, Aston Zhang, Aurelien Rodriguez, Austen Gregerson, Ava Spataru, Baptiste Roziere, Bethany Biron, Binh Tang, Bobbie Chern, Charlotte Caucheteux, Chaya Nayak, Chloe Bi, Chris Marra, Chris McConnell, Christian Keller, Christophe Touret, Chunyang Wu, Corinne Wong, Cristian~Canton Ferrer, Cyrus Nikolaidis, Damien Allonsius, Daniel Song, Danielle Pintz, Danny Livshits, Danny Wyatt, David Esiobu, Dhruv Choudhary, Dhruv Mahajan, Diego Garcia-Olano, Diego Perino, Dieuwke Hupkes, Egor Lakomkin, Ehab AlBadawy, Elina Lobanova, Emily Dinan, Eric~Michael Smith, Filip Radenovic, Francisco Guzmán, Frank Zhang, Gabriel Synnaeve, Gabrielle Lee, Georgia~Lewis Anderson, Govind Thattai, Graeme Nail, Gregoire Mialon, Guan Pang,
  Guillem Cucurell, Hailey Nguyen, Hannah Korevaar, Hu~Xu, Hugo Touvron, Iliyan Zarov, Imanol~Arrieta Ibarra, Isabel Kloumann, Ishan Misra, Ivan Evtimov, Jack Zhang, Jade Copet, Jaewon Lee, Jan Geffert, Jana Vranes, Jason Park, Jay Mahadeokar, Jeet Shah, Jelmer van~der Linde, Jennifer Billock, Jenny Hong, Jenya Lee, Jeremy Fu, Jianfeng Chi, Jianyu Huang, Jiawen Liu, Jie Wang, Jiecao Yu, Joanna Bitton, Joe Spisak, Jongsoo Park, Joseph Rocca, Joshua Johnstun, Joshua Saxe, Junteng Jia, Kalyan~Vasuden Alwala, Karthik Prasad, Kartikeya Upasani, Kate Plawiak, Ke~Li, Kenneth Heafield, Kevin Stone, Khalid El-Arini, Krithika Iyer, Kshitiz Malik, Kuenley Chiu, Kunal Bhalla, Kushal Lakhotia, Lauren Rantala-Yeary, Laurens van~der Maaten, Lawrence Chen, Liang Tan, Liz Jenkins, Louis Martin, Lovish Madaan, Lubo Malo, Lukas Blecher, Lukas Landzaat, Luke de~Oliveira, Madeline Muzzi, Mahesh Pasupuleti, Mannat Singh, Manohar Paluri, Marcin Kardas, Maria Tsimpoukelli, Mathew Oldham, Mathieu Rita, Maya Pavlova, Melanie Kambadur,
  Mike Lewis, Min Si, Mitesh~Kumar Singh, Mona Hassan, Naman Goyal, Narjes Torabi, Nikolay Bashlykov, Nikolay Bogoychev, Niladri Chatterji, Ning Zhang, Olivier Duchenne, Onur Çelebi, Patrick Alrassy, Pengchuan Zhang, Pengwei Li, Petar Vasic, Peter Weng, Prajjwal Bhargava, Pratik Dubal, Praveen Krishnan, Punit~Singh Koura, Puxin Xu, Qing He, Qingxiao Dong, Ragavan Srinivasan, Raj Ganapathy, Ramon Calderer, Ricardo~Silveira Cabral, Robert Stojnic, Roberta Raileanu, Rohan Maheswari, Rohit Girdhar, Rohit Patel, Romain Sauvestre, Ronnie Polidoro, Roshan Sumbaly, Ross Taylor, Ruan Silva, Rui Hou, Rui Wang, Saghar Hosseini, Sahana Chennabasappa, Sanjay Singh, Sean Bell, Seohyun~Sonia Kim, Sergey Edunov, Shaoliang Nie, Sharan Narang, Sharath Raparthy, Sheng Shen, Shengye Wan, Shruti Bhosale, Shun Zhang, Simon Vandenhende, Soumya Batra, Spencer Whitman, Sten Sootla, Stephane Collot, Suchin Gururangan, Sydney Borodinsky, Tamar Herman, Tara Fowler, Tarek Sheasha, Thomas Georgiou, Thomas Scialom, Tobias Speckbacher,
  Todor Mihaylov, Tong Xiao, Ujjwal Karn, Vedanuj Goswami, Vibhor Gupta, Vignesh Ramanathan, Viktor Kerkez, Vincent Gonguet, Virginie Do, Vish Vogeti, Vítor Albiero, Vladan Petrovic, Weiwei Chu, Wenhan Xiong, Wenyin Fu, Whitney Meers, Xavier Martinet, Xiaodong Wang, Xiaofang Wang, Xiaoqing~Ellen Tan, Xide Xia, Xinfeng Xie, Xuchao Jia, Xuewei Wang, Yaelle Goldschlag, Yashesh Gaur, Yasmine Babaei, Yi~Wen, Yiwen Song, Yuchen Zhang, Yue Li, Yuning Mao, Zacharie~Delpierre Coudert, Zheng Yan, Zhengxing Chen, Zoe Papakipos, Aaditya Singh, Aayushi Srivastava, Abha Jain, Adam Kelsey, Adam Shajnfeld, Adithya Gangidi, Adolfo Victoria, Ahuva Goldstand, Ajay Menon, Ajay Sharma, Alex Boesenberg, Alexei Baevski, Allie Feinstein, Amanda Kallet, Amit Sangani, Amos Teo, Anam Yunus, Andrei Lupu, Andres Alvarado, Andrew Caples, Andrew Gu, Andrew Ho, Andrew Poulton, Andrew Ryan, Ankit Ramchandani, Annie Dong, Annie Franco, Anuj Goyal, Aparajita Saraf, Arkabandhu Chowdhury, Ashley Gabriel, Ashwin Bharambe, Assaf Eisenman, Azadeh
  Yazdan, Beau James, Ben Maurer, Benjamin Leonhardi, Bernie Huang, Beth Loyd, Beto~De Paola, Bhargavi Paranjape, Bing Liu, Bo~Wu, Boyu Ni, Braden Hancock, Bram Wasti, Brandon Spence, Brani Stojkovic, Brian Gamido, Britt Montalvo, Carl Parker, Carly Burton, Catalina Mejia, Ce~Liu, Changhan Wang, Changkyu Kim, Chao Zhou, Chester Hu, Ching-Hsiang Chu, Chris Cai, Chris Tindal, Christoph Feichtenhofer, Cynthia Gao, Damon Civin, Dana Beaty, Daniel Kreymer, Daniel Li, David Adkins, David Xu, Davide Testuggine, Delia David, Devi Parikh, Diana Liskovich, Didem Foss, Dingkang Wang, Duc Le, Dustin Holland, Edward Dowling, Eissa Jamil, Elaine Montgomery, Eleonora Presani, Emily Hahn, Emily Wood, Eric-Tuan Le, Erik Brinkman, Esteban Arcaute, Evan Dunbar, Evan Smothers, Fei Sun, Felix Kreuk, Feng Tian, Filippos Kokkinos, Firat Ozgenel, Francesco Caggioni, Frank Kanayet, Frank Seide, Gabriela~Medina Florez, Gabriella Schwarz, Gada Badeer, Georgia Swee, Gil Halpern, Grant Herman, Grigory Sizov, Guangyi, Zhang, Guna
  Lakshminarayanan, Hakan Inan, Hamid Shojanazeri, Han Zou, Hannah Wang, Hanwen Zha, Haroun Habeeb, Harrison Rudolph, Helen Suk, Henry Aspegren, Hunter Goldman, Hongyuan Zhan, Ibrahim Damlaj, Igor Molybog, Igor Tufanov, Ilias Leontiadis, Irina-Elena Veliche, Itai Gat, Jake Weissman, James Geboski, James Kohli, Janice Lam, Japhet Asher, Jean-Baptiste Gaya, Jeff Marcus, Jeff Tang, Jennifer Chan, Jenny Zhen, Jeremy Reizenstein, Jeremy Teboul, Jessica Zhong, Jian Jin, Jingyi Yang, Joe Cummings, Jon Carvill, Jon Shepard, Jonathan McPhie, Jonathan Torres, Josh Ginsburg, Junjie Wang, Kai Wu, Kam~Hou U, Karan Saxena, Kartikay Khandelwal, Katayoun Zand, Kathy Matosich, Kaushik Veeraraghavan, Kelly Michelena, Keqian Li, Kiran Jagadeesh, Kun Huang, Kunal Chawla, Kyle Huang, Lailin Chen, Lakshya Garg, Lavender A, Leandro Silva, Lee Bell, Lei Zhang, Liangpeng Guo, Licheng Yu, Liron Moshkovich, Luca Wehrstedt, Madian Khabsa, Manav Avalani, Manish Bhatt, Martynas Mankus, Matan Hasson, Matthew Lennie, Matthias Reso, Maxim
  Groshev, Maxim Naumov, Maya Lathi, Meghan Keneally, Miao Liu, Michael~L. Seltzer, Michal Valko, Michelle Restrepo, Mihir Patel, Mik Vyatskov, Mikayel Samvelyan, Mike Clark, Mike Macey, Mike Wang, Miquel~Jubert Hermoso, Mo~Metanat, Mohammad Rastegari, Munish Bansal, Nandhini Santhanam, Natascha Parks, Natasha White, Navyata Bawa, Nayan Singhal, Nick Egebo, Nicolas Usunier, Nikhil Mehta, Nikolay~Pavlovich Laptev, Ning Dong, Norman Cheng, Oleg Chernoguz, Olivia Hart, Omkar Salpekar, Ozlem Kalinli, Parkin Kent, Parth Parekh, Paul Saab, Pavan Balaji, Pedro Rittner, Philip Bontrager, Pierre Roux, Piotr Dollar, Polina Zvyagina, Prashant Ratanchandani, Pritish Yuvraj, Qian Liang, Rachad Alao, Rachel Rodriguez, Rafi Ayub, Raghotham Murthy, Raghu Nayani, Rahul Mitra, Rangaprabhu Parthasarathy, Raymond Li, Rebekkah Hogan, Robin Battey, Rocky Wang, Russ Howes, Ruty Rinott, Sachin Mehta, Sachin Siby, Sai~Jayesh Bondu, Samyak Datta, Sara Chugh, Sara Hunt, Sargun Dhillon, Sasha Sidorov, Satadru Pan, Saurabh Mahajan,
  Saurabh Verma, Seiji Yamamoto, Sharadh Ramaswamy, Shaun Lindsay, Shaun Lindsay, Sheng Feng, Shenghao Lin, Shengxin~Cindy Zha, Shishir Patil, Shiva Shankar, Shuqiang Zhang, Shuqiang Zhang, Sinong Wang, Sneha Agarwal, Soji Sajuyigbe, Soumith Chintala, Stephanie Max, Stephen Chen, Steve Kehoe, Steve Satterfield, Sudarshan Govindaprasad, Sumit Gupta, Summer Deng, Sungmin Cho, Sunny Virk, Suraj Subramanian, Sy~Choudhury, Sydney Goldman, Tal Remez, Tamar Glaser, Tamara Best, Thilo Koehler, Thomas Robinson, Tianhe Li, Tianjun Zhang, Tim Matthews, Timothy Chou, Tzook Shaked, Varun Vontimitta, Victoria Ajayi, Victoria Montanez, Vijai Mohan, Vinay~Satish Kumar, Vishal Mangla, Vlad Ionescu, Vlad Poenaru, Vlad~Tiberiu Mihailescu, Vladimir Ivanov, Wei Li, Wenchen Wang, Wenwen Jiang, Wes Bouaziz, Will Constable, Xiaocheng Tang, Xiaojian Wu, Xiaolan Wang, Xilun Wu, Xinbo Gao, Yaniv Kleinman, Yanjun Chen, Ye~Hu, Ye~Jia, Ye~Qi, Yenda Li, Yilin Zhang, Ying Zhang, Yossi Adi, Youngjin Nam, Yu, Wang, Yu~Zhao, Yuchen Hao, Yundi
  Qian, Yunlu Li, Yuzi He, Zach Rait, Zachary DeVito, Zef Rosnbrick, Zhaoduo Wen, Zhenyu Yang, Zhiwei Zhao, and Zhiyu Ma.
\newblock The llama 3 herd of models, 2024.
\newblock URL \url{https://arxiv.org/abs/2407.21783}.

\bibitem[Jiang et~al.(2023)Jiang, Sablayrolles, Mensch, Bamford, Chaplot, de~las Casas, Bressand, Lengyel, Lample, Saulnier, Lavaud, Lachaux, Stock, Scao, Lavril, Wang, Lacroix, and Sayed]{jiang2023mistral7b}
Albert~Q. Jiang, Alexandre Sablayrolles, Arthur Mensch, Chris Bamford, Devendra~Singh Chaplot, Diego de~las Casas, Florian Bressand, Gianna Lengyel, Guillaume Lample, Lucile Saulnier, Lélio~Renard Lavaud, Marie-Anne Lachaux, Pierre Stock, Teven~Le Scao, Thibaut Lavril, Thomas Wang, Timothée Lacroix, and William~El Sayed.
\newblock Mistral 7b, 2023.
\newblock URL \url{https://arxiv.org/abs/2310.06825}.

\bibitem[Jiang et~al.(2024)Jiang, Sablayrolles, Roux, Mensch, Savary, Bamford, Chaplot, de~las Casas, Hanna, Bressand, Lengyel, Bour, Lample, Lavaud, Saulnier, Lachaux, Stock, Subramanian, Yang, Antoniak, Scao, Gervet, Lavril, Wang, Lacroix, and Sayed]{jiang2024mixtralexperts}
Albert~Q. Jiang, Alexandre Sablayrolles, Antoine Roux, Arthur Mensch, Blanche Savary, Chris Bamford, Devendra~Singh Chaplot, Diego de~las Casas, Emma~Bou Hanna, Florian Bressand, Gianna Lengyel, Guillaume Bour, Guillaume Lample, Lélio~Renard Lavaud, Lucile Saulnier, Marie-Anne Lachaux, Pierre Stock, Sandeep Subramanian, Sophia Yang, Szymon Antoniak, Teven~Le Scao, Théophile Gervet, Thibaut Lavril, Thomas Wang, Timothée Lacroix, and William~El Sayed.
\newblock Mixtral of experts, 2024.
\newblock URL \url{https://arxiv.org/abs/2401.04088}.

\bibitem[Team et~al.(2024)Team, Georgiev, Lei, Burnell, Bai, Gulati, Tanzer, Vincent, Pan, Wang, Mariooryad, Ding, Geng, Alcober, Frostig, Omernick, Walker, Paduraru, Sorokin, Tacchetti, Gaffney, Daruki, Sercinoglu, Gleicher, Love, Voigtlaender, Jain, Surita, Mohamed, Blevins, Ahn, Zhu, Kawintiranon, Firat, Gu, Zhang, Rahtz, Faruqui, Clay, Gilmer, Co-Reyes, Penchev, Zhu, Morioka, Hui, Haridasan, Campos, Mahdieh, Guo, Hassan, Kilgour, Vezer, Cheng, de~Liedekerke, Goyal, Barham, Strouse, Noury, Adler, Sundararajan, Vikram, Lepikhin, Paganini, Garcia, Yang, Valter, Trebacz, Vodrahalli, Asawaroengchai, Ring, Kalb, Soares, Brahma, Steiner, Yu, Mentzer, He, Gonzalez, Xu, Kaufman, Shafey, Oh, Hennigan, van~den Driessche, Odoom, Lucic, Roelofs, Lall, Marathe, Chan, Ontanon, He, Teplyashin, Lai, Crone, Damoc, Ho, Riedel, Lenc, Yeh, Chowdhery, Xu, Kazemi, Amid, Petrushkina, Swersky, Khodaei, Chen, Larkin, Pinto, Yan, Badia, Patil, Hansen, Orr, Arnold, Grimstad, Dai, Douglas, Sinha, Yadav, Chen, Gribovskaya, Austin,
  Zhao, Patel, Komarek, Austin, Borgeaud, Friso, Goyal, Caine, Cao, Chung, Lamm, Barth-Maron, Kagohara, Olszewska, Chen, Shivakumar, Agarwal, Godhia, Rajwar, Snaider, Dotiwalla, Liu, Barua, Ungureanu, Zhang, Batsaikhan, Wirth, Qin, Danihelka, Doshi, Chadwick, Chen, Jain, Le, Kar, Gurumurthy, Li, Sang, Liu, Lamprou, Munoz, Lintz, Mehta, Howard, Reynolds, Aroyo, Wang, Blanco, Cassirer, Griffith, Das, Lee, Sygnowski, Fisher, Besley, Powell, Ahmed, Paulus, Reitter, Borsos, Joshi, Pope, Hand, Selo, Jain, Sethi, Goel, Makino, May, Yang, Schalkwyk, Butterfield, Hauth, Goldin, Hawkins, Senter, Brin, Woodman, Ritter, Noland, Giang, Bolina, Lee, Blyth, Mackinnon, Reid, Sarvana, Silver, Chen, Wang, Maggiore, Chang, Attaluri, Thornton, Chiu, Bunyan, Levine, Chung, Eltyshev, Si, Lillicrap, Brady, Aggarwal, Wu, Xu, McIlroy, Badola, Sandhu, Moreira, Stokowiec, Hemsley, Li, Tudor, Shyam, Rahimtoroghi, Haykal, Sprechmann, Zhou, Mincu, Li, Addanki, Krishna, Wu, Frechette, Eyal, Dafoe, Lacey, Whang, Avrahami, Zhang, Taropa,
  Lin, Toyama, Rutherford, Sano, Choe, Tomala, Safranek-Shrader, Kassner, Pajarskas, Harvey, Sechrist, Fortunato, Lyu, Elsayed, Kuang, Lottes, Chu, Jia, Chen, Humphreys, Baumli, Tao, Samuel, dos Santos, Andreassen, Rakićević, Grewe, Kumar, Winkler, Caton, Brock, Dalmia, Sheahan, Barr, Miao, Natsev, Devlin, Behbahani, Prost, Sun, Myaskovsky, Pillai, Hurt, Lazaridou, Xiong, Zheng, Pardo, Li, Horgan, Stanton, Ambar, Xia, Lince, Wang, Mustafa, Webson, Lee, Anil, Wicke, Dozat, Sinha, Piqueras, Dabir, Upadhyay, Boral, Hendricks, Fry, Djolonga, Su, Walker, Labanowski, Huang, Misra, Chen, Skerry-Ryan, Singh, Rijhwani, Yu, Castro-Ros, Changpinyo, Datta, Bagri, Hrafnkelsson, Maggioni, Zheng, Sulsky, Hou, Paine, Yang, Riesa, Rogozinska, Marcus, Badawy, Zhang, Wang, Miller, Greer, Sjos, Nova, Zen, Chaabouni, Rosca, Jiang, Chen, Liu, Sainath, Krikun, Polozov, Lespiau, Newlan, Cankara, Kwak, Xu, Chen, Coenen, Meyer, Tsihlas, Ma, Gottweis, Xing, Gu, Miao, Frank, Cankara, Ganapathy, Dasgupta, Hughes-Fitt, Chen, Reid, Rong,
  Fan, van Amersfoort, Zhuang, Cohen, Gu, Mohananey, Ilic, Tobin, Wieting, Bortsova, Thacker, Wang, Caveness, Chiu, Sezener, Kaskasoli, Baker, Millican, Elhawaty, Aisopos, Lebsack, Byrd, Dai, Jia, Wiethoff, Davoodi, Weston, Yagati, Ahuja, Gao, Pundak, Zhang, Azzam, Sim, Caelles, Keeling, Sharma, Swing, Li, Liu, Bostock, Bansal, Nado, Anand, Lipschultz, Karmarkar, Proleev, Ittycheriah, Yeganeh, Polovets, Faust, Sun, Rrustemi, Li, Shivanna, Liu, Welty, Lebron, Baddepudi, Krause, Parisotto, Soricut, Xu, Bloxwich, Johnson, Neyshabur, Mao-Jones, Wang, Ramasesh, Abbas, Guez, Segal, Nguyen, Svensson, Hou, York, Milan, Bridgers, Gworek, Tagliasacchi, Lee-Thorp, Chang, Guseynov, Hartman, Kwong, Zhao, Kashem, Cole, Miech, Tanburn, Phuong, Pavetic, Cevey, Comanescu, Ives, Yang, Du, Li, Zhang, Iinuma, Hu, Roy, Bijwadia, Zhu, Martins, Saputro, Gergely, Zheng, Jia, Antonoglou, Sadovsky, Gu, Bi, Andreev, Samangooei, Khan, Kocisky, Filos, Kumar, Bishop, Yu, Hodkinson, Mittal, Shah, Moufarek, Cheng, Bloniarz, Lee, Pejman,
  Michel, Spencer, Feinberg, Xiong, Savinov, Smith, Shakeri, Tran, Chesus, Bohnet, Tucker, von Glehn, Muir, Mao, Kazawa, Slone, Soparkar, Shrivastava, Cobon-Kerr, Sharman, Pavagadhi, Araya, Misiunas, Ghelani, Laskin, Barker, Li, Briukhov, Houlsby, Glaese, Lakshminarayanan, Schucher, Tang, Collins, Lim, Feng, Recasens, Lai, Magni, Cao, Siddhant, Ashwood, Orbay, Dehghani, Brennan, He, Xu, Gao, Saroufim, Molloy, Wu, Arnold, Chang, Schrittwieser, Buchatskaya, Radpour, Polacek, Giordano, Bapna, Tokumine, Hellendoorn, Sottiaux, Cogan, Severyn, Saleh, Thakoor, Shefey, Qiao, Gaba, yiin Chang, Swanson, Zhang, Lee, Rubenstein, Song, Kwiatkowski, Koop, Kannan, Kao, Schuh, Stjerngren, Ghiasi, Gibson, Vilnis, Yuan, Ferreira, Kamath, Klimenko, Franko, Xiao, Bhattacharya, Patel, Wang, Morris, Strudel, Sharma, Choy, Hashemi, Landon, Finkelstein, Jhakra, Frye, Barnes, Mauger, Daun, Baatarsukh, Tung, Farhan, Michalewski, Viola, de~Chaumont~Quitry, Lan, Hudson, Wang, Fischer, Zheng, White, Dragan, baptiste Alayrac, Ni, Pritzel,
  Iwanicki, Isard, Bulanova, Zilka, Dyer, Sachan, Srinivasan, Muckenhirn, Cai, Mandhane, Tariq, Rae, Wang, Ayoub, FitzGerald, Zhao, Han, Alberti, Garrette, Krishnakumar, Gimenez, Levskaya, Sohn, Matak, Iturrate, Chang, Xiang, Cao, Ranka, Brown, Hutter, Mirrokni, Chen, Yao, Egyed, Galilee, Liechty, Kallakuri, Palmer, Ghemawat, Liu, Tao, Thornton, Green, Jasarevic, Lin, Cotruta, Tan, Fiedel, Yu, Chi, Neitz, Heitkaemper, Sinha, Zhou, Sun, Kaed, Hulse, Mishra, Georgaki, Kudugunta, Farabet, Shafran, Vlasic, Tsitsulin, Ananthanarayanan, Carin, Su, Sun, V, Carvajal, Broder, Comsa, Repina, Wong, Chen, Hawkins, Filonov, Loher, Hirnschall, Wang, Ye, Burns, Cate, Wright, Piccinini, Zhang, Lin, Gog, Kulizhskaya, Sreevatsa, Song, Cobo, Iyer, Tekur, Garrido, Xiao, Kemp, Zheng, Li, Agarwal, Ngani, Goshvadi, Santamaria-Fernandez, Fica, Chen, Gorgolewski, Sun, Garg, Ye, Eslami, Hua, Simon, Joshi, Kim, Tenney, Potluri, Thiet, Yuan, Luisier, Chronopoulou, Scellato, Srinivasan, Chen, Koverkathu, Dalibard, Xu, Saeta, Anderson,
  Sellam, Fernando, Huot, Jung, Varadarajan, Quinn, Raul, Le, Habalov, Clark, Jalan, Bullard, Singhal, Luong, Wang, Rajayogam, Eisenschlos, Jia, Finchelstein, Yakubovich, Balle, Fink, Agarwal, Li, Dvijotham, Pal, Kang, Konzelmann, Beattie, Dousse, Wu, Crocker, Elkind, Jonnalagadda, Lee, Holtmann-Rice, Kallarackal, Liu, Vnukov, Vats, Invernizzi, Jafari, Zhou, Taylor, Prendki, Wu, Eccles, Liu, Kopparapu, Beaufays, Angermueller, Marzoca, Sarcar, Dib, Stanway, Perbet, Trdin, Sterneck, Khorlin, Li, Wu, Goenka, Madras, Goldshtein, Gierke, Zhou, Liu, Liang, White, Li, Singh, Bahargam, Epstein, Basu, Lao, Ozturel, Crous, Zhai, Lu, Tung, Gaur, Walton, Dixon, Zhang, Globerson, Uy, Bolt, Wiles, Nasr, Shumailov, Selvi, Piccinno, Aguilar, McCarthy, Khalman, Shukla, Galic, Carpenter, Villela, Zhang, Richardson, Martens, Bosnjak, Belle, Seibert, Alnahlawi, McWilliams, Singh, Louis, Ding, Popovici, Simicich, Knight, Mehta, Gupta, Shi, Fatehi, Mitrovic, Grills, Pagadora, Munkhdalai, Petrova, Eisenbud, Zhang, Yates, Mittal,
  Tripuraneni, Assael, Brovelli, Jain, Velimirovic, Akbulut, Mu, Macherey, Kumar, Xu, Qureshi, Comanici, Wiesner, Gong, Ruddock, Bauer, Felt, GP, Arnab, Zelle, Rothfuss, Rosgen, Shenoy, Seybold, Li, Mudigonda, Erdogan, Xia, Simsa, Michi, Yao, Yew, Kan, Caswell, Radebaugh, Elisseeff, Valenzuela, McKinney, Paterson, Cui, Latorre-Chimoto, Kim, Zeng, Durden, Ponnapalli, Sosea, Choquette-Choo, Manyika, Robenek, Vashisht, Pereira, Lam, Velic, Owusu-Afriyie, Lee, Bolukbasi, Parrish, Lu, Park, Venkatraman, Talbert, Rosique, Cheng, Sozanschi, Paszke, Kumar, Austin, Li, Salama, Perz, Kim, Dukkipati, Baryshnikov, Kaplanis, Sheng, Chervonyi, Unlu, de~Las~Casas, Askham, Tunyasuvunakool, Gimeno, Poder, Kwak, Miecnikowski, Mirrokni, Dimitriev, Parisi, Liu, Tsai, Shevlane, Kouridi, Garmon, Goedeckemeyer, Brown, Vijayakumar, Elqursh, Jazayeri, Huang, Carthy, Hoover, Kim, Kumar, Chen, Biles, Bingham, Rosen, Wang, Tan, Engel, Pongetti, de~Cesare, Hwang, Yu, Pullman, Narayanan, Levin, Gopal, Li, Aharoni, Trinh, Lo, Casagrande,
  Vij, Matthey, Ramadhana, Matthews, Carey, Johnson, Goranova, Shah, Ashraf, Dasgupta, Larsen, Wang, Vuyyuru, Jiang, Ijazi, Osawa, Smith, Boppana, Bilal, Koizumi, Xu, Altun, Shabat, Bariach, Korchemniy, Choo, Ronneberger, Iwuanyanwu, Zhao, Soergel, Hsieh, Cai, Iqbal, Sundermeyer, Chen, Bursztein, Malaviya, Biadsy, Shroff, Dhillon, Latkar, Dyer, Forbes, Nicosia, Nikolaev, Greene, Georgiev, Wang, Martin, Sedghi, Zhang, Banzal, Fritz, Rao, Wang, Zhang, Patraucean, Du, Mordatch, Jurin, Liu, Dubey, Mohan, Nowakowski, Ion, Wei, Tojo, Raad, Hudson, Keshava, Agrawal, Ramirez, Wu, Nguyen, Liu, Sewak, Petrini, Choi, Philips, Wang, Bica, Garg, Wilkiewicz, Agrawal, Li, Guo, Xue, Shaik, Leach, Khan, Wiesinger, Jerome, Chakladar, Wang, Ornduff, Abu, Ghaffarkhah, Wainwright, Cortes, Liu, Maynez, Terzis, Samangouei, Mansour, Kępa, Aubet, Algymr, Banica, Weisz, Orban, Senges, Andrejczuk, Geller, Santo, Anklin, Merey, Baeuml, Strohman, Bai, Petrov, Wu, Hassabis, Kavukcuoglu, Dean, and
  Vinyals]{geminiteam2024gemini15unlockingmultimodal}
Gemini Team, Petko Georgiev, Ving~Ian Lei, Ryan Burnell, Libin Bai, Anmol Gulati, Garrett Tanzer, Damien Vincent, Zhufeng Pan, Shibo Wang, Soroosh Mariooryad, Yifan Ding, Xinyang Geng, Fred Alcober, Roy Frostig, Mark Omernick, Lexi Walker, Cosmin Paduraru, Christina Sorokin, Andrea Tacchetti, Colin Gaffney, Samira Daruki, Olcan Sercinoglu, Zach Gleicher, Juliette Love, Paul Voigtlaender, Rohan Jain, Gabriela Surita, Kareem Mohamed, Rory Blevins, Junwhan Ahn, Tao Zhu, Kornraphop Kawintiranon, Orhan Firat, Yiming Gu, Yujing Zhang, Matthew Rahtz, Manaal Faruqui, Natalie Clay, Justin Gilmer, JD~Co-Reyes, Ivo Penchev, Rui Zhu, Nobuyuki Morioka, Kevin Hui, Krishna Haridasan, Victor Campos, Mahdis Mahdieh, Mandy Guo, Samer Hassan, Kevin Kilgour, Arpi Vezer, Heng-Tze Cheng, Raoul de~Liedekerke, Siddharth Goyal, Paul Barham, DJ~Strouse, Seb Noury, Jonas Adler, Mukund Sundararajan, Sharad Vikram, Dmitry Lepikhin, Michela Paganini, Xavier Garcia, Fan Yang, Dasha Valter, Maja Trebacz, Kiran Vodrahalli, Chulayuth
  Asawaroengchai, Roman Ring, Norbert Kalb, Livio~Baldini Soares, Siddhartha Brahma, David Steiner, Tianhe Yu, Fabian Mentzer, Antoine He, Lucas Gonzalez, Bibo Xu, Raphael~Lopez Kaufman, Laurent~El Shafey, Junhyuk Oh, Tom Hennigan, George van~den Driessche, Seth Odoom, Mario Lucic, Becca Roelofs, Sid Lall, Amit Marathe, Betty Chan, Santiago Ontanon, Luheng He, Denis Teplyashin, Jonathan Lai, Phil Crone, Bogdan Damoc, Lewis Ho, Sebastian Riedel, Karel Lenc, Chih-Kuan Yeh, Aakanksha Chowdhery, Yang Xu, Mehran Kazemi, Ehsan Amid, Anastasia Petrushkina, Kevin Swersky, Ali Khodaei, Gowoon Chen, Chris Larkin, Mario Pinto, Geng Yan, Adria~Puigdomenech Badia, Piyush Patil, Steven Hansen, Dave Orr, Sebastien M.~R. Arnold, Jordan Grimstad, Andrew Dai, Sholto Douglas, Rishika Sinha, Vikas Yadav, Xi~Chen, Elena Gribovskaya, Jacob Austin, Jeffrey Zhao, Kaushal Patel, Paul Komarek, Sophia Austin, Sebastian Borgeaud, Linda Friso, Abhimanyu Goyal, Ben Caine, Kris Cao, Da-Woon Chung, Matthew Lamm, Gabe Barth-Maron, Thais
  Kagohara, Kate Olszewska, Mia Chen, Kaushik Shivakumar, Rishabh Agarwal, Harshal Godhia, Ravi Rajwar, Javier Snaider, Xerxes Dotiwalla, Yuan Liu, Aditya Barua, Victor Ungureanu, Yuan Zhang, Bat-Orgil Batsaikhan, Mateo Wirth, James Qin, Ivo Danihelka, Tulsee Doshi, Martin Chadwick, Jilin Chen, Sanil Jain, Quoc Le, Arjun Kar, Madhu Gurumurthy, Cheng Li, Ruoxin Sang, Fangyu Liu, Lampros Lamprou, Rich Munoz, Nathan Lintz, Harsh Mehta, Heidi Howard, Malcolm Reynolds, Lora Aroyo, Quan Wang, Lorenzo Blanco, Albin Cassirer, Jordan Griffith, Dipanjan Das, Stephan Lee, Jakub Sygnowski, Zach Fisher, James Besley, Richard Powell, Zafarali Ahmed, Dominik Paulus, David Reitter, Zalan Borsos, Rishabh Joshi, Aedan Pope, Steven Hand, Vittorio Selo, Vihan Jain, Nikhil Sethi, Megha Goel, Takaki Makino, Rhys May, Zhen Yang, Johan Schalkwyk, Christina Butterfield, Anja Hauth, Alex Goldin, Will Hawkins, Evan Senter, Sergey Brin, Oliver Woodman, Marvin Ritter, Eric Noland, Minh Giang, Vijay Bolina, Lisa Lee, Tim Blyth, Ian
  Mackinnon, Machel Reid, Obaid Sarvana, David Silver, Alexander Chen, Lily Wang, Loren Maggiore, Oscar Chang, Nithya Attaluri, Gregory Thornton, Chung-Cheng Chiu, Oskar Bunyan, Nir Levine, Timothy Chung, Evgenii Eltyshev, Xiance Si, Timothy Lillicrap, Demetra Brady, Vaibhav Aggarwal, Boxi Wu, Yuanzhong Xu, Ross McIlroy, Kartikeya Badola, Paramjit Sandhu, Erica Moreira, Wojciech Stokowiec, Ross Hemsley, Dong Li, Alex Tudor, Pranav Shyam, Elahe Rahimtoroghi, Salem Haykal, Pablo Sprechmann, Xiang Zhou, Diana Mincu, Yujia Li, Ravi Addanki, Kalpesh Krishna, Xiao Wu, Alexandre Frechette, Matan Eyal, Allan Dafoe, Dave Lacey, Jay Whang, Thi Avrahami, Ye~Zhang, Emanuel Taropa, Hanzhao Lin, Daniel Toyama, Eliza Rutherford, Motoki Sano, HyunJeong Choe, Alex Tomala, Chalence Safranek-Shrader, Nora Kassner, Mantas Pajarskas, Matt Harvey, Sean Sechrist, Meire Fortunato, Christina Lyu, Gamaleldin Elsayed, Chenkai Kuang, James Lottes, Eric Chu, Chao Jia, Chih-Wei Chen, Peter Humphreys, Kate Baumli, Connie Tao, Rajkumar
  Samuel, Cicero~Nogueira dos Santos, Anders Andreassen, Nemanja Rakićević, Dominik Grewe, Aviral Kumar, Stephanie Winkler, Jonathan Caton, Andrew Brock, Sid Dalmia, Hannah Sheahan, Iain Barr, Yingjie Miao, Paul Natsev, Jacob Devlin, Feryal Behbahani, Flavien Prost, Yanhua Sun, Artiom Myaskovsky, Thanumalayan~Sankaranarayana Pillai, Dan Hurt, Angeliki Lazaridou, Xi~Xiong, Ce~Zheng, Fabio Pardo, Xiaowei Li, Dan Horgan, Joe Stanton, Moran Ambar, Fei Xia, Alejandro Lince, Mingqiu Wang, Basil Mustafa, Albert Webson, Hyo Lee, Rohan Anil, Martin Wicke, Timothy Dozat, Abhishek Sinha, Enrique Piqueras, Elahe Dabir, Shyam Upadhyay, Anudhyan Boral, Lisa~Anne Hendricks, Corey Fry, Josip Djolonga, Yi~Su, Jake Walker, Jane Labanowski, Ronny Huang, Vedant Misra, Jeremy Chen, RJ~Skerry-Ryan, Avi Singh, Shruti Rijhwani, Dian Yu, Alex Castro-Ros, Beer Changpinyo, Romina Datta, Sumit Bagri, Arnar~Mar Hrafnkelsson, Marcello Maggioni, Daniel Zheng, Yury Sulsky, Shaobo Hou, Tom~Le Paine, Antoine Yang, Jason Riesa, Dominika
  Rogozinska, Dror Marcus, Dalia~El Badawy, Qiao Zhang, Luyu Wang, Helen Miller, Jeremy Greer, Lars~Lowe Sjos, Azade Nova, Heiga Zen, Rahma Chaabouni, Mihaela Rosca, Jiepu Jiang, Charlie Chen, Ruibo Liu, Tara Sainath, Maxim Krikun, Alex Polozov, Jean-Baptiste Lespiau, Josh Newlan, Zeyncep Cankara, Soo Kwak, Yunhan Xu, Phil Chen, Andy Coenen, Clemens Meyer, Katerina Tsihlas, Ada Ma, Juraj Gottweis, Jinwei Xing, Chenjie Gu, Jin Miao, Christian Frank, Zeynep Cankara, Sanjay Ganapathy, Ishita Dasgupta, Steph Hughes-Fitt, Heng Chen, David Reid, Keran Rong, Hongmin Fan, Joost van Amersfoort, Vincent Zhuang, Aaron Cohen, Shixiang~Shane Gu, Anhad Mohananey, Anastasija Ilic, Taylor Tobin, John Wieting, Anna Bortsova, Phoebe Thacker, Emma Wang, Emily Caveness, Justin Chiu, Eren Sezener, Alex Kaskasoli, Steven Baker, Katie Millican, Mohamed Elhawaty, Kostas Aisopos, Carl Lebsack, Nathan Byrd, Hanjun Dai, Wenhao Jia, Matthew Wiethoff, Elnaz Davoodi, Albert Weston, Lakshman Yagati, Arun Ahuja, Isabel Gao, Golan Pundak,
  Susan Zhang, Michael Azzam, Khe~Chai Sim, Sergi Caelles, James Keeling, Abhanshu Sharma, Andy Swing, YaGuang Li, Chenxi Liu, Carrie~Grimes Bostock, Yamini Bansal, Zachary Nado, Ankesh Anand, Josh Lipschultz, Abhijit Karmarkar, Lev Proleev, Abe Ittycheriah, Soheil~Hassas Yeganeh, George Polovets, Aleksandra Faust, Jiao Sun, Alban Rrustemi, Pen Li, Rakesh Shivanna, Jeremiah Liu, Chris Welty, Federico Lebron, Anirudh Baddepudi, Sebastian Krause, Emilio Parisotto, Radu Soricut, Zheng Xu, Dawn Bloxwich, Melvin Johnson, Behnam Neyshabur, Justin Mao-Jones, Renshen Wang, Vinay Ramasesh, Zaheer Abbas, Arthur Guez, Constant Segal, Duc~Dung Nguyen, James Svensson, Le~Hou, Sarah York, Kieran Milan, Sophie Bridgers, Wiktor Gworek, Marco Tagliasacchi, James Lee-Thorp, Michael Chang, Alexey Guseynov, Ale~Jakse Hartman, Michael Kwong, Ruizhe Zhao, Sheleem Kashem, Elizabeth Cole, Antoine Miech, Richard Tanburn, Mary Phuong, Filip Pavetic, Sebastien Cevey, Ramona Comanescu, Richard Ives, Sherry Yang, Cosmo Du, Bo~Li, Zizhao
  Zhang, Mariko Iinuma, Clara~Huiyi Hu, Aurko Roy, Shaan Bijwadia, Zhenkai Zhu, Danilo Martins, Rachel Saputro, Anita Gergely, Steven Zheng, Dawei Jia, Ioannis Antonoglou, Adam Sadovsky, Shane Gu, Yingying Bi, Alek Andreev, Sina Samangooei, Mina Khan, Tomas Kocisky, Angelos Filos, Chintu Kumar, Colton Bishop, Adams Yu, Sarah Hodkinson, Sid Mittal, Premal Shah, Alexandre Moufarek, Yong Cheng, Adam Bloniarz, Jaehoon Lee, Pedram Pejman, Paul Michel, Stephen Spencer, Vladimir Feinberg, Xuehan Xiong, Nikolay Savinov, Charlotte Smith, Siamak Shakeri, Dustin Tran, Mary Chesus, Bernd Bohnet, George Tucker, Tamara von Glehn, Carrie Muir, Yiran Mao, Hideto Kazawa, Ambrose Slone, Kedar Soparkar, Disha Shrivastava, James Cobon-Kerr, Michael Sharman, Jay Pavagadhi, Carlos Araya, Karolis Misiunas, Nimesh Ghelani, Michael Laskin, David Barker, Qiujia Li, Anton Briukhov, Neil Houlsby, Mia Glaese, Balaji Lakshminarayanan, Nathan Schucher, Yunhao Tang, Eli Collins, Hyeontaek Lim, Fangxiaoyu Feng, Adria Recasens, Guangda Lai,
  Alberto Magni, Nicola~De Cao, Aditya Siddhant, Zoe Ashwood, Jordi Orbay, Mostafa Dehghani, Jenny Brennan, Yifan He, Kelvin Xu, Yang Gao, Carl Saroufim, James Molloy, Xinyi Wu, Seb Arnold, Solomon Chang, Julian Schrittwieser, Elena Buchatskaya, Soroush Radpour, Martin Polacek, Skye Giordano, Ankur Bapna, Simon Tokumine, Vincent Hellendoorn, Thibault Sottiaux, Sarah Cogan, Aliaksei Severyn, Mohammad Saleh, Shantanu Thakoor, Laurent Shefey, Siyuan Qiao, Meenu Gaba, Shuo yiin Chang, Craig Swanson, Biao Zhang, Benjamin Lee, Paul~Kishan Rubenstein, Gan Song, Tom Kwiatkowski, Anna Koop, Ajay Kannan, David Kao, Parker Schuh, Axel Stjerngren, Golnaz Ghiasi, Gena Gibson, Luke Vilnis, Ye~Yuan, Felipe~Tiengo Ferreira, Aishwarya Kamath, Ted Klimenko, Ken Franko, Kefan Xiao, Indro Bhattacharya, Miteyan Patel, Rui Wang, Alex Morris, Robin Strudel, Vivek Sharma, Peter Choy, Sayed~Hadi Hashemi, Jessica Landon, Mara Finkelstein, Priya Jhakra, Justin Frye, Megan Barnes, Matthew Mauger, Dennis Daun, Khuslen Baatarsukh, Matthew
  Tung, Wael Farhan, Henryk Michalewski, Fabio Viola, Felix de~Chaumont~Quitry, Charline~Le Lan, Tom Hudson, Qingze Wang, Felix Fischer, Ivy Zheng, Elspeth White, Anca Dragan, Jean baptiste Alayrac, Eric Ni, Alexander Pritzel, Adam Iwanicki, Michael Isard, Anna Bulanova, Lukas Zilka, Ethan Dyer, Devendra Sachan, Srivatsan Srinivasan, Hannah Muckenhirn, Honglong Cai, Amol Mandhane, Mukarram Tariq, Jack~W. Rae, Gary Wang, Kareem Ayoub, Nicholas FitzGerald, Yao Zhao, Woohyun Han, Chris Alberti, Dan Garrette, Kashyap Krishnakumar, Mai Gimenez, Anselm Levskaya, Daniel Sohn, Josip Matak, Inaki Iturrate, Michael~B. Chang, Jackie Xiang, Yuan Cao, Nishant Ranka, Geoff Brown, Adrian Hutter, Vahab Mirrokni, Nanxin Chen, Kaisheng Yao, Zoltan Egyed, Francois Galilee, Tyler Liechty, Praveen Kallakuri, Evan Palmer, Sanjay Ghemawat, Jasmine Liu, David Tao, Chloe Thornton, Tim Green, Mimi Jasarevic, Sharon Lin, Victor Cotruta, Yi-Xuan Tan, Noah Fiedel, Hongkun Yu, Ed~Chi, Alexander Neitz, Jens Heitkaemper, Anu Sinha, Denny
  Zhou, Yi~Sun, Charbel Kaed, Brice Hulse, Swaroop Mishra, Maria Georgaki, Sneha Kudugunta, Clement Farabet, Izhak Shafran, Daniel Vlasic, Anton Tsitsulin, Rajagopal Ananthanarayanan, Alen Carin, Guolong Su, Pei Sun, Shashank V, Gabriel Carvajal, Josef Broder, Iulia Comsa, Alena Repina, William Wong, Warren~Weilun Chen, Peter Hawkins, Egor Filonov, Lucia Loher, Christoph Hirnschall, Weiyi Wang, Jingchen Ye, Andrea Burns, Hardie Cate, Diana~Gage Wright, Federico Piccinini, Lei Zhang, Chu-Cheng Lin, Ionel Gog, Yana Kulizhskaya, Ashwin Sreevatsa, Shuang Song, Luis~C. Cobo, Anand Iyer, Chetan Tekur, Guillermo Garrido, Zhuyun Xiao, Rupert Kemp, Huaixiu~Steven Zheng, Hui Li, Ananth Agarwal, Christel Ngani, Kati Goshvadi, Rebeca Santamaria-Fernandez, Wojciech Fica, Xinyun Chen, Chris Gorgolewski, Sean Sun, Roopal Garg, Xinyu Ye, S.~M.~Ali Eslami, Nan Hua, Jon Simon, Pratik Joshi, Yelin Kim, Ian Tenney, Sahitya Potluri, Lam~Nguyen Thiet, Quan Yuan, Florian Luisier, Alexandra Chronopoulou, Salvatore Scellato, Praveen
  Srinivasan, Minmin Chen, Vinod Koverkathu, Valentin Dalibard, Yaming Xu, Brennan Saeta, Keith Anderson, Thibault Sellam, Nick Fernando, Fantine Huot, Junehyuk Jung, Mani Varadarajan, Michael Quinn, Amit Raul, Maigo Le, Ruslan Habalov, Jon Clark, Komal Jalan, Kalesha Bullard, Achintya Singhal, Thang Luong, Boyu Wang, Sujeevan Rajayogam, Julian Eisenschlos, Johnson Jia, Daniel Finchelstein, Alex Yakubovich, Daniel Balle, Michael Fink, Sameer Agarwal, Jing Li, Dj~Dvijotham, Shalini Pal, Kai Kang, Jaclyn Konzelmann, Jennifer Beattie, Olivier Dousse, Diane Wu, Remi Crocker, Chen Elkind, Siddhartha~Reddy Jonnalagadda, Jong Lee, Dan Holtmann-Rice, Krystal Kallarackal, Rosanne Liu, Denis Vnukov, Neera Vats, Luca Invernizzi, Mohsen Jafari, Huanjie Zhou, Lilly Taylor, Jennifer Prendki, Marcus Wu, Tom Eccles, Tianqi Liu, Kavya Kopparapu, Francoise Beaufays, Christof Angermueller, Andreea Marzoca, Shourya Sarcar, Hilal Dib, Jeff Stanway, Frank Perbet, Nejc Trdin, Rachel Sterneck, Andrey Khorlin, Dinghua Li, Xihui Wu,
  Sonam Goenka, David Madras, Sasha Goldshtein, Willi Gierke, Tong Zhou, Yaxin Liu, Yannie Liang, Anais White, Yunjie Li, Shreya Singh, Sanaz Bahargam, Mark Epstein, Sujoy Basu, Li~Lao, Adnan Ozturel, Carl Crous, Alex Zhai, Han Lu, Zora Tung, Neeraj Gaur, Alanna Walton, Lucas Dixon, Ming Zhang, Amir Globerson, Grant Uy, Andrew Bolt, Olivia Wiles, Milad Nasr, Ilia Shumailov, Marco Selvi, Francesco Piccinno, Ricardo Aguilar, Sara McCarthy, Misha Khalman, Mrinal Shukla, Vlado Galic, John Carpenter, Kevin Villela, Haibin Zhang, Harry Richardson, James Martens, Matko Bosnjak, Shreyas~Rammohan Belle, Jeff Seibert, Mahmoud Alnahlawi, Brian McWilliams, Sankalp Singh, Annie Louis, Wen Ding, Dan Popovici, Lenin Simicich, Laura Knight, Pulkit Mehta, Nishesh Gupta, Chongyang Shi, Saaber Fatehi, Jovana Mitrovic, Alex Grills, Joseph Pagadora, Tsendsuren Munkhdalai, Dessie Petrova, Danielle Eisenbud, Zhishuai Zhang, Damion Yates, Bhavishya Mittal, Nilesh Tripuraneni, Yannis Assael, Thomas Brovelli, Prateek Jain, Mihajlo
  Velimirovic, Canfer Akbulut, Jiaqi Mu, Wolfgang Macherey, Ravin Kumar, Jun Xu, Haroon Qureshi, Gheorghe Comanici, Jeremy Wiesner, Zhitao Gong, Anton Ruddock, Matthias Bauer, Nick Felt, Anirudh GP, Anurag Arnab, Dustin Zelle, Jonas Rothfuss, Bill Rosgen, Ashish Shenoy, Bryan Seybold, Xinjian Li, Jayaram Mudigonda, Goker Erdogan, Jiawei Xia, Jiri Simsa, Andrea Michi, Yi~Yao, Christopher Yew, Steven Kan, Isaac Caswell, Carey Radebaugh, Andre Elisseeff, Pedro Valenzuela, Kay McKinney, Kim Paterson, Albert Cui, Eri Latorre-Chimoto, Solomon Kim, William Zeng, Ken Durden, Priya Ponnapalli, Tiberiu Sosea, Christopher~A. Choquette-Choo, James Manyika, Brona Robenek, Harsha Vashisht, Sebastien Pereira, Hoi Lam, Marko Velic, Denese Owusu-Afriyie, Katherine Lee, Tolga Bolukbasi, Alicia Parrish, Shawn Lu, Jane Park, Balaji Venkatraman, Alice Talbert, Lambert Rosique, Yuchung Cheng, Andrei Sozanschi, Adam Paszke, Praveen Kumar, Jessica Austin, Lu~Li, Khalid Salama, Bartek Perz, Wooyeol Kim, Nandita Dukkipati, Anthony
  Baryshnikov, Christos Kaplanis, XiangHai Sheng, Yuri Chervonyi, Caglar Unlu, Diego de~Las~Casas, Harry Askham, Kathryn Tunyasuvunakool, Felix Gimeno, Siim Poder, Chester Kwak, Matt Miecnikowski, Vahab Mirrokni, Alek Dimitriev, Aaron Parisi, Dangyi Liu, Tomy Tsai, Toby Shevlane, Christina Kouridi, Drew Garmon, Adrian Goedeckemeyer, Adam~R. Brown, Anitha Vijayakumar, Ali Elqursh, Sadegh Jazayeri, Jin Huang, Sara~Mc Carthy, Jay Hoover, Lucy Kim, Sandeep Kumar, Wei Chen, Courtney Biles, Garrett Bingham, Evan Rosen, Lisa Wang, Qijun Tan, David Engel, Francesco Pongetti, Dario de~Cesare, Dongseong Hwang, Lily Yu, Jennifer Pullman, Srini Narayanan, Kyle Levin, Siddharth Gopal, Megan Li, Asaf Aharoni, Trieu Trinh, Jessica Lo, Norman Casagrande, Roopali Vij, Loic Matthey, Bramandia Ramadhana, Austin Matthews, CJ~Carey, Matthew Johnson, Kremena Goranova, Rohin Shah, Shereen Ashraf, Kingshuk Dasgupta, Rasmus Larsen, Yicheng Wang, Manish~Reddy Vuyyuru, Chong Jiang, Joana Ijazi, Kazuki Osawa, Celine Smith, Ramya~Sree
  Boppana, Taylan Bilal, Yuma Koizumi, Ying Xu, Yasemin Altun, Nir Shabat, Ben Bariach, Alex Korchemniy, Kiam Choo, Olaf Ronneberger, Chimezie Iwuanyanwu, Shubin Zhao, David Soergel, Cho-Jui Hsieh, Irene Cai, Shariq Iqbal, Martin Sundermeyer, Zhe Chen, Elie Bursztein, Chaitanya Malaviya, Fadi Biadsy, Prakash Shroff, Inderjit Dhillon, Tejasi Latkar, Chris Dyer, Hannah Forbes, Massimo Nicosia, Vitaly Nikolaev, Somer Greene, Marin Georgiev, Pidong Wang, Nina Martin, Hanie Sedghi, John Zhang, Praseem Banzal, Doug Fritz, Vikram Rao, Xuezhi Wang, Jiageng Zhang, Viorica Patraucean, Dayou Du, Igor Mordatch, Ivan Jurin, Lewis Liu, Ayush Dubey, Abhi Mohan, Janek Nowakowski, Vlad-Doru Ion, Nan Wei, Reiko Tojo, Maria~Abi Raad, Drew~A. Hudson, Vaishakh Keshava, Shubham Agrawal, Kevin Ramirez, Zhichun Wu, Hoang Nguyen, Ji~Liu, Madhavi Sewak, Bryce Petrini, DongHyun Choi, Ivan Philips, Ziyue Wang, Ioana Bica, Ankush Garg, Jarek Wilkiewicz, Priyanka Agrawal, Xiaowei Li, Danhao Guo, Emily Xue, Naseer Shaik, Andrew Leach,
  Sadh~MNM Khan, Julia Wiesinger, Sammy Jerome, Abhishek Chakladar, Alek~Wenjiao Wang, Tina Ornduff, Folake Abu, Alireza Ghaffarkhah, Marcus Wainwright, Mario Cortes, Frederick Liu, Joshua Maynez, Andreas Terzis, Pouya Samangouei, Riham Mansour, Tomasz Kępa, François-Xavier Aubet, Anton Algymr, Dan Banica, Agoston Weisz, Andras Orban, Alexandre Senges, Ewa Andrejczuk, Mark Geller, Niccolo~Dal Santo, Valentin Anklin, Majd~Al Merey, Martin Baeuml, Trevor Strohman, Junwen Bai, Slav Petrov, Yonghui Wu, Demis Hassabis, Koray Kavukcuoglu, Jeff Dean, and Oriol Vinyals.
\newblock Gemini 1.5: Unlocking multimodal understanding across millions of tokens of context, 2024.
\newblock URL \url{https://arxiv.org/abs/2403.05530}.

\bibitem[OpenAI et~al.(2024)OpenAI, Achiam, Adler, Agarwal, Ahmad, Akkaya, Aleman, Almeida, Altenschmidt, Altman, Anadkat, Avila, Babuschkin, Balaji, Balcom, Baltescu, Bao, Bavarian, Belgum, Bello, Berdine, Bernadett-Shapiro, Berner, Bogdonoff, Boiko, Boyd, Brakman, Brockman, Brooks, Brundage, Button, Cai, Campbell, Cann, Carey, Carlson, Carmichael, Chan, Chang, Chantzis, Chen, Chen, Chen, Chen, Chen, Chess, Cho, Chu, Chung, Cummings, Currier, Dai, Decareaux, Degry, Deutsch, Deville, Dhar, Dohan, Dowling, Dunning, Ecoffet, Eleti, Eloundou, Farhi, Fedus, Felix, Fishman, Forte, Fulford, Gao, Georges, Gibson, Goel, Gogineni, Goh, Gontijo-Lopes, Gordon, Grafstein, Gray, Greene, Gross, Gu, Guo, Hallacy, Han, Harris, He, Heaton, Heidecke, Hesse, Hickey, Hickey, Hoeschele, Houghton, Hsu, Hu, Hu, Huizinga, Jain, Jain, Jang, Jiang, Jiang, Jin, Jin, Jomoto, Jonn, Jun, Kaftan, Łukasz Kaiser, Kamali, Kanitscheider, Keskar, Khan, Kilpatrick, Kim, Kim, Kim, Kirchner, Kiros, Knight, Kokotajlo, Łukasz Kondraciuk, Kondrich,
  Konstantinidis, Kosic, Krueger, Kuo, Lampe, Lan, Lee, Leike, Leung, Levy, Li, Lim, Lin, Lin, Litwin, Lopez, Lowe, Lue, Makanju, Malfacini, Manning, Markov, Markovski, Martin, Mayer, Mayne, McGrew, McKinney, McLeavey, McMillan, McNeil, Medina, Mehta, Menick, Metz, Mishchenko, Mishkin, Monaco, Morikawa, Mossing, Mu, Murati, Murk, Mély, Nair, Nakano, Nayak, Neelakantan, Ngo, Noh, Ouyang, O'Keefe, Pachocki, Paino, Palermo, Pantuliano, Parascandolo, Parish, Parparita, Passos, Pavlov, Peng, Perelman, de~Avila Belbute~Peres, Petrov, de~Oliveira~Pinto, Michael, Pokorny, Pokrass, Pong, Powell, Power, Power, Proehl, Puri, Radford, Rae, Ramesh, Raymond, Real, Rimbach, Ross, Rotsted, Roussez, Ryder, Saltarelli, Sanders, Santurkar, Sastry, Schmidt, Schnurr, Schulman, Selsam, Sheppard, Sherbakov, Shieh, Shoker, Shyam, Sidor, Sigler, Simens, Sitkin, Slama, Sohl, Sokolowsky, Song, Staudacher, Such, Summers, Sutskever, Tang, Tezak, Thompson, Tillet, Tootoonchian, Tseng, Tuggle, Turley, Tworek, Uribe, Vallone, Vijayvergiya,
  Voss, Wainwright, Wang, Wang, Wang, Ward, Wei, Weinmann, Welihinda, Welinder, Weng, Weng, Wiethoff, Willner, Winter, Wolrich, Wong, Workman, Wu, Wu, Wu, Xiao, Xu, Yoo, Yu, Yuan, Zaremba, Zellers, Zhang, Zhang, Zhao, Zheng, Zhuang, Zhuk, and Zoph]{openai2024gpt4technicalreport}
OpenAI, Josh Achiam, Steven Adler, Sandhini Agarwal, Lama Ahmad, Ilge Akkaya, Florencia~Leoni Aleman, Diogo Almeida, Janko Altenschmidt, Sam Altman, Shyamal Anadkat, Red Avila, Igor Babuschkin, Suchir Balaji, Valerie Balcom, Paul Baltescu, Haiming Bao, Mohammad Bavarian, Jeff Belgum, Irwan Bello, Jake Berdine, Gabriel Bernadett-Shapiro, Christopher Berner, Lenny Bogdonoff, Oleg Boiko, Madelaine Boyd, Anna-Luisa Brakman, Greg Brockman, Tim Brooks, Miles Brundage, Kevin Button, Trevor Cai, Rosie Campbell, Andrew Cann, Brittany Carey, Chelsea Carlson, Rory Carmichael, Brooke Chan, Che Chang, Fotis Chantzis, Derek Chen, Sully Chen, Ruby Chen, Jason Chen, Mark Chen, Ben Chess, Chester Cho, Casey Chu, Hyung~Won Chung, Dave Cummings, Jeremiah Currier, Yunxing Dai, Cory Decareaux, Thomas Degry, Noah Deutsch, Damien Deville, Arka Dhar, David Dohan, Steve Dowling, Sheila Dunning, Adrien Ecoffet, Atty Eleti, Tyna Eloundou, David Farhi, Liam Fedus, Niko Felix, Simón~Posada Fishman, Juston Forte, Isabella Fulford, Leo
  Gao, Elie Georges, Christian Gibson, Vik Goel, Tarun Gogineni, Gabriel Goh, Rapha Gontijo-Lopes, Jonathan Gordon, Morgan Grafstein, Scott Gray, Ryan Greene, Joshua Gross, Shixiang~Shane Gu, Yufei Guo, Chris Hallacy, Jesse Han, Jeff Harris, Yuchen He, Mike Heaton, Johannes Heidecke, Chris Hesse, Alan Hickey, Wade Hickey, Peter Hoeschele, Brandon Houghton, Kenny Hsu, Shengli Hu, Xin Hu, Joost Huizinga, Shantanu Jain, Shawn Jain, Joanne Jang, Angela Jiang, Roger Jiang, Haozhun Jin, Denny Jin, Shino Jomoto, Billie Jonn, Heewoo Jun, Tomer Kaftan, Łukasz Kaiser, Ali Kamali, Ingmar Kanitscheider, Nitish~Shirish Keskar, Tabarak Khan, Logan Kilpatrick, Jong~Wook Kim, Christina Kim, Yongjik Kim, Jan~Hendrik Kirchner, Jamie Kiros, Matt Knight, Daniel Kokotajlo, Łukasz Kondraciuk, Andrew Kondrich, Aris Konstantinidis, Kyle Kosic, Gretchen Krueger, Vishal Kuo, Michael Lampe, Ikai Lan, Teddy Lee, Jan Leike, Jade Leung, Daniel Levy, Chak~Ming Li, Rachel Lim, Molly Lin, Stephanie Lin, Mateusz Litwin, Theresa Lopez, Ryan
  Lowe, Patricia Lue, Anna Makanju, Kim Malfacini, Sam Manning, Todor Markov, Yaniv Markovski, Bianca Martin, Katie Mayer, Andrew Mayne, Bob McGrew, Scott~Mayer McKinney, Christine McLeavey, Paul McMillan, Jake McNeil, David Medina, Aalok Mehta, Jacob Menick, Luke Metz, Andrey Mishchenko, Pamela Mishkin, Vinnie Monaco, Evan Morikawa, Daniel Mossing, Tong Mu, Mira Murati, Oleg Murk, David Mély, Ashvin Nair, Reiichiro Nakano, Rajeev Nayak, Arvind Neelakantan, Richard Ngo, Hyeonwoo Noh, Long Ouyang, Cullen O'Keefe, Jakub Pachocki, Alex Paino, Joe Palermo, Ashley Pantuliano, Giambattista Parascandolo, Joel Parish, Emy Parparita, Alex Passos, Mikhail Pavlov, Andrew Peng, Adam Perelman, Filipe de~Avila Belbute~Peres, Michael Petrov, Henrique~Ponde de~Oliveira~Pinto, Michael, Pokorny, Michelle Pokrass, Vitchyr~H. Pong, Tolly Powell, Alethea Power, Boris Power, Elizabeth Proehl, Raul Puri, Alec Radford, Jack Rae, Aditya Ramesh, Cameron Raymond, Francis Real, Kendra Rimbach, Carl Ross, Bob Rotsted, Henri Roussez,
  Nick Ryder, Mario Saltarelli, Ted Sanders, Shibani Santurkar, Girish Sastry, Heather Schmidt, David Schnurr, John Schulman, Daniel Selsam, Kyla Sheppard, Toki Sherbakov, Jessica Shieh, Sarah Shoker, Pranav Shyam, Szymon Sidor, Eric Sigler, Maddie Simens, Jordan Sitkin, Katarina Slama, Ian Sohl, Benjamin Sokolowsky, Yang Song, Natalie Staudacher, Felipe~Petroski Such, Natalie Summers, Ilya Sutskever, Jie Tang, Nikolas Tezak, Madeleine~B. Thompson, Phil Tillet, Amin Tootoonchian, Elizabeth Tseng, Preston Tuggle, Nick Turley, Jerry Tworek, Juan Felipe~Cerón Uribe, Andrea Vallone, Arun Vijayvergiya, Chelsea Voss, Carroll Wainwright, Justin~Jay Wang, Alvin Wang, Ben Wang, Jonathan Ward, Jason Wei, CJ~Weinmann, Akila Welihinda, Peter Welinder, Jiayi Weng, Lilian Weng, Matt Wiethoff, Dave Willner, Clemens Winter, Samuel Wolrich, Hannah Wong, Lauren Workman, Sherwin Wu, Jeff Wu, Michael Wu, Kai Xiao, Tao Xu, Sarah Yoo, Kevin Yu, Qiming Yuan, Wojciech Zaremba, Rowan Zellers, Chong Zhang, Marvin Zhang, Shengjia
  Zhao, Tianhao Zheng, Juntang Zhuang, William Zhuk, and Barret Zoph.
\newblock Gpt-4 technical report, 2024.
\newblock URL \url{https://arxiv.org/abs/2303.08774}.

\bibitem[Zhang et~al.(2023{\natexlab{c}})Zhang, Dong, Li, Zhang, Sun, Wang, Li, Hu, Zhang, Wu, et~al.]{zhang2023instruction}
Shengyu Zhang, Linfeng Dong, Xiaoya Li, Sen Zhang, Xiaofei Sun, Shuhe Wang, Jiwei Li, Runyi Hu, Tianwei Zhang, Fei Wu, et~al.
\newblock Instruction tuning for large language models: A survey.
\newblock \emph{arXiv preprint arXiv:2308.10792}, 2023{\natexlab{c}}.

\bibitem[Longpre et~al.(2023)Longpre, Hou, Vu, Webson, Chung, Tay, Zhou, Le, Zoph, Wei, et~al.]{longpre2023flan}
Shayne Longpre, Le~Hou, Tu~Vu, Albert Webson, Hyung~Won Chung, Yi~Tay, Denny Zhou, Quoc~V Le, Barret Zoph, Jason Wei, et~al.
\newblock The flan collection: Designing data and methods for effective instruction tuning.
\newblock In \emph{International Conference on Machine Learning}, pages 22631--22648. PMLR, 2023.

\bibitem[Peng et~al.(2023)Peng, Li, He, Galley, and Gao]{peng2023instruction}
Baolin Peng, Chunyuan Li, Pengcheng He, Michel Galley, and Jianfeng Gao.
\newblock Instruction tuning with gpt-4.
\newblock \emph{arXiv preprint arXiv:2304.03277}, 2023.

\bibitem[Liu et~al.(2023)Liu, Li, Wu, and Lee]{liu2023visual}
Haotian Liu, Chunyuan Li, Qingyang Wu, and Yong~Jae Lee.
\newblock Visual instruction tuning.
\newblock \emph{Advances in neural information processing systems}, 36:\penalty0 34892--34916, 2023.

\bibitem[Wei et~al.(2022)Wei, Wang, Schuurmans, Bosma, Xia, Chi, Le, Zhou, et~al.]{wei2022chain}
Jason Wei, Xuezhi Wang, Dale Schuurmans, Maarten Bosma, Fei Xia, Ed~Chi, Quoc~V Le, Denny Zhou, et~al.
\newblock Chain-of-thought prompting elicits reasoning in large language models.
\newblock \emph{Advances in neural information processing systems}, 35:\penalty0 24824--24837, 2022.

\bibitem[Lyu et~al.(2023)Lyu, Havaldar, Stein, Zhang, Rao, Wong, Apidianaki, and Callison-Burch]{lyu2023faithful}
Qing Lyu, Shreya Havaldar, Adam Stein, Li~Zhang, Delip Rao, Eric Wong, Marianna Apidianaki, and Chris Callison-Burch.
\newblock Faithful chain-of-thought reasoning.
\newblock In \emph{The 13th International Joint Conference on Natural Language Processing and the 3rd Conference of the Asia-Pacific Chapter of the Association for Computational Linguistics (IJCNLP-AACL 2023)}, 2023.

\bibitem[Xia et~al.(2024)Xia, Wang, Liu, Li, Yu, Chen, McAuley, and Li]{xia2024beyond}
Yu~Xia, Rui Wang, Xu~Liu, Mingyan Li, Tong Yu, Xiang Chen, Julian McAuley, and Shuai Li.
\newblock Beyond chain-of-thought: A survey of chain-of-x paradigms for llms.
\newblock \emph{arXiv preprint arXiv:2404.15676}, 2024.

\bibitem[Zhang et~al.(2024)Zhang, Du, Pang, Liu, Gao, and Lin]{NEURIPS2024_00d80722}
Xuan Zhang, Chao Du, Tianyu Pang, Qian Liu, Wei Gao, and Min Lin.
\newblock Chain of preference optimization: Improving chain-of-thought reasoning in llms.
\newblock In A.~Globerson, L.~Mackey, D.~Belgrave, A.~Fan, U.~Paquet, J.~Tomczak, and C.~Zhang, editors, \emph{Advances in Neural Information Processing Systems}, volume~37, pages 333--356. Curran Associates, Inc., 2024.
\newblock URL \url{https://proceedings.neurips.cc/paper_files/paper/2024/file/00d80722b756de0166523a87805dd00f-Paper-Conference.pdf}.

\bibitem[Chen et~al.(2024)Chen, Qin, Wang, Zhou, and Che]{chen2024unlocking}
Qiguang Chen, Libo Qin, Jiaqi Wang, Jingxuan Zhou, and Wanxiang Che.
\newblock Unlocking the capabilities of thought: A reasoning boundary framework to quantify and optimize chain-of-thought.
\newblock \emph{Advances in Neural Information Processing Systems}, 37:\penalty0 54872--54904, 2024.

\bibitem[Ouyang et~al.(2022)Ouyang, Wu, Jiang, Almeida, Wainwright, Mishkin, Zhang, Agarwal, Slama, Ray, Schulman, Hilton, Kelton, Miller, Simens, Askell, Welinder, Christiano, Leike, and Lowe]{ouyang2022traininglanguagemodelsfollow}
Long Ouyang, Jeff Wu, Xu~Jiang, Diogo Almeida, Carroll~L. Wainwright, Pamela Mishkin, Chong Zhang, Sandhini Agarwal, Katarina Slama, Alex Ray, John Schulman, Jacob Hilton, Fraser Kelton, Luke Miller, Maddie Simens, Amanda Askell, Peter Welinder, Paul Christiano, Jan Leike, and Ryan Lowe.
\newblock Training language models to follow instructions with human feedback, 2022.
\newblock URL \url{https://arxiv.org/abs/2203.02155}.

\bibitem[Yao et~al.(2023{\natexlab{a}})Yao, Yu, Zhao, Shafran, Griffiths, Cao, and Narasimhan]{yao2023treethoughtsdeliberateproblem}
Shunyu Yao, Dian Yu, Jeffrey Zhao, Izhak Shafran, Thomas~L. Griffiths, Yuan Cao, and Karthik Narasimhan.
\newblock Tree of thoughts: Deliberate problem solving with large language models, 2023{\natexlab{a}}.
\newblock URL \url{https://arxiv.org/abs/2305.10601}.

\bibitem[Wang et~al.(2023)Wang, Wei, Schuurmans, Le, Chi, Narang, Chowdhery, and Zhou]{wang2023selfconsistencyimproveschainthought}
Xuezhi Wang, Jason Wei, Dale Schuurmans, Quoc Le, Ed~Chi, Sharan Narang, Aakanksha Chowdhery, and Denny Zhou.
\newblock Self-consistency improves chain of thought reasoning in language models, 2023.
\newblock URL \url{https://arxiv.org/abs/2203.11171}.

\bibitem[Yao et~al.(2023{\natexlab{b}})Yao, Zhao, Yu, Du, Shafran, Narasimhan, and Cao]{yao2023reactsynergizingreasoningacting}
Shunyu Yao, Jeffrey Zhao, Dian Yu, Nan Du, Izhak Shafran, Karthik Narasimhan, and Yuan Cao.
\newblock React: Synergizing reasoning and acting in language models, 2023{\natexlab{b}}.
\newblock URL \url{https://arxiv.org/abs/2210.03629}.

\bibitem[Gao et~al.(2023)Gao, Xiong, Gao, Jia, Pan, Bi, Dai, Sun, Wang, and Wang]{gao2023retrieval}
Yunfan Gao, Yun Xiong, Xinyu Gao, Kangxiang Jia, Jinliu Pan, Yuxi Bi, Yixin Dai, Jiawei Sun, Haofen Wang, and Haofen Wang.
\newblock Retrieval-augmented generation for large language models: A survey.
\newblock \emph{arXiv preprint arXiv:2312.10997}, 2:\penalty0 1, 2023.

\bibitem[Shazeer et~al.(2017)Shazeer, Mirhoseini, Maziarz, Davis, Le, Hinton, and Dean]{shazeer2017outrageously}
Noam Shazeer, Azalia Mirhoseini, Krzysztof Maziarz, Andy Davis, Quoc Le, Geoffrey Hinton, and Jeff Dean.
\newblock Outrageously large neural networks: The sparsely-gated mixture-of-experts layer.
\newblock \emph{arXiv preprint arXiv:1701.06538}, 2017.

\bibitem[Bi et~al.(2024)Bi, Zhang, Jiang, Deng, Zheng, and Chen]{bi2024program}
Zhen Bi, Ningyu Zhang, Yinuo Jiang, Shumin Deng, Guozhou Zheng, and Huajun Chen.
\newblock When do program-of-thought works for reasoning?
\newblock In \emph{Proceedings of the AAAI Conference on Artificial Intelligence}, volume~38, pages 17691--17699, 2024.

\bibitem[Liu et~al.(2024{\natexlab{a}})Liu, Li, and Yin]{liu2024travel}
Tianming Liu, Manzi Li, and Yafeng Yin.
\newblock Can large language models capture human travel behavior? evidence and insights on mode choice.
\newblock \emph{arXiv preprint arXiv:2408.12345}, 2024{\natexlab{a}}.

\bibitem[Mo et~al.(2023{\natexlab{a}})Mo, Xu, Zhuang, Ma, Guo, and Zhao]{mo2023llmtravel}
Baichuan Mo, Hanyong Xu, Dingyi Zhuang, Ruoyun Ma, Xiaotong Guo, and Jinhua Zhao.
\newblock Large language models for travel behavior prediction.
\newblock \emph{arXiv preprint arXiv:2312.00819}, 2023{\natexlab{a}}.

\bibitem[Gallegos et~al.(2024)Gallegos, Rossi, Barrow, Tanjim, Yu, Deilamsalehy, Zhang, Kim, and Dernoncourt]{gallegos2024self}
Isabel~O Gallegos, Ryan~A Rossi, Joe Barrow, Md~Mehrab Tanjim, Tong Yu, Hanieh Deilamsalehy, Ruiyi Zhang, Sungchul Kim, and Franck Dernoncourt.
\newblock Self-debiasing large language models: Zero-shot recognition and reduction of stereotypes.
\newblock \emph{arXiv preprint arXiv:2402.01981}, 2024.

\bibitem[Coletta et~al.(2024)Coletta, Zaslavsky, and et~al.]{coletta2024personallm}
Maddalena Coletta, Nolan Zaslavsky, and et~al.
\newblock Personas improve behavioral alignment of large language models.
\newblock \emph{arXiv preprint arXiv:2403.11568}, 2024.

\bibitem[Liu et~al.(2025)Liu, Fu, Yogatama, and Neiswanger]{liu2025dellma}
Ollie Liu, Deqing Fu, Dani Yogatama, and Willie Neiswanger.
\newblock De{LLM}a: Decision making under uncertainty with large language models.
\newblock In \emph{The Thirteenth International Conference on Learning Representations}, 2025.
\newblock URL \url{https://openreview.net/forum?id=Acvo2RGSCy}.

\bibitem[Lu et~al.(2025)Lu, Hu, Foroosh, Jin, and Liu]{lu-etal-2025-strux}
Yiming Lu, Yebowen Hu, Hassan Foroosh, Wei Jin, and Fei Liu.
\newblock {STRUX}: An {LLM} for decision-making with structured explanations.
\newblock In Luis Chiruzzo, Alan Ritter, and Lu~Wang, editors, \emph{Proceedings of the 2025 Conference of the Nations of the Americas Chapter of the Association for Computational Linguistics: Human Language Technologies (Volume 2: Short Papers)}, pages 131--141, Albuquerque, New Mexico, April 2025. Association for Computational Linguistics.
\newblock ISBN 979-8-89176-190-2.
\newblock \doi{10.18653/v1/2025.naacl-short.11}.
\newblock URL \url{https://aclanthology.org/2025.naacl-short.11/}.

\bibitem[Li et~al.(2023)Li, Mellou, Zhang, Pathuri, and Menache]{li2023large}
Beibin Li, Konstantina Mellou, Bo~Zhang, Jeevan Pathuri, and Ishai Menache.
\newblock Large language models for supply chain optimization.
\newblock \emph{CoRR}, 2023.

\bibitem[Mao et~al.(2024)Mao, Ye, Qian, Pavone, and Wang]{mao2024a}
Jiageng Mao, Junjie Ye, Yuxi Qian, Marco Pavone, and Yue Wang.
\newblock A language agent for autonomous driving.
\newblock In \emph{First Conference on Language Modeling}, 2024.
\newblock URL \url{https://openreview.net/forum?id=UPE6WYE8vg}.

\bibitem[Benary et~al.(2023)Benary, Wang, Schmidt, Soll, Hilfenhaus, Nassir, Sigler, Kn{\"o}dler, Keller, Beule, et~al.]{benary2023leveraging}
Manuela Benary, Xing~David Wang, Max Schmidt, Dominik Soll, Georg Hilfenhaus, Mani Nassir, Christian Sigler, Maren Kn{\"o}dler, Ulrich Keller, Dieter Beule, et~al.
\newblock Leveraging large language models for decision support in personalized oncology.
\newblock \emph{JAMA Network Open}, 6\penalty0 (11):\penalty0 e2343689--e2343689, 2023.

\bibitem[Manski(1977)]{manski1977structure}
Charles~F Manski.
\newblock The structure of random utility models.
\newblock \emph{Theory and decision}, 8\penalty0 (3):\penalty0 229, 1977.

\bibitem[McFadden(1972)]{mcfadden1972conditional}
Daniel McFadden.
\newblock Conditional logit analysis of qualitative choice behavior.
\newblock 1972.

\bibitem[Evans and Viscusi(1991)]{evans1991estimation}
William~N Evans and W~Kip Viscusi.
\newblock Estimation of state-dependent utility functions using survey data.
\newblock \emph{The Review of Economics and Statistics}, pages 94--104, 1991.

\bibitem[B{\"a}ck and Schwefel(1993)]{back1993overview}
Thomas B{\"a}ck and Hans-Paul Schwefel.
\newblock An overview of evolutionary algorithms for parameter optimization.
\newblock \emph{Evolutionary computation}, 1\penalty0 (1):\penalty0 1--23, 1993.

\bibitem[Team et~al.(2023)Team, Anil, Borgeaud, Alayrac, Yu, Soricut, Schalkwyk, Dai, Hauth, Millican, et~al.]{team2023gemini}
Gemini Team, Rohan Anil, Sebastian Borgeaud, Jean-Baptiste Alayrac, Jiahui Yu, Radu Soricut, Johan Schalkwyk, Andrew~M Dai, Anja Hauth, Katie Millican, et~al.
\newblock Gemini: a family of highly capable multimodal models.
\newblock \emph{arXiv preprint arXiv:2312.11805}, 2023.

\bibitem[Yuksekgonul et~al.(2025)Yuksekgonul, Bianchi, Boen, Liu, Lu, Huang, Guestrin, and Zou]{yuksekgonul2025optimizing}
Mert Yuksekgonul, Federico Bianchi, Joseph Boen, Sheng Liu, Pan Lu, Zhi Huang, Carlos Guestrin, and James Zou.
\newblock Optimizing generative ai by backpropagating language model feedback.
\newblock \emph{Nature}, 639\penalty0 (8055):\penalty0 609--616, 2025.

\bibitem[Bierlaire et~al.(2001)Bierlaire, Axhausen, and Abay]{swissmetro}
Michel Bierlaire, Kay Axhausen, and Georg Abay.
\newblock The acceptance of modal innovation: The case of swissmetro.
\newblock 01 2001.

\bibitem[Pham et~al.(2022)Pham, Jiang, and Zhang]{pham2022causality}
Huy Pham, Xuan Jiang, and Cong Zhang.
\newblock Causality and advanced models in trip mode prediction: Interest in choosing swissmetro.
\newblock 2022.

\bibitem[Wen and Chen(2025)]{wen2025new}
Xuli Wen and Xin Chen.
\newblock A new breakthrough in travel behavior modeling using deep learning: A high-accuracy prediction method based on a cnn.
\newblock \emph{Sustainability}, 17\penalty0 (2):\penalty0 738, 2025.

\bibitem[Ghorbani et~al.(2025)Ghorbani, Nassir, Lavieri, Beeramoole, and Paz]{ghorbani2025enhanced}
Amir Ghorbani, Neema Nassir, Patricia~Sauri Lavieri, Prithvi~Bhat Beeramoole, and Alexander Paz.
\newblock Enhanced utility estimation algorithm for discrete choice models in travel demand forecasting.
\newblock \emph{Transportation}, pages 1--28, 2025.

\bibitem[Haj-Yahia et~al.(2025)Haj-Yahia, Mansour, and Toledo]{haj2025incorporating}
Shadi Haj-Yahia, Omar Mansour, and Tomer Toledo.
\newblock Incorporating domain knowledge in deep neural networks for discrete choice models.
\newblock \emph{Transportation Research Part C: Emerging Technologies}, 171:\penalty0 105014, 2025.

\bibitem[Lazarus et~al.(2023)Lazarus, Wyka, White, Picchio, Gostin, Larson, Rabin, Ratzan, Kamarulzaman, and El-Mohandes]{lazarus2023survey}
Jeffrey~V Lazarus, Katarzyna Wyka, Trenton~M White, Camila~A Picchio, Lawrence~O Gostin, Heidi~J Larson, Kenneth Rabin, Scott~C Ratzan, Adeeba Kamarulzaman, and Ayman El-Mohandes.
\newblock A survey of covid-19 vaccine acceptance across 23 countries in 2022.
\newblock \emph{Nature medicine}, 29\penalty0 (2):\penalty0 366--375, 2023.

\bibitem[Croson and Gneezy(2009)]{10.1257/jel.47.2.448}
Rachel Croson and Uri Gneezy.
\newblock Gender differences in preferences.
\newblock \emph{Journal of Economic Literature}, 47\penalty0 (2):\penalty0 448–74, June 2009.
\newblock \doi{10.1257/jel.47.2.448}.
\newblock URL \url{https://www.aeaweb.org/articles?id=10.1257/jel.47.2.448}.

\bibitem[Tymula et~al.(2012)Tymula, Belmaker, Roy, Ruderman, Manson, Glimcher, and Levy]{doi:10.1073/pnas.1207144109}
Agnieszka Tymula, Lior A.~Rosenberg Belmaker, Amy~K. Roy, Lital Ruderman, Kirk Manson, Paul~W. Glimcher, and Ifat Levy.
\newblock Adolescents’ risk-taking behavior is driven by tolerance to ambiguity.
\newblock \emph{Proceedings of the National Academy of Sciences}, 109\penalty0 (42):\penalty0 17135--17140, 2012.
\newblock \doi{10.1073/pnas.1207144109}.
\newblock URL \url{https://www.pnas.org/doi/abs/10.1073/pnas.1207144109}.

\bibitem[de~Bruijn and Antonides(2021)]{deBruijn2021PovertyAE}
Ernst-Jan de~Bruijn and Gerrit Antonides.
\newblock Poverty and economic decision making: a review of scarcity theory.
\newblock \emph{Theory and Decision}, 92:\penalty0 5 -- 37, 2021.
\newblock URL \url{https://api.semanticscholar.org/CorpusID:233665419}.

\bibitem[Kojima et~al.(2022)Kojima, Gu, Reid, Matsuo, and Iwasawa]{kojima2022large}
Takeshi Kojima, Shixiang~Shane Gu, Machel Reid, Yutaka Matsuo, and Yusuke Iwasawa.
\newblock Large language models are zero-shot reasoners.
\newblock \emph{Advances in neural information processing systems}, 35:\penalty0 22199--22213, 2022.

\bibitem[Mo et~al.(2023{\natexlab{b}})Mo, Xu, Zhuang, Ma, Guo, and Zhao]{mo2023large}
Baichuan Mo, Hanyong Xu, Dingyi Zhuang, Ruoyun Ma, Xiaotong Guo, and Jinhua Zhao.
\newblock Large language models for travel behavior prediction.
\newblock \emph{arXiv preprint arXiv:2312.00819}, 2023{\natexlab{b}}.

\bibitem[Brown et~al.(2020)Brown, Mann, Ryder, Subbiah, Kaplan, Dhariwal, Neelakantan, Shyam, Sastry, Askell, et~al.]{brown2020language}
Tom Brown, Benjamin Mann, Nick Ryder, Melanie Subbiah, Jared~D Kaplan, Prafulla Dhariwal, Arvind Neelakantan, Pranav Shyam, Girish Sastry, Amanda Askell, et~al.
\newblock Language models are few-shot learners.
\newblock \emph{Advances in neural information processing systems}, 33:\penalty0 1877--1901, 2020.

\bibitem[Liu et~al.(2024{\natexlab{b}})Liu, Li, and Yin]{liu2024can}
Tianming Liu, Manzi Li, and Yafeng Yin.
\newblock Can large language models capture human travel behavior? evidence and insights on mode choice.
\newblock \emph{Evidence and Insights on Mode Choice (August 26, 2024)}, 2024{\natexlab{b}}.

\bibitem[Huang et~al.(2024)Huang, Gong, Lei, Wang, Zhu, and Zhang]{huang2024comparative}
Pengpeng Huang, Lei Gong, Tian Lei, Jia Wang, Cheng Zhu, and Zheng Zhang.
\newblock A comparative study of mnl and machine learning methods for travel mode choice of medical travel.
\newblock In \emph{International Conference on Traffic and Transportation Studies}, pages 305--313. Springer, 2024.

\bibitem[Truong and Nguyen(2021)]{truong2021travel}
Thi My~Thanh Truong and Thi Cam~Van Nguyen.
\newblock Travel time attractiveness in motorcycle dominated cities: An investigation of university students’ travel behavior.
\newblock In \emph{CIGOS 2021, Emerging Technologies and Applications for Green Infrastructure: Proceedings of the 6th International Conference on Geotechnics, Civil Engineering and Structures}, pages 1723--1731. Springer, 2021.

\bibitem[Wu et~al.(2025)Wu, Yildirimoglu, and Zheng]{wu2025evaluating}
Yikang Wu, Mehmet Yildirimoglu, and Zuduo Zheng.
\newblock Evaluating electric micro-mobility related mode choice stated preferences: A latent class choice approach.
\newblock \emph{arXiv preprint arXiv:2504.01237}, 2025.

\bibitem[Kashifi et~al.(2022)Kashifi, Jamal, Kashefi, Almoshaogeh, and Rahman]{kashifi2022predicting}
Mohammad~Tamim Kashifi, Arshad Jamal, Mohammad~Samim Kashefi, Meshal Almoshaogeh, and Syed~Masiur Rahman.
\newblock Predicting the travel mode choice with interpretable machine learning techniques: A comparative study.
\newblock \emph{Travel Behaviour and Society}, 29:\penalty0 279--296, 2022.

\bibitem[Goodfellow et~al.(2016)Goodfellow, Bengio, and Courville]{goodfellow2016deep}
Ian Goodfellow, Yoshua Bengio, and Aaron Courville.
\newblock \emph{Deep Learning}.
\newblock MIT Press, 2016.
\newblock URL \url{https://www.deeplearningbook.org}.

\bibitem[Arik and Pfister(2021)]{arik2021tabnet}
Sercan~{\"O} Arik and Tomas Pfister.
\newblock Tabnet: Attentive interpretable tabular learning.
\newblock In \emph{Proceedings of the AAAI conference on artificial intelligence}, volume~35, pages 6679--6687, 2021.

\bibitem[Devlin et~al.(2019)Devlin, Chang, Lee, and Toutanova]{devlin2019bert}
Jacob Devlin, Ming-Wei Chang, Kenton Lee, and Kristina Toutanova.
\newblock Bert: Pre-training of deep bidirectional transformers for language understanding, 2019.

\bibitem[Shires and {de Jong}(2009)]{SHIRES2009315}
J.D. Shires and G.C. {de Jong}.
\newblock An international meta-analysis of values of travel time savings.
\newblock \emph{Evaluation and Program Planning}, 32\penalty0 (4):\penalty0 315--325, 2009.
\newblock ISSN 0149-7189.
\newblock \doi{https://doi.org/10.1016/j.evalprogplan.2009.06.010}.
\newblock URL \url{https://www.sciencedirect.com/science/article/pii/S0149718909000548}.
\newblock Evaluating the Impact of Transport Projects: Lessons for Other Disciplines.

\bibitem[Abrantes and Wardman(2011)]{abrantes2011}
Pedro Abrantes and Mark Wardman.
\newblock Meta-analysis of uk values of travel time: An update.
\newblock \emph{Transportation Research Part A: Policy and Practice}, 45:\penalty0 1--17, 01 2011.
\newblock \doi{10.1016/j.tra.2010.08.003}.

\bibitem[Chang(2013)]{chang2013}
Yu-Chun Chang.
\newblock Factors affecting airport access mode choice for elderly air passengers.
\newblock \emph{Transportation Research Part E: Logistics and Transportation Review}, 57\penalty0 (C):\penalty0 105--112, None 2013.
\newblock \doi{10.1016/j.tre.2013.01.010}.
\newblock URL \url{https://ideas.repec.org/a/eee/transe/v57y2013icp105-112.html}.

\bibitem[Weis et~al.(2010)Weis, Axhausen, Schlich, and Zbinden]{weis2010}
Claude Weis, Kay~W. Axhausen, Robert Schlich, and René Zbinden.
\newblock Models of mode choice and mobility tool ownership beyond 2008 fuel prices.
\newblock \emph{Transportation Research Record}, 2157\penalty0 (1):\penalty0 86--94, 2010.
\newblock \doi{10.3141/2157-11}.

\bibitem[Frank et~al.(2007)Frank, Bradley, Kavage, Chapman, and Lawton]{Frank2008}
Lawrence Frank, Mark Bradley, Sarah Kavage, James Chapman, and T.~Lawton.
\newblock Urban form, travel time, and cost relationships with tour complexity and mode choice.
\newblock \emph{Transportation}, 35:\penalty0 37--54, 11 2007.
\newblock \doi{10.1007/s11116-007-9136-6}.

\bibitem[Liao et~al.(2020)Liao, Gil, Pereira, Yeh, and Verendel]{Liao2020}
Yuan Liao, Jorge Gil, Rafael Pereira, Sonia Yeh, and Vilhelm Verendel.
\newblock Disparities in travel times between car and transit: Spatiotemporal patterns in cities.
\newblock \emph{Scientific Reports}, 10, 03 2020.
\newblock \doi{10.1038/s41598-020-61077-0}.

\bibitem[Marques et~al.(2025)Marques, Gomes, Ferreira, Rebuá, and Marques]{Marques2025}
Jorge Marques, Sofia Gomes, Mónica Ferreira, Marina Rebuá, and Hugo Marques.
\newblock Generation z and travel motivations: The impact of age, gender, and residence.
\newblock \emph{Tourism and Hospitality}, 6\penalty0 (2), 2025.
\newblock ISSN 2673-5768.
\newblock \doi{10.3390/tourhosp6020082}.
\newblock URL \url{https://www.mdpi.com/2673-5768/6/2/82}.

\bibitem[Green(2023)]{green2023}
Manfred Green.
\newblock Rational and irrational vaccine hesitancy.
\newblock \emph{Israel Journal of Health Policy Research}, 12, 03 2023.
\newblock \doi{10.1186/s13584-023-00560-1}.

\bibitem[Noh et~al.(2022)Noh, Kim, Yoon, Choe, Choe, Jung, Lee, and Shin]{noh2022}
Yunha Noh, Ju~Hwan Kim, Dongwon Yoon, Young~June Choe, Seung-Ah Choe, Jaehun Jung, Sang-Won Lee, and Ju-Young Shin.
\newblock Predictors of covid-19 booster vaccine hesitancy among fully vaccinated adults in korea: a nationwide cross-sectional survey.
\newblock \emph{Epidemiology and Health}, 44, 2022.
\newblock URL \url{https://api.semanticscholar.org/CorpusID:251255072}.

\bibitem[Biswas et~al.(2021)Biswas, Mustapha, Khubchandani, and Price]{biswas2021}
Nirbachita Biswas, Toheeb Mustapha, Jagdish Khubchandani, and James~H. Price.
\newblock The nature and extent of covid-19 vaccination hesitancy in healthcare workers.
\newblock \emph{Journal of Community Health}, 46\penalty0 (6):\penalty0 1244--1251, December 2021.
\newblock \doi{10.1007/s10900-021-00984-3}.
\newblock Epub 2021 Apr 20.

\bibitem[Trent et~al.(2022)Trent, Seale, Chughtai, Salmon, and MacIntyre]{trent2022}
Mallory Trent, Holly Seale, Abrar~Ahmad Chughtai, Daniel Salmon, and C.~Raina MacIntyre.
\newblock Trust in government, intention to vaccinate and covid-19 vaccine hesitancy: A comparative survey of five large cities in the united states, united kingdom, and australia.
\newblock \emph{Vaccine}, 40\penalty0 (17):\penalty0 2498--2505, 2022.
\newblock ISSN 0264-410X.
\newblock \doi{https://doi.org/10.1016/j.vaccine.2021.06.048}.
\newblock URL \url{https://www.sciencedirect.com/science/article/pii/S0264410X21007982}.
\newblock Pandemic Simulation, Pacific Eclipse.

\bibitem[Glanz et~al.(2017)Glanz, Wagner, Narwaney, Kraus, Shoup, Xu, O’Leary, Omer, Gleason, and Daley]{Glanz2017}
Jason Glanz, Nicole Wagner, Komal Narwaney, Courtney Kraus, Jo~Shoup, Stanley Xu, Sean O’Leary, Saad Omer, Kathy Gleason, and Matthew Daley.
\newblock Web-based social media intervention to increase vaccine acceptance: A randomized controlled trial.
\newblock \emph{Pediatrics}, 140:\penalty0 e20171117, 11 2017.
\newblock \doi{10.1542/peds.2017-1117}.

\bibitem[Lohmann and Albarracin(2022)]{Lohmann2022}
Sophie Lohmann and Dolores Albarracin.
\newblock Trust in the public health system as a source of information on vaccination matters most when environments are supportive.
\newblock \emph{Vaccine}, 40, 06 2022.
\newblock \doi{10.1016/j.vaccine.2022.06.012}.

\bibitem[Bajos et~al.(2022)Bajos, Spire, Silberzan, Sireyjol, Jusot, Meyer, Pousson, and Warszawski]{Bajos2022}
Nathalie Bajos, Alexis Spire, Léna Silberzan, Antoine Sireyjol, Florence Jusot, Laurence Meyer, Jeanna-Eve Pousson, and Josiane Warszawski.
\newblock When lack of trust in the government and in scientists reinforces social inequalities in vaccination against covid-19.
\newblock \emph{Frontiers in Public Health}, 10:\penalty0 908152, 07 2022.
\newblock \doi{10.3389/fpubh.2022.908152}.

\end{thebibliography}





\newpage
\section*{NeurIPS Paper Checklist}

\begin{enumerate}

\item {\bf Claims}
    \item[] Question: Do the main claims made in the abstract and introduction accurately reflect the paper's contributions and scope?
    \item[] Answer: \answerYes{} 
    \item[] Justification: Please refer to Section \ref{method}.

\item {\bf Limitations}
    \item[] Question: Does the paper discuss the limitations of the work performed by the authors?
    \item[] Answer: \answerYes{} 
    \item[] Justification: Please refer to Section \ref{conclusion}.

\item {\bf Theory assumptions and proofs}
    \item[] Question: For each theoretical result, does the paper provide the full set of assumptions and a complete (and correct) proof?
    \item[] Answer: \answerYes{} 
    \item[] Justification: Yes, we have full proof and assumption for each result. Please refer to Section \ref{method},\ref{exp}.

    \item {\bf Experimental result reproducibility}
    \item[] Question: Does the paper fully disclose all the information needed to reproduce the main experimental results of the paper to the extent that it affects the main claims and/or conclusions of the paper (regardless of whether the code and data are provided or not)?
    \item[] Answer: \answerYes{} 
    \item[] Justification: Yes, all the details are illustrated in Section \ref{method} and Appendix. We also have released all the code and experiment raw results on GitHub. While the experiments are reproducible given the released code and data, all reported results are from single runs. Multi-seed repetitions will be conducted in future work to assess the stability of the reported performance.

\item {\bf Open access to data and code}
    \item[] Question: Does the paper provide open access to the data and code, with sufficient instructions to faithfully reproduce the main experimental results, as described in supplemental material?
    \item[] Answer: \answerYes{} 
    \item[] Justification: Yes, the code and data will be published on GitHub upon acceptance.

\item {\bf Experimental setting/details}
    \item[] Question: Does the paper specify all the training and test details (e.g., data splits, hyperparameters, how they were chosen, type of optimizer, etc.) necessary to understand the results?
    \item[] Answer: \answerYes{} 
    \item[] Justification: Please refer to Section \ref{exp}.

\item {\bf Experiment statistical significance}
    \item[] Question: Does the paper report error bars suitably and correctly defined or other appropriate information about the statistical significance of the experiments?
    \item[] Answer: \answerNo{} 
    \item[] Justification: We do not report error bars due to the high cost of repeated API calls. We plan to incorporate more thorough uncertainty quantification in future work as computational resources allow.

\item {\bf Experiments compute resources}
    \item[] Question: For each experiment, does the paper provide sufficient information on the computer resources (type of compute workers, memory, time of execution) needed to reproduce the experiments?
    \item[] Answer: \answerYes{} 
    \item[] Justification: Please refer to Appendix.
    
\item {\bf Code of ethics}
    \item[] Question: Does the research conducted in the paper conform, in every respect, with the NeurIPS Code of Ethics \url{https://neurips.cc/public/EthicsGuidelines}?
    \item[] Answer: \answerYes{} 
    \item[] Justification: We follow the NeurIPS Code of Ethics.

\item {\bf Broader impacts}
    \item[] Question: Does the paper discuss both potential positive societal impacts and negative societal impacts of the work performed?
    \item[] Answer: \answerYes{} 
    \item[] Justification: Yes, please refer to Section \ref{intro} and Section \ref{conclusion}.
    
\item {\bf Safeguards}
    \item[] Question: Does the paper describe safeguards that have been put in place for responsible release of data or models that have a high risk for misuse (e.g., pretrained language models, image generators, or scraped datasets)?
    \item[] Answer: \answerNA{} 
    \item[] Justification: The paper poses no such risks.

\item {\bf Licenses for existing assets}
    \item[] Question: Are the creators or original owners of assets (e.g., code, data, models), used in the paper, properly credited and are the license and terms of use explicitly mentioned and properly respected?
    \item[] Answer: \answerYes{} 
    \item[] Justification: All assets used are correctly cited.

\item {\bf New assets}
    \item[] Question: Are new assets introduced in the paper well documented and is the documentation provided alongside the assets?
    \item[] Answer: \answerYes{} 
    \item[] Justification: We release the full implementation of our framework. All assets will be made publicly available upon publication.

\item {\bf Crowdsourcing and research with human subjects}
    \item[] Question: For crowdsourcing experiments and research with human subjects, does the paper include the full text of instructions given to participants and screenshots, if applicable, as well as details about compensation (if any)? 
    \item[] Answer: \answerNA{} 
    \item[] Justification: The paper does not involve crowdsourcing nor research with human subjects.

\item {\bf Institutional review board (IRB) approvals or equivalent for research with human subjects}
    \item[] Question: Does the paper describe potential risks incurred by study participants, whether such risks were disclosed to the subjects, and whether Institutional Review Board (IRB) approvals (or an equivalent approval/review based on the requirements of your country or institution) were obtained?
    \item[] Answer: \answerNA{} 
    \item[] Justification: The study uses publicly available datasets involving human subjects. Based on the dataset documentation, the original data collection was conducted under appropriate ethical review and consent procedures.

\item {\bf Declaration of LLM usage}
    \item[] Question: Does the paper describe the usage of LLMs if it is an important, original, or non-standard component of the core methods in this research? Note that if the LLM is used only for writing, editing, or formatting purposes and does not impact the core methodology, scientific rigorousness, or originality of the research, declaration is not required.
    \item[] Answer: \answerYes{} 
    \item[] Justification: LLMs are used in both stages of our method: to generate candidate symbolic utility expressions during the group-level modeling phase, and to initialize individual-level templates during semantic adaptation. 

\end{enumerate}

\newpage

\appendix

\section{Extended Analysis}

\subsection{Broader Societal Impacts}
The introduction of \textsc{\model}, a two-stage framework that first discovers group-level symbolic utility functions and then develops individual semantic templates, offers potential for positive societal contributions. By combining LLM reasoning with symbolic regression, \textsc{\model} aims to deliver: \textbf{(1) More inclusive public-policy insights:} interpretable utility functions reveal group motivations behind choices in socially critical behaviors, enabling better targeted interventions and policy making. \textbf{(2) Precise and equitable individual adaptation:} reduce the "one-size-fits-all" errors in some high-stake domains, like healthcare and education. 

At the same time, \textsc{\model} also brings concerns. (1) Potential Misuse: Because \textsc{\model} can model fine-grained individual decisions, it could be deployed for manipulative advertising, political micro-targeting, or discriminatory dynamic pricing. Mitigations include restricted licensing and mandatory human oversight for high-impact deployments. (2) Bias and Fairness: The symbolic utility discovery stage may suffer from coarse group partitioning, which can ignore intra-group heterogeneity. Meanwhile, the semantic adaptation stage may inherit biases presented in the LLM’s training data, thereby embedding historical stereotypes.

\subsection{Qualitative Analysis and Case Study}
\label{casestudy}

\begingroup
  \setlist{nolistsep}
  \renewcommand{\arraystretch}{1.3}
  \setlength{\tabcolsep}{6pt}
  \emergencystretch=2em

    \begin{longtable}{>{\raggedright\arraybackslash}p{0.12\textwidth} >{\raggedright\arraybackslash}p{0.88\textwidth}}
    \caption{\textbf{Qualitative case study of \textsc{\model} on the \textit{Swissmetro} dataset.}
    This table contrasts four representative travelers’ attributes, the candidate alternatives,
    the group-level symbolic utility functions learned in Stage-1, and the individualized
    decision rules refined in Stage-2.}\\
    
    \arrayrulecolor{blue}\hline
    \endfirsthead
    
    \arrayrulecolor{blue}\hline
    \endhead


      \arrayrulecolor{blue}\hline
      \rowcolor{lightBlue}
      \textbf{\textcolor{blue}{\textit{Swissmetro} Case 1}} & \\ \hline

      Features $\mathcal{X}$ &
      \begin{minipage}[t]{\linewidth}\raggedright\sloppy
        \begin{itemize}
          \item[-] \textbf{Age}: between 54 and 65 years old
          \item[-] \textbf{Gender}: Male
          \item[-] \textbf{Income}: Over 100
          \item[-] \textbf{Trip Purpose}: business
          \item[-] \textbf{Luggage}: none of luggage
          \item[-] \textbf{Payment Method}: paid by oneself
          \item[-] \textbf{Origin}: St. Gallen
          \item[-] \textbf{Destination}: Bern
          \vspace{0.4em} 
        \end{itemize}
      \end{minipage}
      \\

      \arrayrulecolor{mygray}\hline

      Alternatives $\mathcal{J}$ &
      \begin{minipage}[t]{\linewidth}\raggedright\sloppy
        \begin{itemize}
          \item[-] \textbf{Metro}: travel time of 77 minutes, costing 74 CHF, with a headway of 30 minutes
          \item[-] \textbf{Train}: travel time of 120 minutes, costing 64 CHF, with a headway of 120 minutes
          \item[-] \textbf{Car}: travel time of 169 minutes, costing 60 CHF
          \vspace{0.2em} 
        \end{itemize}
      \end{minipage}
      \\ \hline

      Optimal Group Utility $f_g^*$ &
      \begin{minipage}[t]{\linewidth}\raggedright\sloppy
        \begin{itemize}
        \item[-] \textbf{Train}: 
        $K_1 \cdot \text{trip\_purpose}
        + K_2 \cdot \bigl( \sqrt{\text{num\_luggage} + C_1} + C_2 \bigr) \cdot \log\!\bigl( \text{age} + C_3 \bigr)
        + K_3 \cdot \text{time\_train}
        + K_4 \cdot \sqrt{ \text{train\_headway\_min} }
        + K_5 \cdot \bigl( \text{is\_first\_class} + C_4 \bigr) \cdot \sqrt{ \log\!\bigl( \text{income} + C_5 \bigr) }
        + C_6$
        
        \item[-] \textbf{Car}: 
        $K_1 \cdot \text{trip\_purpose}
        + K_2 \cdot \sqrt{ \text{num\_luggage} + C_1 } \cdot \log\!\bigl( \text{age} + C_2 \bigr)
        + K_3 \cdot \text{time\_car} \cdot \sqrt{ \log\!\bigl( \text{income} + C_3 \bigr) }
        + K_4 \cdot \sqrt{ \lvert \text{cost\_train} + C_4 \rvert }
        + K_5 \cdot \text{train\_service\_headway}
        + K_6 \cdot \bigl( \text{is\_first\_class} + C_5 \bigr) \cdot \sqrt{ \log\!\bigl( \text{income} + C_6 \bigr) }
        + C_7$
        
        \item[-] \textbf{Metro}: 
        $K_1 \cdot \text{trip\_purpose}
        + K_2 \cdot \bigl( \text{num\_luggage} + C_1 \bigr) \cdot \sqrt{ \log\!\bigl( \text{age} + C_2 \bigr) }
        + K_3 \cdot \text{time\_sm} \cdot \sqrt{ \text{income} + C_3 }
        + K_4 \cdot \sqrt{ \text{cost\_sm} + C_4 }
        + K_5 \cdot \bigl( \text{ticket\_payer\_type} + C_5 \bigr)
        + C_6$
      \vspace{0.4em} 
        \end{itemize}
      \end{minipage}
      \\ \hline

      Optimal Personalized Decision Rule $\mathcal{P}_i^*$ &
      \begin{minipage}[t]{\linewidth}\raggedright\sloppy
        \begin{itemize}
          \item[-] \textbf{BALANCED}: As a business traveler, my travel decisions are guided by a dynamic scoring system that prioritizes speed, environmental impact, and cost. I often weigh these factors differently based on specific scenarios (e.g., major conferences vs. regular trips). Real-time traffic/weather and reflections on past trips refine preferences; I consider Swissmetro, trains, cars, buses, and rideshares, evaluating comfort, reliability, and amenities.
        \end{itemize}
          \vspace{0.2em} 
      \end{minipage}
      \\
      \arrayrulecolor{blue}\hline
      \rowcolor{lightBlue}
      \textbf{\textcolor{blue}{\textit{Swissmetro} – Case 2}} & \\ \hline

      Features $\mathcal{X}$ &
      \begin{minipage}[t]{\linewidth}\raggedright\sloppy
        \begin{itemize}
          \item[-] \textbf{Age}: between 39 and 54 years old
          \item[-] \textbf{Gender}: Male
          \item[-] \textbf{Income}: between 50 and 100
          \item[-] \textbf{Trip Purpose}: leisure
          \item[-] \textbf{Luggage}: no luggage
          \item[-] \textbf{Payment Method}: paid by oneself
          \item[-] \textbf{Origin}: Graubünden
          \item[-] \textbf{Destination}: Bern
        \end{itemize}
          \vspace{0.4em} 
      \end{minipage}
      \\

      \arrayrulecolor{mygray}\hline

      Alternatives $\mathcal{J}$ &
      \begin{minipage}[t]{\linewidth}\raggedright\sloppy
        \begin{itemize}
          \item[-] \textbf{Metro}: travel time 142 min, cost 123 CHF, headway 30 min
          \item[-] \textbf{Train}: travel time 180 min, cost 97 CHF, headway 60 min
          \item[-] \textbf{Car}: travel time 136 min, cost 149 CHF
        \end{itemize}
          \vspace{0.2em} 
      \end{minipage}
      \\ \hline

      Optimal Group Utility $f_g^*$ &
      \begin{minipage}[t]{\linewidth}\raggedright\sloppy
        \begin{itemize}
        \item[-] \textbf{Train}:
        $K_1\!\cdot\!\bigl(\text{ticket\_payer\_type}\cdot\text{is\_car\_available}\sqrt{\text{age}+C_1}
        +K_2\cdot\text{is\_first\_class}\cdot\log(\text{age}+C_2)\bigr)
        -K_3\!\cdot\!\dfrac{\text{time\_train}}{\text{age}+C_3}
        +K_4\!\cdot\!\bigl(\text{income}
        +\text{num\_luggage}\sqrt{\text{age}+C_4}\bigr)
        -K_5\!\cdot\!\Bigl(\dfrac{\text{time\_car}}{\text{age}+C_5}+C_6\Bigr)+C_7$
        
        \item[-] \textbf{Car}:
        $K_1\!\cdot\!\bigl(\text{ticket\_payer\_type}\cdot\text{is\_car\_available}\sqrt{\text{age}+C_1}
        +K_2\cdot\text{is\_first\_class}\bigr)
        -K_3\!\cdot\!\dfrac{\text{time\_train}}{\text{age}+C_2}
        +K_4\!\cdot\!\log\!\bigl(\text{income}
        +\text{num\_luggage}+K_5\sqrt{\text{age}+C_3}\bigr)
        -K_6\!\cdot\!\Bigl(\dfrac{\text{time\_car}}{\text{age}+C_4}+C_5\Bigr)+C_6$
        
        \item[-] \textbf{Metro}:
        $K_1\!\cdot\!\bigl(\text{sm\_headway\_min}
        -\dfrac{\text{time\_sm}}{\log(\text{age}+C_1)}\bigr)
        +K_2\!\cdot\!\bigl(\text{has\_ga\_travel\_pass}\sqrt{\text{age}+C_2}
        +\text{gender}\cdot\log(\text{income}+\text{num\_luggage}+C_3)\bigr)
        + C_4$
          \vspace{0.4em} 
        \end{itemize}
      \end{minipage}
      \\ \hline

      Optimal Personalized Decision Rule $\mathcal{P}_i^*$ &
      \begin{minipage}[t]{\linewidth}\raggedright\sloppy
        \begin{itemize}
          \item[-] \textbf{BALANCED, INFORMED, SUSTAINABLE, USER-CENTRIC, CONTEXT-ADAPTIVE, COMFORT-FOCUSED}:
          Dynamically trades off time, cost, and scenic enjoyment. Eco-friendliness and comfort (leg-room, quiet cars) matter, even with longer trips or higher prices; prior-trip feedback and real-time context refine recommendations.
        \end{itemize}
          \vspace{0.3em} 
      \end{minipage}
      \\
      \arrayrulecolor{blue}\hline
      \rowcolor{lightBlue}
      \textbf{\textcolor{blue}{\textit{Swissmetro} – Case 3}} & \\ \hline

      Features $\mathcal{X}$ &
      \begin{minipage}[t]{\linewidth}\raggedright\sloppy
        \begin{itemize}
          \item[-] \textbf{Age}: between 39 and 54 years old
          \item[-] \textbf{Gender}: Male
          \item[-] \textbf{Income}: under 50
          \item[-] \textbf{Trip Purpose}: leisure
          \item[-] \textbf{Luggage}: no luggage
          \item[-] \textbf{Payment Method}: paid by unknown people
          \item[-] \textbf{Origin}: Zurich
          \item[-] \textbf{Destination}: Bern
        \end{itemize}
          \vspace{0.4em} 
      \end{minipage}
      \\

      \arrayrulecolor{mygray}\hline

      Alternatives $\mathcal{J}$ &
      \begin{minipage}[t]{\linewidth}\raggedright\sloppy
        \begin{itemize}
          \item[-] \textbf{Metro}: travel time 56 min, cost 42 CHF, headway 10 min
          \item[-] \textbf{Train}: travel time 111 min, cost 36 CHF, headway 30 min
          \item[-] \textbf{Car}: travel time 88 min, cost 60 CHF
        \end{itemize}
          \vspace{0.2em} 
      \end{minipage}
      \\ \hline

      Optimal Group Utility $f_g^*$ &
      \begin{minipage}[t]{\linewidth}\raggedright\sloppy
        \begin{itemize}
        \item[-] \textbf{Train}:
        $C_1 + K_1\!\cdot\!\lvert \text{time\_train} - \text{time\_car} \rvert
        \bigl(\log(\text{income} + C_2) + \sqrt{\text{age} + C_3}\bigr)
        \bigl(\text{gender}\sqrt{\text{num\_luggage} + C_4} + \text{is\_first\_class}\bigr)
        + K_2\!\cdot\!\dfrac{\text{trip\_purpose}}{\sqrt{\text{age} + C_5} + C_6}
        - K_3\!\cdot\!\text{cost\_train}\cdot\text{trip\_purpose}$
        
        \item[-] \textbf{Car}:
        $C_1 + K_1\!\cdot\!\bigl(
        \text{num\_luggage}\cdot\text{trip\_purpose}\cdot
        \lvert \log(\text{income} + C_2) \rvert\cdot
        (C_3 + \sqrt{\text{age} + C_4})
        + \sqrt{\text{age} + C_5}\cdot
        \lvert \text{time\_train} - \text{time\_car} \rvert\cdot
        (C_6 + \text{is\_first\_class})
        \bigr)
        - K_2\!\cdot\!(C_7 + \text{train\_service\_headway\_min})
        + K_3\!\cdot\!(\text{income} + C_8)^{C_9}$
        
        \item[-] \textbf{Metro}:
        $C_1 + K_1\!\cdot\!\biggl(
        \dfrac{\text{time\_sm}^{C_2}\cdot\text{trip\_purpose}\cdot
        \sqrt{\text{num\_luggage} + C_3}\cdot
        \log(\text{income} + C_4)}
        {\sqrt{\text{age} + C_5} + C_6}
        + K_2\!\cdot\!\bigl(
        \text{cost\_sm}\cdot(\log(\text{age} + C_7)
        + \text{is\_first\_class})
        \bigr)\cdot(C_8 + \sqrt{\text{time\_car}})
        \biggr)$
        \end{itemize}
          \vspace{0.4em} 
      \end{minipage}
      \\ \hline

      Optimal Personalized Decision Rule $\mathcal{P}_i^*$ &
      \begin{minipage}[t]{\linewidth}\raggedright\sloppy
        \begin{itemize}
          \item[-] \textbf{COST\_SAVING}: Prefers options at least 10\% cheaper than peak-fare averages, values minimal transfers and amenities (Wi-Fi, food), uses carbon-footprint data, and favors off-peak scheduling; feedback refines future recommendations.
        \end{itemize}
          \vspace{0.2em} 
      \end{minipage}
      \\
      \arrayrulecolor{blue}\hline
      \rowcolor{lightBlue}
      \textbf{\textcolor{blue}{\textit{Swissmetro} – Case 4}} & \\ \hline

      Features $\mathcal{X}$ &
      \begin{minipage}[t]{\linewidth}\raggedright\sloppy
        \begin{itemize}
          \item[-] \textbf{Age}: over 65 years old
          \item[-] \textbf{Gender}: Male
          \item[-] \textbf{Income}: over 100
          \item[-] \textbf{Trip Purpose}: shopping
          \item[-] \textbf{Luggage}: no luggage
          \item[-] \textbf{Payment Method}: paid half-half
          \item[-] \textbf{Origin}: Vaud
          \item[-] \textbf{Destination}: Geneva
        \end{itemize}
          \vspace{0.4em} 
      \end{minipage}
      \\

      \arrayrulecolor{mygray}\hline

      Alternatives $\mathcal{J}$ &
      \begin{minipage}[t]{\linewidth}\raggedright\sloppy
        \begin{itemize}
          \item[-] \textbf{Metro}: travel time 21 min, cost 226 CHF, headway 10 min
          \item[-] \textbf{Train}: travel time 42 min, cost 209 CHF, headway 60 min
          \item[-] \textbf{Car}: travel time 40 min, cost 67 CHF
        \end{itemize}
          \vspace{0.2em} 
      \end{minipage}
      \\ \hline

      Optimal Group Utility $f_g^*$ &
      \begin{minipage}[t]{\linewidth}\raggedright\sloppy
        \begin{itemize}
        \item[-] \textbf{Train}:
        $K_1\!\cdot\!\bigl(\text{time\_train}
        +\text{num\_luggage}\cdot\text{age}^{C_1}
        +\text{age}^{C_2}
        +\text{time\_car}\bigr)
        +K_2\!\cdot\!\bigl(\text{income}\cdot
        \text{is\_first\_class}\cdot\text{gender}\cdot
        \log(\text{age}+C_3)\bigr)
        +\text{num\_luggage}^{C_4}
        + C_5$
        
        \item[-] \textbf{Car}:
        $K_1\!\cdot\!\bigl(\text{time\_car}
        +\text{num\_luggage}\cdot\text{age}^{C_1}
        +\text{income}\sqrt{\text{is\_first\_class}}
        +\text{car\_travel\_cost\_chf}
        +\text{num\_luggage}\cdot
        \exp(\text{age}/C_2)\bigr)
        + C_3$
        
        \item[-] \textbf{Metro}:
        $K_1\!\cdot\!\text{time\_sm}
        +K_2\!\cdot\!\bigl(
        \text{cost\_sm}
        +\text{income}\cdot(\text{is\_first\_class}
        +\text{gender})\cdot\exp(\text{age}^{C_1})
        +\text{num\_luggage}\,\text{age}^{C_2}
        +\text{age}/C_3
        \bigr)
        + C_4$
        \end{itemize}
          \vspace{0.4em} 
      \end{minipage}
      \\ \hline

      Optimal Personalized Decision Rule $\mathcal{P}_i^*$ &
      \begin{minipage}[t]{\linewidth}\raggedright\sloppy
        \begin{itemize}
          \item[-] \textbf{COMFORT\_SEEKING}:
          Prioritizes spacious seating, quiet cars, and onboard services; willing to pay up to 20\% premium. Prefers real-time updates and easy boarding for accessibility; social events may nudge to more social modes; feedback refines future recommendations.
        \end{itemize}
      \end{minipage}
      \\
      \hline
    \end{longtable}
\endgroup


\begingroup
  \setlist{nolistsep}
  \renewcommand{\arraystretch}{1.3}
  \emergencystretch=2em

    \begin{longtable}{>{\raggedright\arraybackslash}p{0.12\textwidth} >{\raggedright\arraybackslash}p{0.88\textwidth}}
    \caption{\textbf{Qualitative case study of \textsc{\model} on the \textit{Vaccine} dataset.}
    This table contrasts four representative individuals' attributes, the candidate alternatives, the group-level symbolic utility functions learned in Stage-1, and the individualized decision rules refined in Stage-2.} \\

    \arrayrulecolor{blue}\hline
    \endfirsthead
    
    \arrayrulecolor{blue}\hline
    \endhead

      \arrayrulecolor{blue}\hline
      \rowcolor{lightBlue}
      \textbf{\textcolor{blue}{\textit{Vaccine} – Case 1}} & \\ \hline

      Features $\mathcal{X}$ &
      \begin{minipage}[t]{\linewidth}\raggedright
        \begin{itemize}
          \item[-] \textbf{Age}: 25
          \item[-] \textbf{Gender}: Male
          \item[-] \textbf{Occupation}: Nurse
          \item[-] \textbf{Education}: No university degree
          \item[-] \textbf{Income}: Above-median
          \item[-] \textbf{COVID-19 Threat Perception}: Moderate
          \item[-] \textbf{Risk Perception}: Disease risk $>$ vaccine risk
          \item[-] \textbf{Trust in Government}: Moderate
          \item[-] \textbf{Trust in Science}: Moderate
          \item[-] \textbf{Perceived Vaccine Safety}: Fairly safe
          \item[-] \textbf{Family COVID Infection}: $>$1 yr ago
          \item[-] \textbf{Attention to Vaccine News}: Increased
        \end{itemize}
          \vspace{0.4em} 
      \end{minipage}
      \\

      \arrayrulecolor{mygray}\hline

      Alternatives $\mathcal{J}$ &
      \begin{minipage}[t]{\linewidth}\raggedright
        \begin{itemize}
          \item[-] \textbf{Unvaccinated}
          \item[-] \textbf{Vaccinated\_No\_Booster}
          \item[-] \textbf{Booster}
        \end{itemize}
          \vspace{0.4em} 
      \end{minipage}
      \\ \hline

      Optimal Group Utility $f_g^*$ &
      \begin{minipage}[t]{\linewidth}\raggedright\sloppy
        \begin{itemize}
        \item[-] \textbf{Unvaccinated}:
        $C_1\!\cdot\!\text{covid\_threat}\bigl(C_2
        +\text{trust\_gov}\cdot\text{trust\_sci}\cdot
        \log(\text{age}+C_3)\bigr)
        \cdot\text{risk\_covid\_gt\_vax}
        +K_1\!\cdot\!\text{family\_covid}\cdot
        \log(\text{age}+C_4)$
        
        \item[-] \textbf{Vaccinated\_No\_Booster}:
        $C_1\!\cdot\!\text{covid\_threat}
        +C_2\!\cdot\!\text{vax\_safe}
        +K_1\!\cdot\!\bigl(
        \text{trust\_gov}\cdot\text{trust\_sci}\cdot
        \text{more\_attention}\,\sqrt{\text{age}+C_3}
        \bigr)$
        
        \item[-] \textbf{Booster}:
        $C_1\!\cdot\!e^{\text{age}^{C_2}}\,
        \text{covid\_threat}\,\sqrt{\text{vax\_protect\_long}}
        +C_3\!\cdot\!\text{vax\_safe}
        +K_1\!\cdot\!\bigl(
        \text{trust\_gov}\cdot\text{trust\_sci}\cdot
        \text{nurse}\cdot\sqrt{\text{age}+C_4}
        \bigr)$
        \end{itemize}
          \vspace{0.4em} 
      \end{minipage}
      \\ \hline

      Optimal Personalized Decision Rule $\mathcal{P}_i^*$ &
      \begin{minipage}[t]{\linewidth}\raggedright\sloppy
        \begin{itemize}
          \item[-] \textbf{TRUSTING\_AUTHORITY}: This persona represents a cautiously informed healthcare worker who values evidence-based guidance and may favor “Vaccinated (No Booster)” given safety concerns, while remaining open to updates as new data emerge; family and social influence are considered.
        \end{itemize}
          \vspace{0.4em} 
      \end{minipage}
      \\

      \arrayrulecolor{blue}\hline
      \rowcolor{lightBlue}
      \textbf{\textcolor{blue}{\textit{Vaccine} – Case 2}} & \\ \hline

      Features $\mathcal{X}$ &
      \begin{minipage}[t]{\linewidth}\raggedright
        \begin{itemize}
          \item[-] \textbf{Age}: 55
          \item[-] \textbf{Gender}: Male
          \item[-] \textbf{Education}: No university degree
          \item[-] \textbf{Income}: Below-median
          \item[-] \textbf{COVID-19 Threat Perception}: Moderate
          \item[-] \textbf{Trust in Government Delivery}: High
          \item[-] \textbf{Trust in Science}: Some
          \item[-] \textbf{Risk Perception}: Disease risk $>$ vaccine risk
          \item[-] \textbf{Family COVID}: None
          \item[-] \textbf{Attention to Vaccine News}: Decreased
        \end{itemize}
          \vspace{0.4em} 
      \end{minipage}
      \\

      \arrayrulecolor{mygray}\hline

      Alternatives $\mathcal{J}$ &
      \begin{minipage}[t]{\linewidth}\raggedright
        \begin{itemize}
          \item[-] \textbf{Unvaccinated}
          \item[-] \textbf{Vaccinated\_No\_Booster}
          \item[-] \textbf{Booster}
          \vspace{0.4em} 
        \end{itemize}
      \end{minipage}
      \\ \hline

      Optimal Group Utility $f_g^*$ &
      \begin{minipage}[t]{\linewidth}\raggedright\sloppy
        \begin{itemize}
        \item[-] \textbf{Unvaccinated}:
        $K_1\!\sqrt{
        \text{covid\_threat}\bigl(
        \text{risk\_covid\_gt\_vax}
        +\text{gender}\sqrt{\text{age}}+C_1
        \bigr)
        }
        \cdot\bigl(
        (\text{trust\_gov}\,\text{trust\_sci})^{C_2}
        + C_3
        \bigr)
        + K_2\!\cdot\!\text{more\_attention}
        - K_3\!\cdot\!\text{low\_income}
        \bigl(
        C_4+\text{has\_degree}\cdot\text{trust\_gov}\cdot\text{trust\_sci}
        \bigr)
        + C_5$
        
        \item[-] \textbf{Vaccinated\_No\_Booster}:
        $K_1\!\cdot\!\bigl(
        \text{vax\_safe}
        +\text{trust\_gov}\cdot\text{trust\_sci}\cdot
        \sqrt{
        \sqrt{\text{age}+C_1}
        +\text{income\_unknown}
        +C_2
        }
        + C_3
        \bigr)
        + K_2\!\cdot\!\dfrac{
        \text{more\_attention}
        }{
        \text{less\_attention}+C_4
        }
        + C_5$
        
        \item[-] \textbf{Booster}:
        $K_1\!\cdot\!\bigl(
        \text{family\_covid}
        +\text{physician}\cdot
        \text{trust\_gov}\,\text{trust\_sci}\,
        (\sqrt{\text{age}+C_1}+C_2)
        +\text{nurse}\cdot
        \text{trust\_sci}\sqrt{\text{age}+C_3}
        + C_4
        \bigr)
        + C_5$
        \end{itemize}
          \vspace{0.4em} 
      \end{minipage}
      \\ \hline

      Optimal Personalized Decision Rule $\mathcal{P}_i^*$ &
      \begin{minipage}[t]{\linewidth}\raggedright\sloppy
        \begin{itemize}
          \item[-] \textbf{SKEPTICAL}: Prefers conservative choices due to perceived safety concerns; may remain unvaccinated unless convinced by trusted figures; open to “Vaccinated\_No\_Booster” or “Booster” if necessity and safety are clearly established.
        \end{itemize}
      \end{minipage}
      \\

      \arrayrulecolor{blue}\hline
      \rowcolor{lightBlue}
      \textbf{\textcolor{blue}{\textit{Vaccine} – Case 3}} & \\ \hline

      Features $\mathcal{X}$ &
      \begin{minipage}[t]{\linewidth}\raggedright
        \begin{itemize}
          \item[-] \textbf{Age}: 86
          \item[-] \textbf{Gender}: Male
          \item[-] \textbf{Education}: No university degree
          \item[-] \textbf{Income}: Below-median
          \item[-] \textbf{COVID-19 Threat View}: Moderate
          \item[-] \textbf{Perceived Vaccine Safety}: High
          \item[-] \textbf{Long-COVID Protection Belief}: Uncertain
          \item[-] \textbf{Family COVID}: None
          \item[-] \textbf{Attention to Vaccine News}: Increased
        \end{itemize}
          \vspace{0.4em} 
      \end{minipage}
      \\

      \arrayrulecolor{mygray}\hline

      Alternatives $\mathcal{J}$ &
      \begin{minipage}[t]{\linewidth}\raggedright
        \begin{itemize}
          \item[-] \textbf{Unvaccinated}
          \item[-] \textbf{Vaccinated\_No\_Booster}
          \item[-] \textbf{Booster}
        \end{itemize}
          \vspace{0.4em} 
      \end{minipage}
      \\ \hline

      Optimal Group Utility $f_g^*$ &
      \begin{minipage}[t]{\linewidth}\raggedright\sloppy
        \begin{itemize}
        \item[-] \textbf{Unvaccinated}:
        $C_1
        + K_1\!\cdot\!\bigl(
        \text{covid\_threat}\cdot
        \text{trust\_gov}\cdot
        \text{trust\_sci}\cdot
        \text{age}^{C_2}
        (C_3+\text{family\_covid})
        \bigr)
        - K_2\!\cdot\!\bigl(
        \text{risk\_covid\_gt\_vax}
        (C_4+\text{income\_unknown}\cdot
        e^{-K_3\cdot\text{more\_attention}\,\text{age}^{C_5}})
        \bigr)$
        
        \item[-] \textbf{Vaccinated\_No\_Booster}:
        $C_1
        - K_1\!\cdot\!\bigl(
        \text{age}^{C_2}(C_3+\text{low\_income})(C_4-\text{trust\_sci})
        \bigr)
        + K_2\!\cdot\!\bigl(
        \text{vax\_safe}\,\text{trust\_gov}\,
        (C_5+\text{more\_attention}\,\text{age}^{C_6})
        \bigr)$
        
        \item[-] \textbf{Booster}:
        $C_1
        + K_1\!\cdot\!\bigl(
        \text{vax\_protect\_long}\cdot
        e^{-K_2(\text{age}^{C_2}
        +\text{low\_income}\cdot\text{family\_covid})}
        \bigr)
        - K_3\!\cdot\!\bigl(
        \text{less\_attention}\cdot
        \text{trust\_sci}\,(\text{age}/C_3)
        \bigr)$
        \end{itemize}
          \vspace{0.4em} 
      \end{minipage}
      \\ \hline

      Optimal Personalized Decision Rule $\mathcal{P}_i^*$ &
      \begin{minipage}[t]{\linewidth}\raggedright\sloppy
        \begin{itemize}
          \item[-] \textbf{THREAT\_AVOIDING}: Perceives high disease risk; ranks \emph{Booster} $>$ \emph{Vaccinated\_No\_Booster} $>$ \emph{Unvaccinated}; considers logistics and side-effect concerns while relying on trusted sources.
        \end{itemize}
      \end{minipage}
      \\

      \arrayrulecolor{blue}\hline
      \rowcolor{lightBlue}
      \textbf{\textcolor{blue}{\textit{Vaccine} – Case 4}} & \\ \hline

      Features $\mathcal{X}$ &
      \begin{minipage}[t]{\linewidth}\raggedright
        \begin{itemize}
          \item[-] \textbf{Age}: 52
          \item[-] \textbf{Gender}: Male
          \item[-] \textbf{Education}: No university degree
          \item[-] \textbf{Income}: Below-median
          \item[-] \textbf{COVID-19 Threat Perception}: Strong
          \item[-] \textbf{Belief in Vaccine Prevention}: Low
          \item[-] \textbf{Trust in Science}: Moderate
          \item[-] \textbf{Risk Perception}: Disease risk $>$ vaccine risk
          \item[-] \textbf{Attention to Vaccine News}: Unchanged
          \item[-] \textbf{Family COVID}: None
        \end{itemize}
          \vspace{0.4em} 
      \end{minipage}
      \\

      \arrayrulecolor{mygray}\hline

      Alternatives $\mathcal{J}$ &
      \begin{minipage}[t]{\linewidth}\raggedright
        \begin{itemize}
          \item[-] \textbf{Unvaccinated}
          \item[-] \textbf{Vaccinated\_No\_Booster}
          \item[-] \textbf{Booster}
        \end{itemize}
          \vspace{0.4em} 
      \end{minipage}
      \\ \hline

      Optimal Group Utility $f_g^*$ &
      \begin{minipage}[t]{\linewidth}\raggedright\sloppy
        \begin{itemize}
        \item[-] \textbf{Unvaccinated}:
        $K_1\!\sqrt{
        \text{covid\_threat}\bigl(
        \text{risk\_covid\_gt\_vax}
        +\text{gender}\sqrt{\text{age}}+C_1
        \bigr)
        }
        \cdot\bigl(
        (\text{trust\_gov}\,\text{trust\_sci})^{C_2}
        + C_3
        \bigr)
        + K_2\!\cdot\!\text{more\_attention}
        - K_3\!\cdot\!\text{low\_income}
        \bigl(
        C_4+\text{has\_degree}\cdot
        \text{trust\_gov}\cdot
        \text{trust\_sci}
        \bigr)
        + C_5$
        
        \item[-] \textbf{Vaccinated\_No\_Booster}:
        $K_1\!\cdot\!\bigl(
        \text{vax\_safe}
        +\text{trust\_gov}\cdot
        \text{trust\_sci}\cdot
        \sqrt{
        \sqrt{\text{age}+C_1}
        +\text{income\_unknown}
        +C_2
        }
        + C_3
        \bigr)
        + K_2\!\cdot\!\dfrac{
        \text{more\_attention}
        }{
        \text{less\_attention}+C_4
        }
        + C_5$
        
        \item[-] \textbf{Booster}:
        $K_1\!\cdot\!\bigl(
        \text{family\_covid}
        +\text{physician}\cdot
        \text{trust\_gov}\cdot
        \text{trust\_sci}\,
        (\sqrt{\text{age}+C_1}+C_2)
        +\text{nurse}\cdot
        \text{trust\_sci}\sqrt{\text{age}+C_3}
        + C_4
        \bigr)
        + C_5$
        \end{itemize}
          \vspace{0.5em} 
      \end{minipage}
      \\ \hline

      Optimal Personalized Decision Rule $\mathcal{P}_i^*$ &
      \begin{minipage}[t]{\linewidth}\raggedright\sloppy
        \begin{itemize}
          \item[-] \textbf{BALANCED}: Cautious yet data-driven; moderate trust in authorities; open to boosters with clear evidence; weighs prior experiences and accessibility.
        \end{itemize}
      \end{minipage}
      \\
      \hline
    \end{longtable}
\endgroup



\subsection{Semantically Similar Choices Analysis}
\label{cm}

\FloatBarrier
\begin{figure}[t]
  \centering
  
  \begin{subfigure}[b]{0.9\linewidth}
    \centering
    \includegraphics[width=\textwidth]{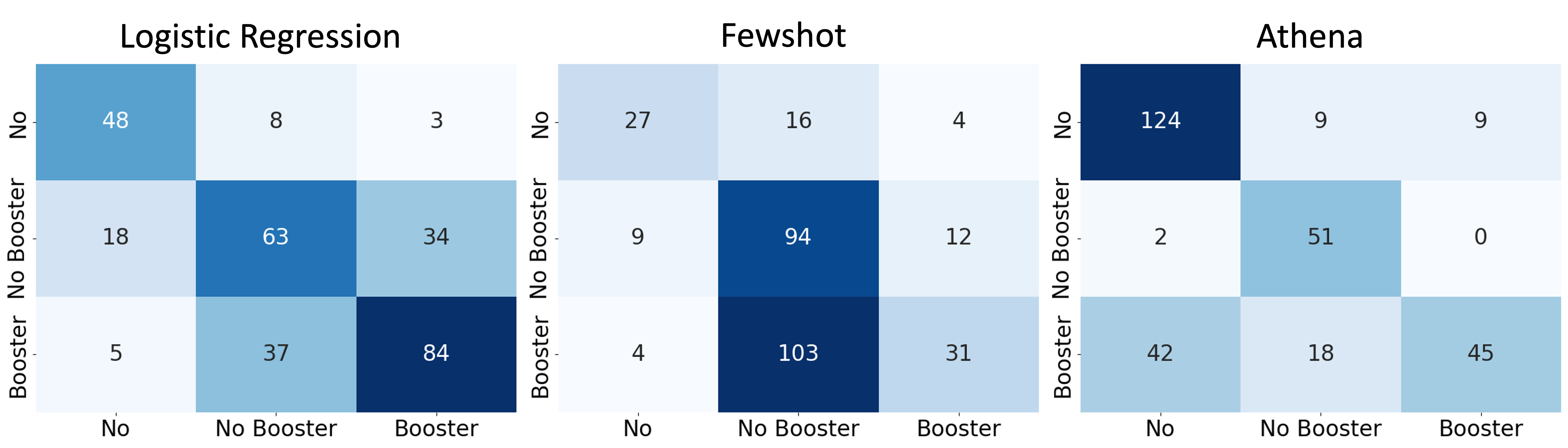} 
    \caption{\textbf{Vaccine-uptake task.} 
      \textsc{\model} removes all \emph{34} cases in which the \emph{Vaccinated\_no\_booster} class was previously misclassified as \emph{Booster}, thereby preserving the integrity of booster-demand estimates.}
    \label{fig:vaccine_conf}
  \end{subfigure}
  
  \vspace{1em} 
  
  \begin{subfigure}[b]{0.9\linewidth}
    \centering
    \includegraphics[width=\textwidth]{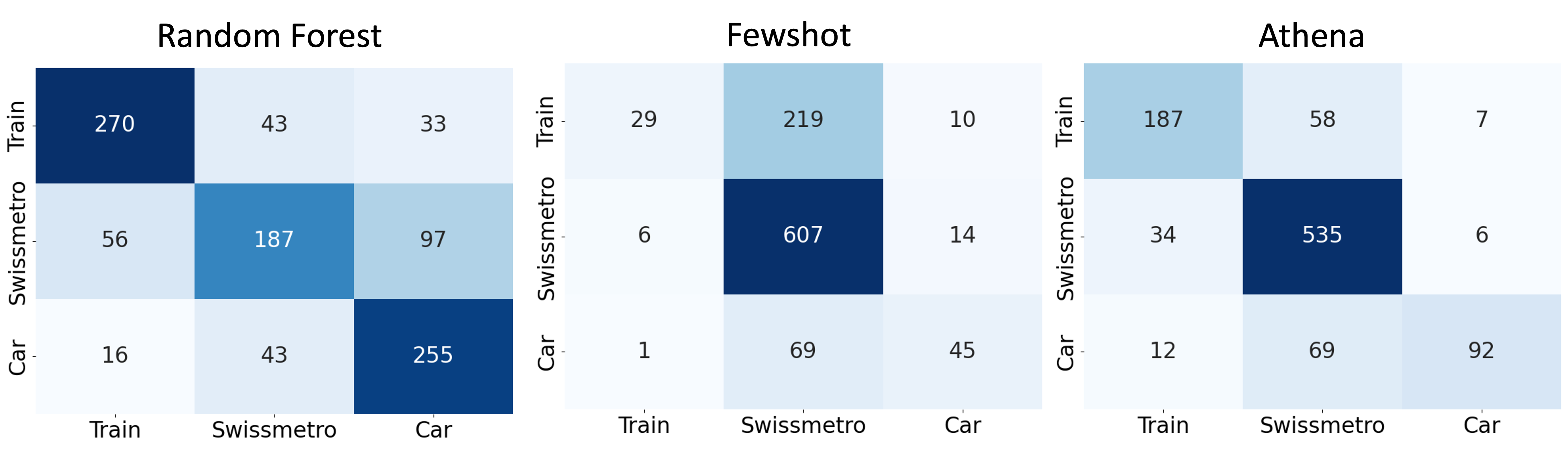}  
    \caption{\textbf{Travel-mode choice task.} 
      \textsc{\model} cuts the \emph{Swissmetro}-versus-\emph{Car} confusion from \emph{83} to \emph{6} instances, refining forecasts of low-carbon rail adoption.}
    \label{fig:travel_conf}
  \end{subfigure}
  
  \caption{\textsc{\model} yields improvements on the classes that matter most yet were previously hard to distinguish.}
  \label{fig:confusion_matrices}
\end{figure}

As illustrated in Figure~\ref{fig:confusion_matrices}, \textsc{\model} not only raises aggregated accuracy but also improves decision-critical boundaries, offering more reliable evidence for public-health and transport-policy planning.

\subsection{Extended Interpretability Showcase}
\label{symbolic_interpretability}
We provide representative full symbolic utilities discovered by \textsc{\model} on \textit{Swissmetro} and \textit{Vaccine} datasets. These examples illustrate how the symbolic structure translates into actionable insights for transportation and public health domains.
\subsubsection{Representative Example — Swissmetro Dataset}


\vspace{-1em}
\begin{table}[H]
\centering
\begin{tabularx}{\textwidth}{l X}
\toprule
\textbf{Mode} & \textbf{Discovered symbolic utility} \\ 
\midrule

\textbf{Train} & 
$K_1 \!\cdot\! \bigl(
\text{train\_time} + \text{metro\_time} 
+ \text{luggage}\cdot\log(\text{age}+C_1)
+ \text{age} + \text{is\_male}
\bigr)
+ C_2 \!\cdot\! (\text{first\_class} + \text{income})
- C_3 \!\cdot\! (\text{GA\_pass} + \text{headway})$ \\[6pt]

\textbf{Car} & 
$K_1 \!\cdot\! \bigl(
\text{car\_time} + \text{train\_time} 
+ \text{luggage}\cdot\log(\text{age}+C_1)
+ \text{age}
\bigr)
+ C_2 \!\cdot\! (\text{first\_class} + \text{income})
- C_3 \!\cdot\! (\text{GA\_pass} + \text{metro\_fare} + \text{is\_male})$ \\[6pt]

\textbf{Metro} & 
$K_1 \!\cdot\! \bigl(
\text{metro\_time} + \text{luggage} + \text{age} + \text{is\_male}
\bigr)
+ C_2 \!\cdot\! (\text{first\_class} + \text{income})
- C_3 \!\cdot\! (\text{headway} + \text{GA\_pass} + \text{is\_male})$ \\

\bottomrule
\end{tabularx}
\end{table}
\vspace{-1em}
\noindent
\textbf{Feature:} Between 39 and 54 years old, identify as female, and have an income between 50 and 100.

\paragraph{Key take-aways for domain experts}

\begin{itemize}[leftmargin=*]
  \item \textbf{Time dominates.} Large negative coefficients on travel-time variables show this segment is \textbf{highly time-sensitive} $\rightarrow$ investments that shorten door-to-door time (e.g., skip-stop service) should shift demand \cite{SHIRES2009315}.
  
  \item \textbf{Comfort premium.} Positive weight on $(\text{first\_class} + \text{income})$ across all modes indicates a willingness to pay for comfort that scales with income $\rightarrow$ targeted upselling (seat reservations, quiet cars) is effective \cite{abrantes2011}.
  
  \item \textbf{Luggage burden grows with age.} The interaction $\text{luggage}\cdot \log(\text{age}+C_1)$ reveals baggage becomes disproportionately painful for older travelers $\rightarrow$ facilities such as luggage trolleys or porter services may raise train/metro share \cite{chang2013}.
  
  \item \textbf{GA pass effect.} Owning a GA pass biases travellers away from modes that still incur extra fares. Extending GA coverage to Swissmetro would therefore raise its relative appeal \cite{weis2010}.
\end{itemize}

\vspace{-1em}
\begin{table}[H]
\centering
\begin{tabularx}{\textwidth}{l X}
\toprule
\textbf{Mode} & \textbf{Discovered symbolic utility} \\ 
\midrule

\textbf{Train} & 
$K_1 \!\cdot\! \Bigl(
\text{purpose} 
+ \bigl| \text{payer\_type}\!\cdot\! C_1 \bigr|
- \text{first\_class} 
+ \bigl| \text{luggage} \bigr| \sqrt{\bigl| \text{age} + C_2 \bigr|} 
+ \bigl| \text{train\_time} + C_3 \bigr| 
+ \log(\text{income} + C_4)
\Bigr)
- C_5$ \\[6pt]

\textbf{Car} & 
$K_1 \!\cdot\! \Bigl(
\bigl| \text{car\_time} + C_1 \bigr|
+ \bigl| \text{car\_time} - \text{train\_time} + C_2 \bigr|
- \bigl| \text{car\_cost} + \text{train\_cost} + C_3 \bigr|
+ \bigl| \text{headway} \bigr| \sqrt{\text{income} + C_4}
\Bigr)
+ C_5$ \\[6pt]

\textbf{Metro} & 
$K_1 \!\cdot\! \Bigl(
\bigl| \text{metro\_time} + C_1 \bigr|
+ \bigl| \text{metro\_cost} + C_2 \bigr|
+ \sqrt{\bigl| \text{age} + C_3 \bigr|}
+ \log\!\Bigl(\exp(\text{income} + C_4) + C_5\Bigr)
\Bigr)
- C_6$ \\

\bottomrule
\end{tabularx}
\end{table}
\vspace{-1em}
\noindent
\textbf{Feature:} Male travelers younger than 24 years, annual income 50–100k.

\paragraph{Key take-aways for domain experts}
\begin{itemize}[leftmargin=*]
  \item \textbf{Time still trumps money.} Travel time appears in all utilities, while fare only in Car/Metro. For under-25 travelers, each minute lost matters more than an extra franc $\rightarrow$ prioritizing faster transfers or signal priority is especially effective \cite{Frank2008}.
  
  \item \textbf{Headway frustration fuels car use.} The term $ \text{headway} \cdot \sqrt{\text{income}}$ shows that infrequent trains push young people toward cars, and irritation rises with income $\rightarrow$ higher-frequency rail services can curb car switching \cite{Liao2020}.
  
  \item \textbf{First-class indifference.} The negative $ \text{first\_class}$ coefficient suggests little interest in upgrades $\rightarrow$ amenities in standard class (Wi-Fi, gaming lounges) may be more persuasive than premium seating \cite{Marques2025}.
\end{itemize}

\subsubsection{Representative Example — Vaccine Dataset}
\vspace{-1em}
\begin{table}[H]
\centering
\begin{tabularx}{\textwidth}{l X}
\toprule
\textbf{Mode} & \textbf{Discovered symbolic utility} \\
\midrule

\textbf{Unvaccinated} & 
$C_1 \!\cdot\! \text{covid\_threat}
\cdot \Bigl(
C_2 + \text{trust\_government}\cdot\text{trust\_science}
\cdot \log(\text{age} + C_3)
\Bigr)
\cdot \text{risk\_of\_covid\_greater\_than\_vax}
+ K_1 \!\cdot\! 
\text{have\_covid\_sick\_family\_member}
\cdot \log(\text{age} + C_4)$ \\[6pt]

\textbf{Vaccinated (no booster)} & 
$C_1 \!\cdot\! \text{covid\_threat}
+ C_2 \!\cdot\! \text{vaccine\_safe\_to\_me}
+ K_1 \!\cdot\!
\bigl(
\text{trust\_government}\cdot
\text{trust\_science}\cdot
\text{more\_attention\_to\_vax\_info}\cdot
\sqrt{\text{age} + C_3}
\bigr)$ \\[6pt]

\textbf{Booster} & 
$C_1 \!\cdot\! e^{\text{age}^{C_2}}
\cdot \text{covid\_threat}
\cdot \sqrt{\text{vax\_protect\_long\_yes}}
+ C_3 \!\cdot\! \text{vaccine\_safe\_to\_me}
+ K_1 \!\cdot\!
\bigl(
\text{trust\_government}\cdot
\text{trust\_science}\cdot
\text{nurse}\cdot
\sqrt{\text{age} + C_4}
\bigr)$ \\

\bottomrule
\end{tabularx}
\end{table}
\vspace{-1em}

\noindent
\textbf{Feature:} Age 18--38, income above county median.

\paragraph{Key take-aways for domain experts}

\begin{itemize}[leftmargin=*]
  \item \textbf{Risk trade-off in vaccination choice.} The product $\text{covid\_threat} \times \text{risk\_of\_covid\_greater\_than\_vax}$ captures a critical decision-making trade-off. Messaging must narrow this perceived risk gap, e.g., by emphasizing robust evidence on vaccine safety \cite{green2023}.
  
  \item \textbf{Booster demand rises steeply with age.} The factor $e^{\text{age}^{C_2}}$ generates a nonlinear age effect: as age increases, perceived vaccine benefit grows rapidly. This reflects age-associated increases in risk perception and vulnerabilities \cite{noh2022}.
  
  \item \textbf{Prior belief and healthcare occupation.} The presence of $\text{vax\_protect\_long\_yes}$ and nurse occupation in the booster equation means emphasizing extended protection and occupation will push this group further along the vaccination ladder \cite{biswas2021}.
  
  \item \textbf{Trust is pivotal for vaccine uptake.} The multiplicative $\text{trust\_government}\times \text{trust\_science}$ term appears in every vaccinated utility, signalling that confidence in both institutions amplifies willingness \cite{trent2022}.
\end{itemize}

\vspace{-1em}
\begin{table}[H]
\centering
\begin{tabularx}{\textwidth}{l X}
\toprule
\textbf{Mode} & \textbf{Discovered symbolic utility} \\ 
\midrule

\textbf{Unvaccinated} & 
$K_1 \!\cdot\!
\sqrt{
\text{covid\_threat}
\cdot
\bigl(
\text{risk\_of\_covid\_greater\_than\_vax}
+ \sqrt{\text{age}}\!\cdot\!\text{gender}
+ C_1
\bigr)
}
\cdot
\bigl(
(\text{trust\_government}\!\cdot\!\text{trust\_science})^{2}
+ C_2
\bigr)
+ K_2 \!\cdot\! \text{more\_attention\_to\_vax\_info}
- K_3 \!\cdot\!
\bigl(
\text{income\_below\_median}
\!\cdot\!
\text{have\_university\_degree}
\!\cdot\!
(\text{trust\_government}\!\cdot\!\text{trust\_science})
\bigr)
+ C_3$ \\[6pt]

\textbf{Vaccinated (no booster)} & 
$K_1 \!\cdot\!
\bigl(
\text{vaccine\_safe\_to\_me}
+ \text{trust\_government}\!\cdot\!\text{trust\_science}
\!\cdot\!
\sqrt{
\sqrt{\text{age}+C_1}
+ \text{income\_unknown}
+ C_2
}
\bigr)
+ K_2 \!\cdot\!
\dfrac{
\text{more\_attention\_to\_vax\_info}
}{
\text{less\_attention\_to\_vax\_info}+C_3
}
+ C_4$ \\[6pt]

\textbf{Booster} & 
$K_1 \!\cdot\!
\bigl(
\text{have\_covid\_sick\_family\_member}
+ \text{physician}\!\cdot\!
(\text{trust\_government}\!\cdot\!\text{trust\_science})
\!\cdot\!
(\sqrt{\text{age}+C_1}+C_2)
+ \text{nurse}\!\cdot\!
(\text{trust\_science}\!\cdot\!\sqrt{\text{age}+C_3})
\bigr)
+ C_4$ \\

\bottomrule
\end{tabularx}
\end{table}
\vspace{-1em}
\noindent
\textbf{Feature:} Adults with varied trust, income, and education profiles.

\paragraph{Key take-aways for domain experts}
\begin{itemize}[leftmargin=*]
  \item \textbf{Information attention as lever.} Positive weights on $\text{more\_attention\_to\_vax\_info}$ indicate that engagement with vaccine information consistently increases uptake $\rightarrow$ interactive campaigns remain essential \cite{Glanz2017}.
  
  \item \textbf{Nonlinear trust amplification.} The squared term $(\text{trust\_government}\cdot\text{trust\_science})^2$ highlights a super-additive effect $\rightarrow$ boosting both trust dimensions together disproportionately reduces hesitancy \cite{Lohmann2022}.
  
  \item \textbf{Education buffers income hesitancy.} The negative income effect is mitigated by education–trust interactions $\rightarrow$ higher education plus trust can offset low-income hesitancy, pointing to education-focused outreach \cite{Bajos2022}.
\end{itemize}

\section{Baseline Setup}

\subsection{Utility-Based Models}
\vspace{-5mm}
\begin{table}[htbp]
  \caption{Utility-based models and key settings (train : test = 0.8 : 0.2)}
  \vspace{-3mm}
  \label{tab:choice-hparams}
  \begin{tabularx}{\textwidth}{p{0.30\textwidth} >{\raggedright\arraybackslash}X}
    \arrayrulecolor{blue}\hline
    \rowcolor{lightBlue}
    \textbf{Model} & \textbf{Key (Non-default) Settings} \\ \hline
    SimpleMNL &
      \texttt{intercept="item"}; \texttt{optimizer="adam"} \\[2pt]
    ConditionalLogit &
      \texttt{optimizer="adam"}; added intercept for items 1 \& 2 \\[2pt]
    Latent Class MNL &
      \texttt{n\_latent\_classes=2}; \texttt{fit\_method="mle"};
      \texttt{optimizer="adam"}; \texttt{epochs=1000} \\ \hline
  \end{tabularx}
\end{table}

\subsection{Machine Learning Models}

\vspace{-5mm}
\begin{table}[htbp]
  \centering\small
  \caption{Machine learning models and key settings (train : test = 0.8 : 0.2)}
  \vspace{-2mm}
  \label{tab:ml-hparams}
  \begin{tabularx}{\textwidth}{
      >{\raggedright\arraybackslash}p{0.22\textwidth}  
      >{\raggedright\arraybackslash}X}                 
    \arrayrulecolor{blue}\hline
    \rowcolor{lightBlue}
    \textbf{Model} & \textbf{Best hyper-parameters} \\ \hline

    Logistic Regression &
    \ttfamily C=10,\; penalty=l2,\; solver=saga \\[4pt]

    Random Forest &
    \ttfamily bootstrap=False,\; max\_depth=None,\; min\_samples\_leaf=1,\;
    min\_samples\_split=2,\; n\_estimators=600 \\[4pt]

    XGBoost &
    \ttfamily colsample\_bytree=0.8,\; learning\_rate=0.05,\; max\_depth=6,\;
    n\_estimators=500,\; subsample=0.8 \\[4pt] \hline
  \end{tabularx}
\end{table}

\subsection{LLM-Based Models}
Take \textbf{Swissmetro} dataset as an example.

\begin{hintbox}[hb:example]{\textit{Swissmetro} - Zeroshot}
[SYS] You are a decision assistant that predicts a probability distribution over three travel modes, Swissmetro, Train, and Car, for a single trip.

You will receive two blocks of text:

<TRIP_INFO>  
… details like trip purpose, luggage, payment, origin, destination …  
</TRIP_INFO>

<TRANSPORT_OPTIONS>  
… list of modes with travel time, cost, headway …  
</TRANSPORT_OPTIONS>

**Instructions:**  

1. Use only the information in <TRIP_INFO> and <TRANSPORT_OPTIONS>.  

2. Estimate a probability for each mode so they sum to 1. 

3. **Output only** a JSON object, for example:  

```json
\{
  "Swissmetro": <float between 0 and 1>,
  "Train":      <float between 0 and 1>,
  "Car":        <float between 0 and 1>
\}
```

No additional text; just the JSON object with normalized probabilities.

[USR] <TRIP_INFO>
\{trip_info\}
</TRIP_INFO>

<TRANSPORT_OPTIONS>
\{transport_options\}
</TRANSPORT_OPTIONS>
\end{hintbox}

\begin{hintbox}[hb:example]{\textit{Swissmetro} - Zeroshot-CoT}
[SYS] You are a decision assistant that predicts a probability distribution over three travel modes, Swissmetro, Train, and Car, for a single trip.

You will receive two blocks of text:

<TRIP_INFO>  
… details like trip purpose, luggage, payment, origin, destination …  
</TRIP_INFO>

<TRANSPORT_OPTIONS>  
… list of modes with travel time, cost, headway …  
</TRANSPORT_OPTIONS>

**Instructions:**  

1. Use only the information in <TRIP_INFO> and <TRANSPORT_OPTIONS>.  

2. Estimate a probability for each mode so they sum to 1.  

3. **Output only** a JSON object, for example:  

```json
\{
  "Swissmetro": <float between 0 and 1>,
  "Train":      <float between 0 and 1>,
  "Car":        <float between 0 and 1>
\}
```

No additional text; just the JSON object with normalized probabilities.

Let's think step-by-step.

[USR] <TRIP_INFO>
\{trip_info\}
</TRIP_INFO>

<TRANSPORT_OPTIONS>
\{transport_options\}
</TRANSPORT_OPTIONS>
\end{hintbox}

\newpage
\begin{hintbox}[hb:example]{\textit{Swissmetro} - Fewshot}
[SYS] You are a decision assistant that predicts a probability distribution over three travel modes—Swissmetro, Train, and Car—for a set of travel records.

You will receive multiple records. Each record consists of three blocks:
<TRIP_INFO>  
… trip details: purpose, luggage, payment, origin, destination …  
</TRIP_INFO>
<TRANSPORT_OPTIONS>  
… each mode's travel time, cost, headway …  
</TRANSPORT_OPTIONS>
<CHOICE>  
… either a JSON object with probabilities (for examples), or left empty for the record to predict …  
</CHOICE>

**Instructions:**  
- For records where <CHOICE> is filled, treat them as examples.  
- For the final record (with an empty <CHOICE>), output **only** the JSON object of normalized probabilities (summing to 1), with no extra text.

[USR] <TRIP_INFO>
\{trip_info_1\}
</TRIP_INFO>
<TRANSPORT_OPTIONS>
\{transport_options_1\}
</TRANSPORT_OPTIONS>
<CHOICE>
\{choice_1\}
</CHOICE>
<TRIP_INFO>
\{trip_info_2\}
</TRIP_INFO>
<TRANSPORT_OPTIONS>
\{transport_options_2\}
</TRANSPORT_OPTIONS>
<CHOICE>
\{choice_2\}
</CHOICE>
<TRIP_INFO>
\{trip_info_3\}
</TRIP_INFO>
<TRANSPORT_OPTIONS>
\{transport_options_3\}
</TRANSPORT_OPTIONS>
<CHOICE>
\{choice_3\}
</CHOICE>
<TRIP_INFO>
\{trip_info_4\}
</TRIP_INFO>
<TRANSPORT_OPTIONS>
\{transport_options_4\}
</TRANSPORT_OPTIONS>
<CHOICE>
\{choice_4\}
</CHOICE>
<TRIP_INFO>
\{trip_info_5\}
</TRIP_INFO>
<TRANSPORT_OPTIONS>
\{transport_options_5\}
</TRANSPORT_OPTIONS>
<CHOICE>
\{choice_5\}
</CHOICE>
<TRIP_INFO>
\{trip_info_6\}
</TRIP_INFO>
<TRANSPORT_OPTIONS>
\{transport_options_6\}
</TRANSPORT_OPTIONS>
<CHOICE>
Please predict the travel mode for this trip.
</CHOICE>
\end{hintbox}

\begin{hintbox}[hb:example]{\textit{Swissmetro} - TextGrad}
[INITIAL FULL PROMPT + SOLUTION] Task: Estimate the probability distribution over three travel modes
(Swissmetro, Train, Car) for a single trip.

<TRIP_INFO>
\{trip_info\}
</TRIP_INFO>

<TRANSPORT_OPTIONS>
\{transport_options\}
</TRANSPORT_OPTIONS>

Solution (JSON):
\{{"Swissmetro": 0.333, "Train": 0.333, "Car": 0.334\}}

[GRADING PROMPT] You are a transport-economics expert.
Given the trip info, transport options, and predicted probabilities in the
user's message, output a single line ONLY:
Score: <float between 0 and 1>
1 = probabilities look highly reasonable, 0 = implausible.
Remember: THE PREDICTION MUST BE A JSON DICT.
\end{hintbox}

\newpage 
\section{Additional Experiments: Reasoning LLMs and End-to-End Baselines}
\label{moreexp}

\noindent\textbf{Purpose and setup.}
This section probes how much backbone model capacity matters on our tasks. 
For a controlled comparison, we randomly sample $100$ individuals from the $500$-person pool to form a compact evaluation subset (same preprocessing, metrics, and decoding settings as in the main experiments). For each individual, we randomly sample one record.
We evaluate \textsc{\model} with five backbones: two state-of-the-art open-source reasoning models (\texttt{Qwen3-32B}, \texttt{DeepSeek-R1-Distill-Qwen-32B}) and three leading commercial offerings (\texttt{GPT-4o-mini}, \texttt{GPT-4o}, \texttt{Gemini-2.0-Flash}). 
Across both tasks, \textsc{\model} attains \emph{state-of-the-art classification performance} among LLM-based methods—consistently delivering the highest \emph{Accuracy} and \emph{F1}, with \emph{AUC} that is competitive or superior to prompt-only LLM baselines (see Tables~\ref{tab:results} and~\ref{tab:baseline_comparison}). 

\medskip
\noindent\textbf{Structure dominates model size; stronger reasoning yields modest, consistent gains.}
Under \textsc{\model}, swapping \texttt{GPT-4o-mini} for larger “reasoning” backbones (e.g., \texttt{GPT-4o}, \texttt{Qwen3-32B}, \texttt{DeepSeek-R1}) yields \emph{incremental} but \emph{consistent} improvements, especially on the more interaction-heavy \emph{Vaccine} task. 
The effect is smaller on \emph{Swissmetro}, where dominant explanatory factors (time/cost) are already well captured by the \emph{symbolic discovery $\to$ textual refinement} pipeline. 
Intuitively, Stage~1 constrains the hypothesis space to interpretable utility forms, and Stage~2 makes small, directed edits to those forms; this turns the problem into guided search plus local adjustments. 
As a result, \emph{structural bias} (symbolic utility discovery + semantic adaptation) shoulders most of the lift, while \emph{backbone capacity} primarily fine-tunes edge cases (nonlinear interactions, atypical profiles), producing a steady but not dramatic gain.

\medskip
\noindent\textbf{Prompt-only methods are brittle and poorly calibrated; \textsc{\model} regularizes both decisions and probabilities.}
Zero-shot / CoT / Few-shot prompting shows visible volatility across metrics: \emph{Accuracy/F1} can spike on one dataset yet drop on another, and \emph{AUC/CE} often swing with decoding details (temperature, sampling count, score-to-probability mapping). 
\textsc{\model} markedly reduces this variance: the symbolic stage enforces cross-person consistency (shared operators, shared concept library), while the textual refinement stage adjusts \emph{within} those constraints, leading to better class separability and more conservative probability mass. 
Empirically this manifests as stronger and more stable \emph{F1/AUC}, with \emph{CE} reflecting improved calibration compared to prompt-only baselines. 
In short, structure acts as \emph{regularization} for both decisions and confidence.

\medskip
\noindent\textbf{End-to-end baselines trail on interpretability and robustness; \textsc{\model}’s decomposition captures heterogeneity with explicit utility logic.}
Machine learning-based models can be competitive on single metrics in isolated settings, but they do not expose explicit, policy-relevant utility functions and are less consistent across tasks/splits. 
They must implicitly learn both \emph{which} attributes matter and \emph{how} they combine, from scratch. 
\textsc{\model} instead \emph{decouples} the problem: Stage~1 discovers globally interpretable utility structure (operators, interactions), and Stage~2 adapts those structures to individual semantics. 
This yields \textit{(i)} stronger across-task consistency in \emph{Accuracy/F1/AUC}, \textit{(ii)} end-to-end interpretability of the discovered utilities.

\begin{table*}[t]
\centering
\caption{Performance comparison across methods on \textit{Swissmetro} and \textit{Vaccine} datasets.}
\label{tab:results}
\setlength{\tabcolsep}{4.5pt}
\renewcommand{\arraystretch}{1.15}

\begin{adjustbox}{max width=0.9\textwidth}
\begin{tabular}{l l cccc cccc}
\toprule
\textbf{Method} & \textbf{LLM Model}
& \multicolumn{4}{c}{\textbf{Swissmetro}}
& \multicolumn{4}{c}{\textbf{Vaccine}} \\
\cmidrule(lr){3-6}\cmidrule(lr){7-10}
& & Acc.$\uparrow$ & F1$\uparrow$ & CE$\downarrow$ & AUC$\uparrow$
& Acc.$\uparrow$ & F1$\uparrow$ & CE$\downarrow$ & AUC$\uparrow$ \\
\midrule
Zeroshot        & gemini-2.0-flash & 0.5800 & 0.3046 & 0.9059 & 0.6829 & 0.6000 & 0.5386 & 0.8317 & 0.7433 \\
                & GPT-4o-mini      & 0.6100 & 0.2763 & 0.9253 & 0.5556 & 0.5500 & 0.5302 & 0.8271 & 0.7500 \\
                & GPT-4o           & 0.5900 & 0.3310 & 0.8646 & 0.6946 & 0.6000 & 0.5465 & 0.8052 & 0.7306 \\
                & Qwen3            & 0.5900 & 0.4158 & 1.4047 & 0.6126 & 0.5400 & 0.5729 & 0.8676 & 0.7519 \\
                & DeepSeek-r1      & 0.5900 & 0.4339 & 1.2473 & 0.6608 & 0.6300 & \textbf{0.6531} & 0.8244 & 0.7692 \\
Zeroshot-CoT    & gemini-2.0-flash & 0.5300 & 0.2809 & 0.9858 & 0.6409 & 0.6100 & 0.5485 & 0.8128 & 0.7604 \\
                & GPT-4o-mini      & 0.6100 & 0.2763 & 0.9109 & 0.6156 & 0.5900 & 0.5820 & 0.8161 & 0.7714 \\
                & GPT-4o           & 0.5800 & 0.3162 & 0.9237 & 0.6404 & 0.6300 & 0.5717 & 0.7785 & 0.7700 \\
                & Qwen3            & 0.6200 & 0.4632 & 1.7101 & 0.6284 & 0.5400 & 0.5843 & 0.9073 & 0.7364 \\
                & DeepSeek-r1      & 0.5800 & 0.3018 & 0.9522 & 0.6173 & 0.5200 & 0.5492 & 0.8761 & 0.7373 \\
Few-shot        & gemini-2.0-flash & 0.7200 & 0.6922 & 10.0922 & 0.7984 & 0.5300 & 0.5508 & 14.2531 & 0.6747 \\
                & GPT-4o-mini      & 0.6800 & 0.5320 & 5.0402  & 0.7516 & 0.5200 & 0.5278 & 7.7066  & 0.6975 \\
                & GPT-4o           & 0.7300 & 0.6654 & 3.9386  & 0.8423 & 0.5700 & 0.5967 & 5.6261  & 0.7467 \\
                & Qwen3            & 0.7400 & 0.6760 & 7.0435  & 0.8033 & 0.5400 & 0.5487 & 9.5013  & 0.7060 \\
                & DeepSeek-r1      & 0.7000 & 0.6282 & 3.6154  & 0.8439 & 0.5500 & 0.5392 & 8.8895  & 0.6855 \\
TextGrad        & gemini-2.0-flash & 0.5400 & 0.2432 & 1.1934  & 0.4718 & 0.5000 & 0.4511 & 4.1290  & 0.7345 \\
                & GPT-4o-mini      & 0.5700 & 0.3111 & 0.9551  & 0.5292 & 0.5000 & 0.4686 & 4.7960  & 0.6460 \\
                & GPT-4o           & 0.5600 & 0.3316 & 0.9441  & 0.6256 & 0.5600 & 0.5321 & 2.3468  & 0.6721 \\
                & Qwen3            & 0.5100 & 0.3669 & 2.0180  & 0.5322 & 0.4600 & 0.4276 & 6.4994  & 0.6303 \\
                & DeepSeek-r1      & 0.5800 & 0.3356 & 0.9631  & 0.6344 & 0.4300 & 0.4235 & 3.1080  & 0.5948 \\
\textbf{\textsc{\model} (ours)} & gemini-2.0-flash & \textbf{0.7900} & 0.7185 & \textbf{0.6121} & \textbf{0.9153} & 0.6500 & 0.5978 & 0.8305 & 0.7998 \\
                & GPT-4o-mini      & 0.7600 & \textbf{0.7304} & 1.4208 & 0.8577 & 0.6500 & 0.6079 & 0.8034 & 0.8133 \\
                & GPT-4o           & 0.7700 & 0.7085 & 1.0417 & 0.8697 & \textbf{0.6700} & 0.6213 & \textbf{0.7765} & \textbf{0.8279} \\
                & Qwen3            & 0.7400 & 0.7040 & 4.9132 & 0.7754 & 0.5700 & 0.5650 & 1.1393 & 0.7637 \\
                & DeepSeek-r1      & 0.7100 & 0.6612 & 0.8437 & 0.8353 & 0.6600 & 0.6501 & 0.8115 & 0.8212 \\
\bottomrule
\end{tabular}
\end{adjustbox}
\end{table*}

\newpage
\section{Empirical Scalability Evidence}
\label{app:scalability}
We benchmarked wall-clock time and token usage using \texttt{gpt-4o-mini} on the Swissmetro subset.

\subsection{Stage 2 – Individual adaptation}
\begin{table}[h]
\centering
\caption{Runtime and token usage for Stage 2 (individual-level semantic adaptation) under different iteration counts $T'$.}
\begin{tabular}{c c c c}
\toprule
$T'$ & Time (s) & s/iter & tokens/iter \\
\midrule
1 & 48.16 & 48.16 & 1079.6 \\
3 & 176.89 & 58.96 & 1195.83 \\
5 & 315.66 & 63.13 & 1249.28 \\
\bottomrule
\end{tabular}
\end{table}

\subsection{Stage 1 – Group-level discovery}
\begin{table}[h]
\centering
\caption{Runtime and token usage for Stage 1 (group-level symbolic utility discovery) under different iteration counts $T$.}
\begin{tabular}{c c c}
\toprule
$T$ & Time (min) & tokens total \\
\midrule
5 & 30.61 & 215{,}281 \\
15 & 36.69 & 251{,}974 \\
30 & 65.64 & 479{,}751 \\
\bottomrule
\end{tabular}
\end{table}

These results confirm that runtime and token usage scale approximately linearly with the number of iterations, consistent with the theoretical analysis.

\newpage
\section{Prompts}

Take \textbf{Swissmetro} dataset as an example.

\begin{hintbox}[hb:example]{\textit{Swissmetro} - Symbolic Utility Initialization}

Step 1:

[SYS] You are a transportation planner specializing in analyzing the relationships among 
various factors that influence travel behavior. You will be provided with two types of information: individual 
features (delimited by <FEATURES> and </FEATURES>) and preliminary travel mode knowledge (delimited by <KNOWLEDGE> 
and </KNOWLEDGE>). Your task is to carefully review these inputs and in detailed sentence describe how the provided 
features interrelate. Ensure your response includes as many specific details as possible about the relationships, 
but do not propose any new features or suggest modifications to the existing ones. Example: Time: quadratic, Cost: log, luggage: linear.

[USR] <GROUP DESCRIPTION>\{description\}</GROUP DESCRIPTION>
<FEATURES>\{features\}</FEATURES>
<KNOWLEDGE>\{knowledge\}</KNOWLEDGE>

You should ONLY provide the relations between the features.
YOU MUST return your assumption in this exact format: ```["relation_0","relation_1", ...]

Step 2:

[SYS] You are a helpful assistant that proposes mathematical expressions based on some provided 
suggestions. Your goal is to:

0. **Task**: Generate utility functions for travel mode choice of group of \{description\}.

1. **Use only** the specified variables: \{variables\}

2. **Represent all constants** with the symbol "C", and all coefficients with the symbol "K".

3. **Restrict** yourself to the following operators: {operators}

4. **For each group**, suggest utility functions for train, car, and Swissmetro respectively.

Your response must:
- Propose exactly **\{N\}** groups of expressions.
- MUST return in this exact format: ```[("expressions_car","expressions_train","expressions_metro"), ...]```, replace expressions_mode with your proposed expressions.

[USR] Suggestions: \{suggestions\}

\end{hintbox}

\begin{hintbox}[hb:example]{\textit{Swissmetro} - Results Analysis}
[SYS] You are a creative and insightful mathematical research assistant. You have been provided with two sets of utility expressions: one function group labeled “Good Expressions” and one labeled “Bad Expressions.” Your objective is to hypothesize about the underlying assumptions or principles that might generate the good expressions yet exclude the bad ones. 

Key Points:

1. Focus primarily on the good expressions' mathematical structures and any connections they might have to physical or applied contexts. 

2. Capital “C” in any expression is just an arbitrary constant.

3. Do not discuss or compare the expressions in terms of their simplicity or complexity.

4. Provide your reasoning step by step, but keep it very concise and genuinely insightful. No more than 5 lines.

[USR] Good Expression 1: (train: \{texpr1\}, car: \{cexpr1\}, metro: \{mexpr1\}), accuracy: \{acc1\}

Good Expression 2: (train: \{texpr2\}, car: \{cexpr2\}, metro: \{mexpr2\}), accuracy: \{acc2\}

Bad Expression 1: (train: \{bexpr1\}, car: \{bexpr2\}, metro: \{bexpr3\}), accuracy: \{acc3\}

Above expressions are travel mode choice utility functions of group of \{description\}. Propose \{N\} hypotheses that would be appropriate given the expressions. Provide short commentary for each of your decisions. Do not talk about topics related to the simplicity or complexity of the expressions. I want ideas that are unique and interesting enough to amaze the world's best mathematicians.
\end{hintbox}











\newpage
\begin{hintbox}[hb:example]{\textit{Swissmetro} - Crossover}
[SYS] You are a helpful assistant that recombines two mathematical expressions based on some provided suggestions. Your goal is to produce new expressions that:\\
1. Blend or merge elements from both reference expressions in a way that reflects the suggestions.\\
2. Adhere to the following constraints:\\
- You may only use the variables in library: \{variables\}\\
- All constants must be represented with the symbol C\\
- Only the following operators are allowed: \{operators\}\\
Guidelines:\\
- Propose exactly \{N\} new expressions.\\
- Each new expression should integrate elements of both reference expressions. You can also propose new terms with variables that are in the library but not in the old expressions.\\
- If any suggestions appear contradictory, reconcile them reasonably.\\
MUST return in this exact format:\\
\texttt{[("expressions\_car","expressions\_train","expressions\_metro"), ...]}\\
Replace \texttt{expressions\_} with your proposed expressions.

[USR] Suggestion: \{suggestions\}\\
Reference Expression group 1: (train: \{texpr1\}, car: \{cexpr1\}, metro: \{mexpr1\})\\
Reference Expression group 2: (train: \{texpr2\}, car: \{cexpr2\}, metro: \{mexpr2\})\\
\\
Propose \{N\} expressions that would be appropriate given the suggestions and references.\\
\end{hintbox}

\begin{hintbox}[hb:example]{\textit{Swissmetro} - Mutation}
[SYS] You are a helpful assistant that generates mutated variants of a **triplet** of mathematical expressions
(car, train, metro) based on provided mutation strategies.  
Your goal is to produce new expression triplets that:  
1. Mutate the reference expressions by applying mutation operations (e.g., adjust coefficients, swap variables,
   alter operators) in a way that reflects the suggestions.  
2. Adhere to the following constraints:  
   - You may only use the variables in library: \{variables\}  
   - All constants must be represented with the symbol C  
   - Only the following operators are allowed: \{operators\}  

Guidelines:  
- Produce exactly \{M\} mutated **triplets**.  
- Within each triplet you must provide one mutated expression for **car**, one for **train** and one for **metro**.  
- A mutation can modify any combination of variable, operator or constant, but each expression must remain
  syntactically valid under the constraints.  

MUST return in this exact format:  

\texttt{[("mut\_car1","mut\_train1","mut\_metro1"), ...]}  

[USR] Generate \{M\} mutated variants of the following mathematical expression triplet according to these mutation strategies:  
– You may only use variables from: \{variables\}  
– All constants must be written as C  
– Only these operators are allowed: \{operators\}  

Mutation strategies: \{suggestions\}  

Reference expressions:  
(car): \{cexpr\}  
(train): \{texpr\}  
(metro): \{mexpr\}  

Please return exactly \{M\} new, syntactically valid triplets in the JSON list format shown above.
\end{hintbox}

\newpage
\begin{hintbox}[hb:example]{\textit{Swissmetro} - Semantic Adaptation Initialization}
[SYS] You are a travel-behavior preference selector.
You will be given two blocks of information:

<DEMOGRAPHICS> ... </DEMOGRAPHICS>
<UTILITY\_FUNCTION> ... </UTILITY\_FUNCTION>

Your goal: choose the single best-matching high-level preference template for this group **exactly** from the catalogue below and output **only** the template name (uppercase).

CATALOGUE
- TIME\_EFFICIENCY   : travellers primarily minimise total travel time.

- COST\_SAVING       : travellers primarily minimise direct monetary cost.

- COMFORT\_SEEKING   : travellers value comfort/service frequency and dislike crowding.

- BALANCED          : sensitivities are evenly distributed across factors.

...//OTHER POSSIBLE TEMPLATE

Return nothing else — no commentary, no punctuation, just the template name.

[USR] <DEMOGRAPHICS>\{demographics\}</DEMOGRAPHICS>

<UTILITY\_FUNCTION>\{utility\}</UTILITY\_FUNCTION>
\end{hintbox}

\begin{hintbox}[hb:example]{\textit{Swissmetro} - Semantic Adaptation Loss Function}
Evaluate the travel mode prediction based on the individual's profile and alternatives. Compare it to the actual choice and identify any discrepancies. Be concise and focus on why the prediction might be incorrect. Return 0 if they match, 1 otherwise.
\end{hintbox}

\begin{hintbox}[hb:example]{\textit{Swissmetro} - Prediction}
[SYS] You are a decision assistant that recommends the most suitable travel mode for an individual trip by estimating a probability distribution over three options: Swissmetro, Train, and Car.

You will receive three blocks:
<TEMPLATE> … optimized preference template … </TEMPLATE>

<PROFILE> … individual profile … </PROFILE>

<ALTERNATIVES> … attributes of Swissmetro, Train, and Car … </ALTERNATIVES>

**Instructions:**
1. **Use the <TEMPLATE> as a guide** for understanding the individual's likely preference bias (e.g., time efficiency, cost saving, comfort seeking, balanced).

2. **Analyze the <PROFILE>** (age, gender, income, trip details) **and the <ALTERNATIVES>** (travel time, cost, headway).

3. **Estimate and output a probability** for each travel mode, such that all three probabilities sum to 1.

**Output format (JSON only):**
```json
\{
  "Swissmetro": <float between 0 and 1>,
  "Train":      <float between 0 and 1>,
  "Car":        <float between 0 and 1>
\}
```
No additional text; just the JSON object with normalized probabilities.

[USR] <TEMPLATE>\{template_name\}</TEMPLATE>
<PROFILE>
\{individual_block\}
</PROFILE>
<ALTERNATIVES>
\{options\}
</ALTERNATIVES>
\end{hintbox}


\end{document}